\documentclass[10pt,twocolumn,letterpaper]{article}

\usepackage{cvpr}              

\usepackage{color}
\usepackage{colortbl}
\definecolor{tableBlue}{rgb}{0.424,0.557,0.749}
\definecolor{lightblue}{HTML}{dfebf7}
\definecolor{verylightgray}{HTML}{EDEDED}
\newcommand{\tableLineColorGray}{\rowcolor{verylightgray}}
\definecolor{persiangreen}{rgb}{0.0, 0.65, 0.58}
\definecolor{cadmiumred}{rgb}{0.89, 0.0, 0.13}
\newcommand{\tableLineColor}{\rowcolor{lightblue}}
\definecolor{increaseGreen}{rgb}{0.51,0.702,0.4}
\definecolor{decreaseRed}{rgb}{0.647,0.337,0.318}

\usepackage{amsthm}
\usepackage{booktabs}
\usepackage{algorithm}
\usepackage{algpseudocode}
\usepackage{amsmath}
\usepackage{amssymb}
\usepackage{mathtools}
\newtheorem{theorem}{Theorem}

\usepackage{pifont}
\usepackage{makecell}
\usepackage{titletoc}

\titlecontents{section}[3.8em]
  {}
  {\hyperlink{section.\thecontentslabel}{\contentslabel{1.8em}}}  
  {\hspace*{-2.em}}
  {\titlerule*[0.5pc]{.}\contentspage}

\titlecontents{subsection}[6.5em]
  {}
  {\hyperlink{subsection.\thecontentslabel}{\contentslabel{2.5em}}}
  {\hspace*{-3.em}}
  {\titlerule*[0.5pc]{.}\contentspage}

\newcommand{\cmark}{\textcolor{green!70!black}{\ding{51}}}  
\newcommand{\xmark}{\textcolor{red!80!black}{\ding{55}}}  

\definecolor{cvprblue}{rgb}{0.21,0.49,0.74}
\usepackage[pagebackref,breaklinks,colorlinks,allcolors=cvprblue]{hyperref}

\def\paperID{6063} 
\def\confName{CVPR}
\def\confYear{2026}

\title{CHIPS: Efficient CLIP Adaptation via\\Curvature-aware Hybrid Influence-based Data Selection}

\author{
    {Xinlin Zhuang}\textsuperscript{1,2,5} \quad
    {Yichen Li}\textsuperscript{1,3} \quad
    {Xiwei Liu}\textsuperscript{1} \quad
    {Haolin Yang}\textsuperscript{1} \quad
    {Yifan Lu}\textsuperscript{1} \\
    {Ziyun Zou}\textsuperscript{1} \quad
    {Yulong Li}\textsuperscript{1} \quad
    {Huifa Li}\textsuperscript{1} \quad  
    {Dongliang Chen}\textsuperscript{2} \quad
    {Qinglei Wang}\textsuperscript{1} \\
    {Weiyang Liu}\textsuperscript{5} \quad
    {Ying Qian}\textsuperscript{2}\thanks{Corresponding authors.} \quad
    {Jiangming Shi}\textsuperscript{4}\footnotemark[1] \quad
    {Imran Razzak}\textsuperscript{1}\footnotemark[1] \\[6pt] 
    \textsuperscript{1}MBZUAI \quad
    \textsuperscript{2}East China Normal University \quad
    \textsuperscript{3}Huazhong University of Science and Technology \\
    \textsuperscript{4}Xiamen University \quad
    \textsuperscript{5}The Chinese University of Hong Kong \\
    {\tt\footnotesize xinlinzhuang@stu.ecnu.edu.cn, yqian@cs.ecnu.edu.cn, jiangming.shi@outlook.com, imran.razzak@mbzuai.ac.ae}
}

\begin{document}
\maketitle
\begin{abstract}
Adapting CLIP to vertical domains is typically approached by novel fine-tuning strategies or by continual pre-training (CPT) on large domain-specific datasets. 
Yet, data itself remains an underexplored factor in this process.
We revisit this task from a data-centric perspective: {Can effective data selection substitute for large-scale datasets in CPT?}
We introduce \textbf{CHIPS} (\textbf{C}urvature-aware \textbf{H}ybrid \textbf{I}nfluence in \textbf{P}rojection \textbf{S}ubspace), which assigns each image-text pair a utility score that integrates three complementary factors aligned with three goals: \textit{faithfulness} via a curvature-aware and Newton-style alignment computed in CLIP's end-point subspace; \textit{scalability} via an InfoNCE-aware curvature estimator with Johnson-Lindenstrauss (JL) sketching; and \textit{retention} via a selection-aware relevance weight combined with learnability to balance target adaptation against general-domain preservation. 
We justify this design theoretically by proving a lower-bound guarantee on the proxy's correlation with full-parameter alignment and by characterizing the bias-variance trade-offs introduced by curvature mixing and JL sketching.
We evaluate CHIPS empirically across various settings: 1) CHIPS attains state-of-the-art performance among selection baselines on \textbf{17 medical benchmarks}, matches full-dataset CPT with 30\% of the data, and outperforms half-dataset CPT using only 10\%; 2) on \textbf{31 general-domain benchmarks}, CHIPS yields the least performance drop under all retention ratios. 
\end{abstract}    
\section{Introduction}
\label{sec:intro}

\begin{table}[t]
  \centering
  \footnotesize
  \setlength{\tabcolsep}{4pt}
\begin{tabular}{lccccc}
  \toprule
  \textbf{Method} &
  \makecell{\textbf{Target}} &
  \makecell{\textbf{Sketch}} &
  \makecell{\textbf{Curvature}} &
  \makecell{\textbf{Domain}} &
  \makecell{\textbf{CLIP}} \\
  \midrule
  Full CPT              & \xmark & \xmark & \xmark & \xmark & \xmark \\
  Random Selection      & \xmark & \xmark & \xmark & \xmark & \xmark \\
  Rule-filter           & \xmark & \xmark & \xmark & \xmark & \xmark \\
  CLIPscore             & \xmark & \xmark & \xmark & \xmark & \xmark \\
  Dot                   & \cmark & \cmark & \xmark & \xmark & \xmark \\
  TracIn                & \cmark & \cmark & \xmark & \xmark & \xmark \\
  TRAK                  & \cmark & \cmark & \cmark & \xmark & \xmark \\
  \textbf{CHIPS (ours)} & \cmark & \cmark & \cmark & \cmark & \cmark \\
  \bottomrule
\end{tabular}
  \caption{Properties of data selection methods for CLIP adaptation. \textit{Target}: whether the strategy utilizes target evaluation set information. \textit{Sketch}: whether JL Sketch is supported. \textit{Curvature}: whether second-order information is employed.  \textit{Domain}: whether general-target domain balance is supported.  \textit{CLIP}: whether the method is specifically designed for CLIP models.}
  \label{tab:method_comparison}
\end{table}

Vision-language models such as CLIP \cite{clip} achieve strong zero-shot recognition in general domains, but their performance degrades sharply in vertical settings (e.g., medical imaging and biology) where vocabulary, acquisition protocols, and label taxonomies shift substantially \cite{medical_clip_survey}. 
To adapt CLIP to such domains, two main paradigms have emerged: 1) \textit{Model-centric} methods modify training or parameterization (e.g., probabilistic fine-tuning \cite{clap4clip}, many-to-many contrastive learning \cite{cplip}, and PEFT variants \cite{SPT, BSR, clip_ast, clip_refine, liu2024parameter}); 2) \textit{Data-centric} methods pursue continual pre-training (CPT) on large, domain-specific datasets, ranging from millions to hundreds of millions of pairs across medical and biodiversity settings \cite{pmc-clip,biomedclip,biomedica,quilt1m,ophthalmology,dermatology,bioclip,bioclip2,biotrove,bioscan,dalip,openinsects}. 
However, for vertical domains, collecting, curating, and processing increasingly large datasets is costly, and indiscriminate upsampling can even impair learning by introducing redundant, low-utility samples. 
This naturally raises a question: \textit{Is sheer scale truly necessary for effective CPT?}

Meanwhile, a complementary line of work studies data attribution \cite{if_survey} to quantify how individual samples affect training dynamics and downstream generalization.
Classical influence functions~\cite{if} and variants such as TracIn~\cite{tracin}, EL2N~\cite{EL2N}, Arnoldi/GEX/FVM~\cite{arnoldi,gex,fvm}, were developed for single-tower model on supervised classification with additive cross-entropy and full-parameter updates. 
In CLIP, three mechanisms break these assumptions and systematically mis-rank examples. 
\textbf{(A) Cross-modal curvature of dual encoders.} 
CLIP’s dual encoders induce non-block-diagonal second-order curvature and block-diagonal proxies ignore this coupling and mis-rank samples.
\textbf{(B) Non-local gradients under InfoNCE.} 
Each sample’s gradient depends on the softmax normalizer over the full negative set, making influence \textit{batch-/global-dependent} rather than per-example additive and thus sensitive to batch size, queue design, and negative composition.
\textbf{(C) Endpoint dominance at the projection heads.} 
Empirically, the projection heads and temperature parameter drive early shifts in similarity distributions, rendering full-parameter influence computation relatively unnecessary for CLIP.

Motivated by (A)-(C), we propose \textbf{CHIPS} (\textbf{C}urvature-aware \textbf{H}ybrid \textbf{I}nfluence in \textbf{P}rojection \textbf{S}ubspace), a CLIP-specific selector that scores each candidate by the expected one-step drop of the target evaluation loss and selects the top examples for CPT. 
CHIPS is designed for three goals, \textit{faithfulness}, \textit{scalability}, and \textit{retention}, with one component per goal:
(i) a Newton-style alignment computed on the projection heads and temperature that \textit{provably lower-bounds} full-parameter alignment and enjoys better conditioning (Sec.~\ref{sec:alignment});
(ii) an InfoNCE-aware curvature preconditioner built from positive/negative gradient moments and compressed via Johnson-Lindenstrauss (JL) sketching to achieve near-linear time/memory with a quantified variance-bias trade-off (Sec.~\ref{sec:neg_curv});
and (iii) a selection-aware \textit{soft} weighting that combines per-sample learnability with target-domain relevance to explicitly control the adaptation-retention balance (Sec.~\ref{sec:weights}).
Tab.~\ref{tab:method_comparison} positions CHIPS against several common baselines from practical perspectives.

Across \textbf{17 medical benchmarks}, CHIPS \textit{matches} full-dataset CPT using only 30\% of the training pool, and \textit{outperforms} half-dataset CPT using merely 10\% data, achieving SOTA among selection baselines. 
For \textbf{31 general-domain benchmarks}, while domain adaptation inevitably reduces average performance, CHIPS consistently \textit{reduces} the drop relative to the previous SOTA selector, indicating stronger retention of general-domain capabilities. 
In short, with principled attribution and CLIP-aware curvature, \textit{effective CPT does not require extreme scale}.

Our contributions are threefold\footnote{\url{https://github.com/mihara-bot/CHIPS}.}.
First, we propose an alignment score computed in end-point space of CHIP, which ranks samples by expected one-step decrease on evaluation loss.
Second, we incorporate negative-pair geometry to capture cross-example curvature and analyze the variance-bias trade-off of the estimator.
Third, we combine learnability and domain-relevance weights to emphasize decision-boundary, on-domain samples while softly limiting selection drift and mitigating catastrophic forgetting.
Experimental results demonstrate the effectiveness of CHIPS on both medical and general downstream tasks.
\begin{figure*}[t]
  \centering
  \includegraphics[width=1.0\textwidth]{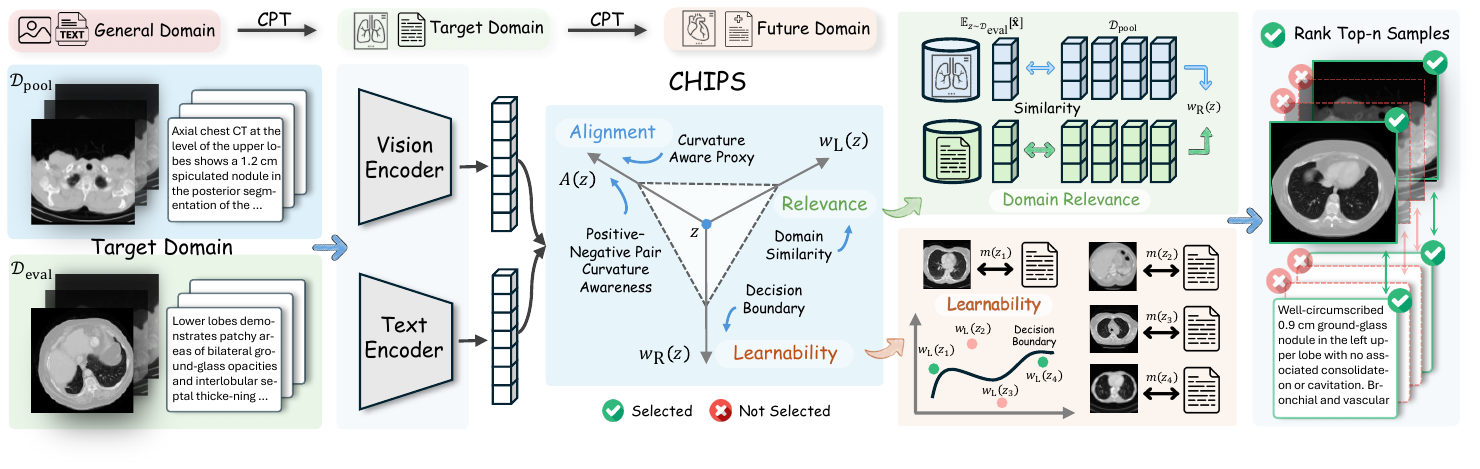}
  \caption{Workflow of CHIPS. For each training sample, CHIPS computes a curvature-aware proxy Newton alignment in CLIP’s end-point space (projection heads and temperature), where curvature is approximated by mixing self and negative-pair cross moments from symmetric InfoNCE and scaled efficiently via JL sketching. The alignment is then modulated by learnability and target-domain relevance to yield a single selection utility, and the top-$n$ samples are chosen to CPT CLIP models for domain adaptation.}

  \label{fig:workflow}
\end{figure*}

\section{Method}
\label{sec:method}

CHIPS ranks training samples using an \textit{alignment} score that estimates how much a one-step update on a given sample would reduce the evaluation loss on the target domain, together with auxiliary \textit{learnability} and \textit{domain relevance} scores.
More specifically, CHIPS consists of three tightly coupled components:
(i) a curvature-aware proxy alignment score computed in CLIP’s end-point geometry, namely the projection heads and temperature;
(ii) a negative-pair curvature estimator that captures the second-order coupling induced by symmetric InfoNCE and is made scalable through JL sketching; and
(iii) two multiplicative weights for learnability and target-domain relevance, forming a single selection utility. 
Fig.~\ref{fig:workflow} provides the workflow of CHIPS.

\subsection{Problem Setup}
\label{sec:setup}

We consider a pre-trained general CLIP model with parameters $\theta$ and a large target-domain training pool $\mathcal{D}_{\text{pool}}=\{z_i\}_{i=1}^N$ of image-text pairs.
Our goal is to select a valuable subset from $\mathcal{D}_{\text{pool}}$ for CPT so as to maximize performance on a held-out target-domain evaluation set $\mathcal{D}_{\text{eval}}$ consisting of samples from various validation sets of downstream tasks.
Let $\mathcal{L}_{\text{eval}}(\theta)\triangleq \mathbb{E}_{z\sim \mathcal{D}_{\text{eval}}}[\ell(z;\theta)]$ denote the evaluation loss, 
where $\ell$ is the symmetric InfoNCE objective \cite{clip} with a learnable logit-scale (temperature) $\tau>0$. 
Denote the evaluation mean gradient by $\mathbf{u}\triangleq \nabla_\theta \mathcal{L}_{\text{eval}}(\theta)=\mathbb{E}_{z\sim \mathcal{D}_{\text{eval}}}[\nabla_\theta \ell(z;\theta)]$.
During CPT, one Stochastic Gradient Descent (SGD) step draws a mini-batch of size $B$ from a selection distribution $q$ and computes the training gradient 
$\widehat{\mathbf{g}}=\tfrac{1}{B}\sum_{i=1}^B \nabla_\theta \ell(z_i;\theta)$ to update parameters via $\theta'=\theta-\eta\,\widehat{\mathbf{g}}$, where $\eta>0$ is the learning rate.
We also define the training mean gradient under $q$ by $\mathbf{g}_q \triangleq \mathbb{E}_{z\sim q}[\nabla_\theta \ell(z;\theta)]$.
Assume $\mathcal{L}_{\text{eval}}$ is twice continuously differentiable in a neighborhood of $\theta$ and let 
$\mathbf{H}_{\text{eval}}(\theta)=\nabla_\theta^2 \mathcal{L}_{\text{eval}}(\theta)$.
A second-order Taylor expansion with integral remainder and Hessian-Lipschitz constant $L_H$ yields the following bound for $\Delta\theta=-\eta\,\widehat{\mathbf{g}}$:
\begin{equation}
\label{eq:delta_eval_second}
\begin{aligned}
    \mathbb{E}\big[\Delta  \mathcal{L}_{\text{eval}}\big]
    \;\le\;&
    -\eta\, \mathbf{g}_q^\top \mathbf{u}
    +\tfrac{1}{2}\,\eta^2\,\mathbb{E}\!\big[\,\widehat{\mathbf{g}}^{\!\top}\mathbf H_{\text{eval}}(\Theta)\,\widehat{\mathbf{g}}\,\big] \\
    &+\tfrac{L_H}{6}\,\eta^3\,\mathbb{E}\!\big[\|\widehat{\mathbf{g}}\|^{3}\big],
\end{aligned}
\end{equation}
for some $\Theta$ on the segment between $\theta$ and $\theta'$.
In a local quadratic model, the steepest descent direction is the Newton direction 
$\mathbf{H}_{\text{eval}}^{-1}\mathbf{u}$, and a good selection distribution $q$ should \textbf{increase} the first-order alignment term $-\eta\,\mathbf{g}_q^\top \mathbf{u}$ while controlling curvature.
For a more lightweight assumption, if $\mathcal{L}_{\text{eval}}$ is $\rho$-smooth (gradient-Lipschitz), then
\begin{equation}
\label{eq:delta_eval_first}
\mathbb{E}\!\left[\Delta \mathcal{L}_{\text{eval}}\right]
\;\le\;
-\eta\,\mathbf{g}_q^\top \mathbf{u}
+\tfrac{\rho}{2}\,\eta^2\,\mathbb{E}\!\big[\|\widehat{\mathbf{g}}\|^2\big].
\end{equation}
Eq.~\eqref{eq:delta_eval_second} and Eq.~\eqref{eq:delta_eval_first} emphasize the same principle:
\textit{descent is driven by the alignment between the expected training gradient and the evaluation mean gradient}.
While we present SGD for clarity, the alignment direction naturally extends to AdamW and full proofs are provided in App.~\ref{app:proofs_alignment_direction}. 
These observations motivate us to prioritize samples whose updates align with (a proxy of) the Newton direction; the construction of such a curvature-aware proxy in CLIP’s end-point geometry is given next in Sec.~\ref{sec:alignment}.

\subsection{Curvature-aware Proxy Alignment}
\label{sec:alignment}

CHIPS is implemented on the end-point subspace of CLIP
\(\vartheta=\{\mathbf W_v,\mathbf W_t,\tau\}\), consisting of the two projection heads and the temperature.
For a sample \(z\), let $\mathbf g_\vartheta(z)=\nabla_\vartheta \ell(z;\theta)$
denote its gradient restricted to \(\vartheta\), and let \(\mathbf u_\vartheta\) be the corresponding mean evaluation gradient.
Motivated by the one-step descent view in Sec.~\ref{sec:setup}, the ideal update direction on this subspace is the Newton direction
\(\mathbf H_\vartheta^{-1}\mathbf u_\vartheta\), where \(\mathbf H_\vartheta\) denotes a population curvature on \(\vartheta\).
Accordingly, we define the curvature-aware proxy alignment score:
\begin{equation}
\label{eq:alignment_score_main}
A(z)\ \triangleq\ \mathbf g_\vartheta(z)^\top \mathbf M^{-1}\mathbf u_\vartheta,
\end{equation}
where \(\mathbf M\succ 0\) is a tractable curvature surrogate detailed in Sec.~\ref{sec:neg_curv}.
A large positive \(A(z)\) means that a one-step update on \(z\) moves the model more strongly along a descent direction for the evaluation loss.

\par
Our central hypothesis is that alignment measured in the end-point subspace is a reliable proxy for alignment in the full parameter space.
Eq.~\eqref{eq:alignment_score_main} can be rewritten as 
$A(z)=\tilde{\mathbf g}_\vartheta(z)^\top \tilde{\mathbf u}_\vartheta$ with
$\tilde{\mathbf g}_\vartheta=\mathbf M^{-1/2}\mathbf g_\vartheta$ and 
$\tilde{\mathbf u}_\vartheta=\mathbf M^{-1/2}\mathbf u_\vartheta$.
This indicates that the curvature-aware score is simply the standard inner-product alignment after a fixed, sample-independent preconditioning.
Therefore, the proxy-full analysis below applies directly to this curvature-aware form.
The explicit transformed covariance and coupling operators are given in App.~\ref{app:proxy_influence}.
To formalize this proxy view, consider the local linearization
\begin{equation}
\label{eq:proxy_linearization}
\nabla_\theta \ell(z)=\mathbf J\,\mathbf g_\vartheta(z)+\mathbf r(z),
\qquad
\mathbf u=\bar{\mathbf J}\,\mathbf u_\vartheta+\boldsymbol\varepsilon,
\end{equation}
where \(\mathbf J\) is a local linear map that lifts a sample's end-point gradient
\(\mathbf g_\vartheta(z)\) to its full-parameter gradient,
and \(\bar{\mathbf J}\) is the analogous local linear map for the mean evaluation gradient.
The residual terms \(\mathbf r(z)\) and \(\boldsymbol\varepsilon\) capture the corresponding approximation errors.

Define the subspace alignment $X(z)=\mathbf g_\vartheta(z)^\top \mathbf u_\vartheta$ and the full alignment $Y(z)=\nabla_\theta \ell(z)^\top \mathbf u$.
Let
$\Sigma_g=\operatorname{Cov}[\mathbf g_\vartheta(z)]$,
$\mathbf S=\tfrac12(\mathbf J^\top\bar{\mathbf J}+\bar{\mathbf J}^\top\mathbf J)$,
$\mathbf B=\Sigma_g^{1/2}\mathbf S\,\Sigma_g^{-1/2}$, and let $\sigma_\zeta^2=\operatorname{Var}[\zeta(z)]$ denote the variance of the mismatch term induced by \((\mathbf r,\boldsymbol\varepsilon)\). 
We have the following lower bound:

\begin{theorem}[Proxy-full alignment correlation]
\label{thm:proxy_general}
If \(\zeta(z)\) is uncorrelated with \(\mathbf g_\vartheta(z)\), then the Pearson correlation between \(X(z)\) and \(Y(z)\) satisfies
\begin{equation}
\label{eq:proxy_lb_general}
\rho_{XY}\ \ge\
\frac{\lambda_{\min}\!\big(\mathrm{sym}(\mathbf B)\big)}
{\sqrt{\ \|\mathbf B\|_2^2\;+\;\sigma_\zeta^2\,/\,\big(\mathbf u_\vartheta^\top \Sigma_g\,\mathbf u_\vartheta\big)\ }}\,.
\end{equation}
\end{theorem}

Theorem~\ref{thm:proxy_general} shows that the proxy score tracks the full score when two conditions hold:
(i) the subspace-to-full map is well aligned along \(\mathbf u_\vartheta\), as summarized by \(\mathbf S\), and
(ii) the residual mismatch variance \(\sigma_\zeta^2\) is not too large.
Under approximate isotropy or commuting structure, the bound simplifies further (provided in App.~\ref{app:proxy_influence}).

However, CHIPS uses the alignment score for \textit{ranking} samples.
Therefore, the relevant empirical quantity is the \textit{Spearman correlation} between proxy and full scores.
On a 100K subset of BIOMEDICA \cite{biomedica}, we observe a Spearman correlation of 0.83, proving that end-point subspace scores preserve the ordering induced by full alignment well.

\subsection{Curvature Estimation}
\label{sec:neg_curv}

Symmetric InfoNCE couples each positive pair with multiple negatives through the softmax normalizer, creating \textit{cross-example} curvature that a positive-only or diagonal surrogate fails to capture.
Since our alignment score $A(z)$ relies on a curvature-preconditioned inner product, missing this cross mass yields a biased Newton proxy and unstable rankings along highly coupled directions.
Working in the end-point subspace $\vartheta$, we approximate the population curvature by combining \textit{self} and \textit{cross} second moments of subspace gradients:
\begin{equation}
\label{eq:phi_posneg}
\begin{aligned}
\Phi_{\mathrm{pos}}
&=\tfrac{1}{N}\sum_{i=1}^N \mathbf g_\vartheta(z_i)\,\mathbf g_\vartheta(z_i)^\top, \\
\Phi_{\mathrm{neg}}
&=\tfrac{1}{N(N-1)}\!\!\sum_{i\neq j}\mathbf g_\vartheta(z_i)\,\mathbf g_\vartheta(z_j)^\top,
\end{aligned}
\end{equation}
where $\Phi_{\mathrm{pos}}$ captures \textit{self} curvature, while $\Phi_{\mathrm{neg}}$ restores the \textit{cross} mass induced by random negatives in InfoNCE.
We then form a trace- and spectrum-controlled surrogate
\begin{equation}
\label{eq:h_m_def}
\mathbf H_\vartheta^{(\alpha)}=(1-\alpha)\,\Phi_{\mathrm{pos}}+\alpha\,\Phi_{\mathrm{neg}},
\qquad
\mathbf M=\mathbf H_\vartheta^{(\alpha)}+\lambda\,\mathbf I,
\end{equation}
with a mixing weight $\alpha\in[0,1]$ and a small Tikhonov parameter $\lambda>0$ to ensure numerical stability in matrix inversion.
Plugging $\mathbf M$ into Eq.~\eqref{eq:alignment_score_main} yields the \textit{curvature-aware} proxy Newton score $A(z)$.
By mixing $\Phi_{\mathrm{pos}}$ with the cross moment $\Phi_{\mathrm{neg}}$ (with $\alpha>0$), we restore the off-diagonal curvature induced by InfoNCE’s negatives that positive-only surrogates ignore. 
The surrogate $\mathbf M\!\succ\!0$ thus preconditions $A(z)$ in a fixed metric without altering the proxy-full correlation framework (Sec.~\ref{sec:alignment}); $\alpha$ controls coupling strength and $\lambda$ provides ridge stability. 
This yields an alignment score that is sensitive to the true second-order structure yet robust.
Let the ideal scalar be $A^\star(z)=\mathbf g_\vartheta(z)^\top \mathbf H_\vartheta^{-1}\mathbf u_\vartheta$ and the curvature-approximated score $A_\alpha(z)=\mathbf g_\vartheta(z)^\top \mathbf M^{-1}\mathbf u_\vartheta$ with $\mathbf M$ as in Eq.~\eqref{eq:h_m_def}, and define $\Delta_\alpha=\mathbf H_\vartheta-\mathbf H_\vartheta^{(\alpha)}$.
For further efficiency, we apply a $k$-dimensional JL sketch to compute a sketched score $\widehat A_\alpha(z)$.
We discuss the choice of JL sketch in Sec.~\ref{sec:rp}.

\begin{theorem}[Error bound for curvature mixing]
\label{thm:var_bias_alignment_simpler}
Assume bounded spectra for $\mathbf H_\vartheta$ and $\mathbf H_\vartheta^{(\alpha)}$.
Then, with probability at least $1-\delta$ over the JL sketch (vacuous if no sketch is used), there exist constants $C_1,C_2>0$, depending only on spectral bounds of $\mathbf H_\vartheta$ and $\mathbf H_\vartheta^{(\alpha)}$, such that
\begin{equation}
\label{eq:sketch_var_bias_s33}
\begin{aligned}
\mathbb E_z\!\big[\big(\widehat A_\alpha(z)-A^\star(z)\big)^2\big]
&\le \underbrace{\frac{C_1\log(1/\delta)}{k}}_{\text{projection variance}}
\\
&\quad+\underbrace{C_2\,\|\Delta_\alpha\|_F^2\,\|\mathbf H_\vartheta^{-1}\mathbf u_\vartheta\|_2^2}_{\text{curvature bias}}.
\end{aligned}
\end{equation}
\end{theorem}

Eq.~\eqref{eq:sketch_var_bias_s33} separates an $O(1/k)$ \textit{projection variance} term and a \textit{curvature bias} term that shrinks as $\mathbf H_\vartheta^{(\alpha)}$ better matches $\mathbf H_\vartheta$.
Because $\Phi_{\mathrm{neg}}$ restores off-diagonal mass whenever cross-example coupling is present, any $\alpha>0$ that does so reduces $\|\Delta_\alpha\|_F$ and tightens the bound; $\lambda$ stabilizes the inverse without changing the decomposition.
All derivations and proofs are deferred to App.~\ref{app:neg_curv_proof}.

\subsection{Learnability and Domain Relevance}

\label{sec:weights}

The curvature-aware alignment score $A(z)$ in Eq.~\eqref{eq:alignment_score_main} indicates whether a sample points in a \textit{useful} descent direction, but it does not distinguish already-solved cases from borderline ones, nor does it compensate for the distributional gap between $\mathcal D_{\text{pool}}$ and $\mathcal D_{\text{eval}}$.
Therefore, we introduce another two multiplicative weights to modulate $A(z)$ by \textit{learnability} and \textit{target-domain relevance}.

\paragraph{Learnability.}
For a mini-batch of paired image--text examples $\{(x_i,y_i)\}_{i=1}^B$, let the end-point embeddings be
$\hat{\mathbf x}_i=\frac{\mathbf W_v\mathbf h_i}{\|\mathbf W_v\mathbf h_i\|}$ and
$\hat{\mathbf y}_j=\frac{\mathbf W_t\mathbf t_j}{\|\mathbf W_t\mathbf t_j\|}$ with temperature $\tau>0$.
The CLIP similarity is $s_{ij}=\tau\,\hat{\mathbf x}_i^\top \hat{\mathbf y}_j$, with softmax probabilities
$p^{i2t}_{ij}=\frac{\exp(s_{ij})}{\sum_{j'}\exp(s_{ij'})}$ and
$p^{t2i}_{ij}=\frac{\exp(s_{ij})}{\sum_{i'}\exp(s_{i'j})}$.
Define the average correctness and a hardest-negative margin as
\begin{equation}
\begin{aligned}
p_{\mathrm{corr}}(z) & = \tfrac12\big(p^{i2t}_{ii}+p^{t2i}_{ii}\big),\\
m(z)                 & = s_{ii}-\max\!\big\{\max_{j\neq i}s_{ij},\,\max_{i'\neq i}s_{i'i}\big\}.
\end{aligned}
\end{equation}
We then weight toward decision-boundary examples and away from saturated ones:
\begin{equation}
\label{eq:wL}
w_{\mathrm L}(z)=\big(1-p_{\mathrm{corr}}(z)\big)\,\big(1+\sigma(-m(z))\big),
\end{equation}
where $\sigma(\cdot)$ is the sigmoid function. 
The first factor downweights already-correct and high-confidence pairs to reduce their impact while the second emphasizes small/negative margins that are most learnable in one step.

\paragraph{Target-domain Relevance.}
To softly favor samples consistent with $\mathcal D_{\text{eval}}$, we compute evaluation embeddings $\boldsymbol\mu_x=\mathbb E_{z\sim\mathcal D_{\text{eval}}}[\hat{\mathbf x}]$, $\boldsymbol\mu_y=\mathbb E_{z\sim\mathcal D_{\text{eval}}}[\hat{\mathbf y}]$, and define
\begin{equation}
\label{eq:wR}
w_{\mathrm R}(z)=\sigma\!\Big((1-\beta)\,\cos(\hat{\mathbf x},\boldsymbol\mu_x)+\beta\,\cos(\hat{\mathbf y},\boldsymbol\mu_y)\Big), 
\end{equation}
where $\beta\in[0,1]$ controls the balance between two modalities and $\cos(\cdot,\cdot)$ denotes the cosine similarity.
Notably, the relevance logit $s(z)=(1-\beta)\,\cos(\hat{\mathbf x},\boldsymbol\mu_x)+\beta\,\cos(\hat{\mathbf y},\boldsymbol\mu_y)$ in Eq.~\eqref{eq:wR} lies in $[-1,1]$, therefore $w_{\mathrm R}(z)\in[\sigma(-1),\sigma(1)]\approx[0.27,0.73]$.
This implements soft re-weighting to some extent rather than hard filtering as no sample is zeroed out.
\par
Hence, \(w_{\mathrm R}\) acts as a gentle target-domain bias rather than a hard filter: it nudges selection toward target-relevant samples while preventing the final distribution from drifting too far from the original alignment-driven one.
This helps balance adaptation and retention, mitigating catastrophic forgetting of general-domain knowledge.

\paragraph{Final utility score and selection.}
We combine the components multiplicatively into a single ranking score.
\begin{equation}
\label{eq:final_utility}
\mathcal I_{\text{CHIPS}}(z)=\widehat A_\alpha(z)\cdot w_{\mathrm L}(z)\cdot w_{\mathrm R}(z),
\end{equation}
and select the top-$n$ examples from $\mathcal D_{\text{pool}}$ for CPT.
Implementation details of CHIPS are provided in App.~\ref{app:data_selection}.
\section{Experiment}

\begin{table*}[tb]
\centering
\small
\renewcommand{\arraystretch}{1.05}
\setlength{\tabcolsep}{8.75pt} 
\begin{tabular}{lccccccccccc}
\toprule
\multicolumn{1}{c}{\textbf{Model}}    & \multicolumn{9}{c}{\textbf{Medical}}                                            & \multicolumn{2}{c}{\textbf{General}} \\ \midrule
\multicolumn{1}{c}{}                  & OPH   & RAD   & DER   & HEM   & PAT   & NEU   & HIS   & BIO   & Avg             & CLS               & RET              \\ \midrule
PubMedCLIP                            & 35.92 & 20.27 & 9.63  & 8.45  & 25.77 & 60.50 & 8.45  & 25.27 & 26.21           & 19.05             &   24.21               \\
BioMedCLIP                            & 14.42 & 15.59 & 1.20  & 10.14 & 21.17 & 44.79 & 4.66  & 72.91 & 21.45           & 7.68              &  23.55                \\
BMCLIP                                & 38.47 & 17.44 & 61.65 & 9.44  & 49.75 & 38.26 & 5.02  & 17.28 & 29.86           & 57.97             & 28.27            \\ \midrule
Vanilla                               & 15.91 & 19.76 & 10.82 & 17.95 & 30.09 & 57.96 & 5.24  & 10.56 & 23.51           & 53.15             &  29.23                \\
\tableLineColorGray Full Dataset                                  & 33.30 & 12.75 & 11.72 & 23.33 & 49.57 & 82.79 & 12.40 & 9.81  & 31.51           & 49.72             &  24.20                \\ \midrule
\textit{$r$=10\%}                     &       &       &       &       &       &       &       &       &                 &                   &                  \\
Random                                & 36.01 & 9.93  & 11.27 & 20.61 & 39.71 & 43.96 & 4.68  & 10.18 & 24.78           & \textbf{52.21}             & 29.28            \\
Concept-Balance                       & 33.18 & 12.11 & 11.67 & 21.81 & 40.40 & 46.82 & 5.03  & 10.87 & 25.50           & \underline{51.95}             & \underline{30.14}            \\
Concept-Filter                        & 31.40 & 10.81 & 11.72 & 23.41 & 43.14 & 44.53 & 5.14  & 11.06 & 25.15           & 51.81             & 29.19            \\
CLIPScore                             & 15.48 & 12.75 & 11.47 & 20.99 & 38.23 & 64.15 & 5.25  & 10.56 & 24.16           & 53.39             & \textbf{31.27}            \\
Dot                                   & 42.01 & 8.54  & 12.17 & 16.87 & 36.98 & 45.49 & 4.16  & 23.51 & 25.32           & 48.27             & 26.47            \\
TracIn                                & 41.30 & 10.47 & 12.02 & 16.69 & 35.69 & 51.79 & 4.96  & 26.34 & \underline{26.46} & 47.26             & 26.36            \\
TRAK                                  & 30.29 & 12.37 & 12.17 & 25.93 & 39.04 & 45.95 & 4.46  & 12.88 & 25.19           & 48.24             & 27.17            \\
\tableLineColor \textbf{CHIPS (ours)} & 38.93 & 11.05 & 11.87 & 21.43 & 37.99 & 50.22 & 5.29  & 25.96 & \textbf{27.03}  & 47.88             & 25.71            \\ \midrule
\textit{$r$=20\%}                     &       &       &       &       &       &       &       &       &                 &                   &                  \\
Random                                & 35.79 & 13.82 & 11.62 & 21.92 & 39.35 & 36.41 & 6.60  & 11.13 & 25.00           & \textbf{51.39}             & \underline{29.57}            \\
Concept-Balance                       & 32.77 & 13.62 & 11.97 & 24.00 & 44.24 & 43.16 & 5.42  & 10.37 & 26.20           & \underline{51.38}             & \textbf{30.31}            \\
Concept-Filter                        & 31.36 & 11.70 & 11.97 & 25.46 & 44.98 & 35.85 & 6.42  & 15.21 & 25.23           & 51.25             & 29.54            \\
CLIPScore                             & 17.33 & 9.83  & 11.27 & 18.82 & 38.01 & 32.35 & 4.45  & 11.25 & 20.01           & 51.25             & 28.81            \\
Dot                                   & 39.20 & 11.19 & 11.77 & 13.30 & 35.46 & 53.72 & 5.38  & 33.63 & 26.39           & 47.50             & 26.93            \\
TracIn                                & 42.47 & 12.28 & 11.82 & 19.70 & 37.92 & 46.13 & 5.07  & 19.17 & \underline{26.63} & 46.79             & 25.09            \\
TRAK                                  & 43.25 & 10.12 & 11.02 & 16.75 & 37.66 & 28.83 & 5.41  & 28.16 & 24.54           & 47.54             & 25.81            \\
\tableLineColor \textbf{CHIPS (ours)} & 44.13 & 13.02 & 11.87 & 22.49 & 40.79 & 38.54 & 5.92  & 30.36 & \textbf{28.20}  & 47.45             & 25.90            \\ \midrule
\textit{$r$=30\%}                     &       &       &       &       &       &       &       &       &                 &                   &                  \\
Random                                & 36.77 & 11.90 & 12.07 & 18.42 & 43.35 & 47.08 & 6.42  & 10.94 & \underline{26.28}           & \underline{50.98}             & \textbf{29.85}            \\
Concept-Balance                       & 31.95 & 11.88 & 12.07 & 24.61 & 42.52 & 35.88 & 6.48  & 10.43 & 24.56           & \textbf{51.13}             & \underline{29.66}            \\
Concept-Filter                        & 32.37 & 11.23 & 12.02 & 25.96 & 41.40 & 41.02 & 6.81  & 21.06 & 25.37           & 50.63             & 29.54            \\
CLIPScore                             & 21.93 & 8.25  & 11.32 & 13.94 & 38.81 & 26.49 & 4.49  & 11.94 & 19.01           & 50.76             & 28.09            \\
Dot                                   & 38.42 & 11.05 & 11.62 & 15.40 & 37.52 & 35.04 & 7.69  & 20.74 & 23.97           & 46.39             & 25.67            \\
TracIn                                & 39.63 & 14.18 & 11.97 & 14.24 & 38.91 & 41.09 & 5.69  & 22.19 & 25.68 & 46.23             & 25.15            \\
TRAK                                  & 38.95 & 10.49 & 11.42 & 16.92 & 36.96 & 25.70 & 7.75  & 25.14 & 23.54           & 46.29             & 24.04            \\
\tableLineColor \textbf{CHIPS (ours)} & 41.66 & 13.56 & 11.37 & 25.02 & 43.10 & 46.53 & 7.69  & 36.34 & \textbf{29.96}  & 46.34             & 26.11            \\ \midrule
\textit{$r$=50\%}                     &       &       &       &       &       &       &       &       &                 &                   &                  \\
Random                                & 34.47 & 13.86 & 11.82 & 18.82 & 45.82 & 41.00 & 10.06 & 10.56 & 26.26           & 50.56             &  29.29                \\ \bottomrule
\end{tabular}
\vspace{-1mm}
\caption{Downstream performance of MetaCLIP-B16-400M \cite{metaclip} and variants continually pre-trained with data subsets from BIOMEDICA \cite{biomedica} selected by different methods, under a fixed retention ratio $r \in \{10\%,20\%, 30\%, 50\%\}$. Abbreviations: for medical specialties, OPH = Ophthalmology, RAD = Radiology, DER = Dermatology, HEM = Hematology, PAT = Pathology, NEU = Neuropathology, HIS = Histology, BIO = Non-clinical Biology; for metrics in general domain, \textit{CLS} denotes classification accuracy, and \textit{RET} denotes the average R@1 across image-to-text and text-to-image retrieval tasks. In each setting, the best result is \textbf{bolded} and the second best is \underline{underlined}.
}
\label{tab:main_results}
\vspace{-1.35mm}
\end{table*}

\subsection{Experimental Setup}

\paragraph{Training.}

For a comprehensive comparison, we perform continual pre-training using the initialization weights of MetaCLIP-B32/B16/L14/H14 \cite{metaclip}.
We select \textbf{medical} as target domain and two recently introduced large-scale medical multimodal datasets are employed as training pools: BIOMEDICA (24M samples) \cite{biomedica} and MedTrinity (18M samples) \cite{medtrinity}.
We select data retention ratio $r$ as 10\%, 20\%, 30\%, and 50\% of the complete dataset, with the number of training epochs fixed at 5.
All models are optimized using the AdamW optimizer ($\beta_1=0.9, \beta_2=0.98, \epsilon=10^{-6}$) and a cosine learning rate scheduler (initial learning rate as $10^{-6}$), with a global batch size of 32,768.
Each experiment is conducted on an Ubuntu 24.04 server equipped with 8x NVIDIA H200 (141GB) GPUs.
Further details regarding the training datasets, model architectures, and specific hyperparameter settings are provided in App.~\ref{app:train}.

\paragraph{Evaluation.}

We evaluate 48 tasks spanning general and medical domains to enable a comprehensive assessment. 
The general-domain suite includes Cars \cite{Cars}, Country211 \cite{clip}, Fer2013 \cite{Fer}, Aircraft \cite{Aircraft}, Food101 \cite{Food101}, GTSRB \cite{GTSRB}, ImageNet-A, ImageNet-O \cite{ImageNet-AO}, ImageNet-1K \cite{ImageNet}, ImageNetV2 \cite{ImageNetV2}, MNIST \cite{MNIST}, Rendered-SST2 \cite{clip}, STL-10 \cite{STL10}, SUN397 \cite{SUN}, VOC \cite{VOC2007}, Caltech-101 \cite{Caltech101}, CIFAR-10, CIFAR-100 \cite{CIFAR}, CLEVR (Closest-Object-Distance/Count-All) \cite{CLEVR}, DTD \cite{DTD}, EuroSAT \cite{Helber2019EuroSAT}, Oxford Flowers \cite{Flowers102}, KITTI \cite{Geiger2013KITTI}, Pets \cite{Pets}, RESISC45 \cite{RESISC}, smallNORB (Azimuth), smallNORB (Elevation) \cite{NORB}, SVHN \cite{SVHN}, Flickr8k \cite{flickr8k}, Flickr30k \cite{flickr30k}, and MSCOCO \cite{mscoco}.
For the medical domain, we consider 17 classification tasks across eight specialties. 
These include Ophthalmology (Diabetic \cite{diabetic}, OCTMNIST, RetinaMNIST \cite{medmnistv2}); Radiology (ChestMNIST, ChestX-ray14 \cite{chestx-ray14}, OrganAMNIST, OrganCMNIST, OrganSMNIST); Dermatology (DermaMNIST); Hematology (BloodMNIST); Pathology (PCAM \cite{pcam}, LC25000 \cite{LC25000}, PathMNIST); Neuropathology (Amyloid CAA/Diffuse \cite{amyloid}); Histology (TissueMNIST); and Non-clinical Biology (Pollen \cite{icpr2020_pollen}). 
For classification tasks, the average accuracy is reported.
For retrieval tasks, bidirectional (image-to-text and text-to-image) performance is evaluated using $R@1$ recall metric.
Further details are provided in App.~\ref{app:eval}.

\paragraph{Baselines.}

We select a total of seven data selection baselines for comparison, including (1) \textbf{random selection}, which selects samples uniformly from the training pool without any selective control; (2) three \textbf{heuristics-based methods}: CLIPScore \cite{clipscore}, Concept-Balance, and Concept-Filter \cite{biomedica}; (3) \textbf{Influence-based methods}: Dot \cite{diq}, TracIn \cite{tracin}, and TRAK \cite{trak}.
Moreover, we also include three commonly used medical-specific CLIP models for reference: PubMedCLIP \cite{pubmedclip}, BioMedCLIP \cite{biomedclip}, and BMCLIP \cite{biomedica}.
Implementation details of all baseline methods and our CHIPS method are provided in App.~\ref{app:data_selection}.

\begin{table}[t]
\centering
\footnotesize
\resizebox{\columnwidth}{!}{
\begin{tabular}{lccc}
\toprule
\multicolumn{1}{c}{\textbf{Model}} & \textbf{Medical CLS} & \textbf{General CLS} & \textbf{General RET} \\ \midrule
\textit{B32-400M}                     &                &                &                \\
Random                                & 27.15          & \textbf{49.31} & \textbf{27.33} \\
TracIn                                & \underline{27.48}          & 47.19          & 25.10          \\
\tableLineColor \textbf{CHIPS (ours)} & \textbf{27.83} & \underline{47.90}          & \underline{25.65}          \\ \midrule
\textit{B32-CC}                       &                &                &                \\
Random                                & \underline{27.95}          & \textbf{50.69} & \textbf{29.28} \\
TracIn                                & 26.25          & 48.90          & 25.72          \\
\tableLineColor \textbf{CHIPS (ours)} & \textbf{28.13} & \underline{49.48}          & \underline{26.79}          \\ \midrule
\textit{B16-400M}                     &                &                &                \\
Random                                & 24.78          & \textbf{52.21} & \textbf{29.28} \\
TracIn                                & \underline{26.46}          & 47.26          & \underline{26.36}          \\
\tableLineColor \textbf{CHIPS (ours)} & \textbf{27.03} & \underline{47.88}          & 25.71          \\ \midrule
\textit{B16-CC}                       &                &                &                \\
Random                                & \underline{26.30}          & \textbf{55.91}          & \textbf{30.16} \\
TracIn                                & 25.89          & 50.17          & 27.37          \\
\tableLineColor \textbf{CHIPS (ours)} & \textbf{26.93} & \underline{51.27}          & \underline{28.17}          \\ \midrule
\textit{L14-400M}                     &                &                &                \\
Random                                & \underline{29.33}          & \textbf{57.07} & \textbf{33.35} \\
TracIn                                & 27.08          & 52.99          & \underline{28.17}          \\
\tableLineColor \textbf{CHIPS (ours)} & \textbf{29.73} & \underline{53.65}          & \underline{28.17}          \\ \midrule
\textit{L14-CC}                       &                &                &                \\
Random                                & 30.75          & \textbf{60.74} & \textbf{32.34} \\
TracIn                                & \underline{31.54}          & 56.98          & 28.26          \\
\tableLineColor \textbf{CHIPS (ours)} & \textbf{31.74} & \underline{57.93}          & \underline{30.05}          \\ \midrule
\textit{H14-CC}                       &                &                &                \\
Random                                & \underline{35.23}          & \textbf{61.36} & \textbf{32.82} \\
TracIn                                & 35.10          & \underline{58.27}          & 31.60          \\
\tableLineColor \textbf{CHIPS (ours)} & \textbf{35.48} & 58.24          & \underline{32.09}          \\ \bottomrule
\end{tabular}
}
\caption{Generalization experiment of MetaCLIP-series adapted models continually pre-trained on 10\% kept data from different data selection methods. \textit{400M} denotes the model was pre-trained on 400M general image-text pairs and \textit{CC} denotes the model was pre-trained on 2.5B general image-text pairs.}
\label{tab:scaling}
\end{table}

\subsection{Main Results}

\paragraph{Effectiveness of CHIPS.}
As shown in Tab.~\ref{tab:main_results}, across all retention ratios, CHIPS achieves the best average performance on 17 medical tasks among data selection methods, with Medical Avg scores of 27.03, 28.20, and 29.96, exceeding the second-best method by +0.57, +1.57, and +3.68 points, respectively. 
Notably, with only 10\% of the full BIOMEDICA, CHIPS outperforms a 50\% random subset (27.03 vs. 26.26); with 30\%, it achieves \textbf{95.1\%} of the full-dataset performance (29.96 vs. 31.51).
It is also competitive with specialized medical CLIP models pre-trained on larger datasets: at $r=30\%$, CHIPS slightly surpasses BMCLIP (29.96 vs. 29.86) and consistently outperforms PubMedCLIP and BioMedCLIP across $r$. 
Even though CHIPS is not uniformly best on every specialty, it yields the highest overall average. 
Importantly, CHIPS better preserves general-domain abilities than the previous SOTA selection method TracIn: for \textit{CLS}, CHIPS retains 90.1\%, 89.3\%, and 87.2\% of the vanilla model at $r=10\%,20\%,30\%$ (vs. 88.9\%, 88.0\%, 87.0\% for TracIn), and for \textit{RET} it is comparable, slightly lower at $r=10\%$ but higher at $r\ge20\%$, yielding a higher average RET across retention ratios. 
Overall, CHIPS delivers the strongest medical performance while incurring \textit{less} degradation on the general domain than TracIn.


\paragraph{Generalization abilities of CHIPS.}
With a single scoring pass on MetaCLIP-B16-400M (10\% kept), we reuse the computed CHIPS scores to train diverse backbones (B32/B16/L14/H14) and pre-training scales (\textit{400M} and \textit{CC}). 
As shown in Tab.~\ref{tab:scaling}, across all seven settings, CHIPS attains the best Medical performance, outperforming TracIn by 0.20-2.65 points. 
On general-domain metrics, CHIPS typically ranks second (behind Random) and generally matches or exceeds TracIn on CLS and RET, with a single RET exception on B16-400M. 
Thus, CHIPS transfers across architectures and scales, allowing cached scores to be reused while delivering strong medical gains with minimal loss of general ability.

\vspace{-3mm}
\paragraph{Ablation study.}
As shown in Tab.~\ref{tab:ablation}, we ablate CHIPS by progressively adding components from Eq.~\eqref{eq:final_utility}. 
On the medical benchmarks, CHIPS achieves the highest performance at all budgets, surpassing the strongest ablation by +1.05, +0.28, and +1.46 points at $r=10\%,20\%,30\%$, respectively. 
These gains support the multiplicative combination of alignment, learnability, and domain relevance, with the relevance term becoming especially beneficial at larger budgets ($r=30\%$). 
Meanwhile, general-domain performance remains close to the best ablation (within $\leq$0.53 for General CLS and $\leq$0.99 for General RET), indicating minimal trade-off for the medical improvements.
On the general domain, the retrieval gap to the best ablation narrows as $r$ grows (0.99 $\rightarrow$ 0.66 $\rightarrow$ 0.37 for General RET), suggesting controlled specialization towards the target domain rather than catastrophic forgetting.

\vspace{-1em}

\paragraph{Cost analysis.}

We measure scoring cost as total FLOPs over $\mathcal{D}_\text{pool}$. 
CHIPS requires $50.9475{\times}10^{15}$ FLOPs, 3.1\% lower than TracIn ($52.5891{\times}10^{15}$) and effectively on par with TRAK ($50.9458{\times}10^{15}$). 
Despite equal-or-lower cost, CHIPS yields consistently stronger medical performance than TracIn (Tab.~\ref{tab:main_results}), improving by +0.57, +1.57, and +4.28 points across three retention settings, respectively. 
In practice, the one-time scoring overhead is further amortized because CHIPS scores can be cached and reused across architectures and pre-training scales (Tab.~\ref{tab:scaling}). 
The full FLOPs computation process is provided in App.~\ref{app:computation}.

\begin{table}[t]
\centering
\footnotesize
\begin{tabular}{lccc}
\toprule
\multicolumn{1}{c}{\textbf{Model}} & \textbf{Medical CLS} & \textbf{General CLS} & \textbf{General RET} \\ \midrule
\textit{r=10\%}       &                &                &                \\
Alignment-only        & 25.98          & 48.33          & \textbf{26.70} \\
Alignment-Margin      & 25.95          & \textbf{48.41} & 26.25          \\
\tableLineColor
\textbf{CHIPS (ours)} & \textbf{27.03} & 47.88          & 25.71          \\ \midrule
\textit{r=20\%}       &                &                &                \\
Alignment-only        & 27.52          & 46.67          & 26.25          \\
Alignment-Margin      & 27.92          & \textbf{46.88} & \textbf{26.56} \\
\tableLineColor
\textbf{CHIPS (ours)} & \textbf{28.20} & 46.73          & 25.90          \\ \midrule
\textit{r=30\%}       &                &                &                \\
Alignment-only        & 27.84          & \textbf{46.78} & 25.23          \\
Alignment-Margin      & 28.50          & 46.75          & \textbf{25.52} \\
\tableLineColor
\textbf{CHIPS (ours)} & \textbf{29.96} & 46.34          & 25.15          \\ \bottomrule
\end{tabular}
\caption{Ablation experiment of CHIPS on MetaCLIP-B16-400M under 10\%, 20\%, and 30\% data retention ratios. \textit{Alignment-only} uses single alignment score, \textit{Alignment-Margin} additionally introduces the margin term, and CHIPS multiplies all three metrics.}
\label{tab:ablation}
\end{table}
\section{Analysis}
\label{sec:analysis}

\subsection{Alignment}

\begin{figure*}[tb]
    \centering
    \includegraphics[width=0.85\linewidth]{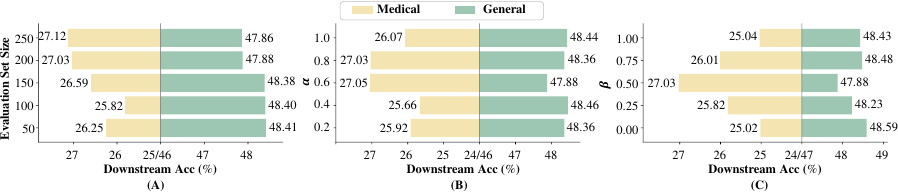}
    \caption{Downstream results of MetaCLIP-B16-400M continually pre-trained on 10\% data selected from CHIPS under: \textbf{(A)} different evaluation set sizes, \textbf{(B)} different mixing  $\alpha$ in computing alignment scores, and \textbf{(C)} different balance $\beta$ in computing relevance scores.}
    \label{fig:eval_alpha_beta_ablation}
\end{figure*}

\paragraph{Effect of target evaluation set size.}
Expanding the target evaluation set $\mathcal{D}_\text{eval}$ sharpens the alignment direction $\mathbf{u}_\vartheta$ and improves medical accuracy with minimal general-domain drift. 
As shown in Fig.~\ref{fig:eval_alpha_beta_ablation}(A), as samples per task increase from 50 to 200, Medical CLS improves from 26.25 to 27.03 (+0.78), then saturates at 250 (27.12, +0.09). 
The slight dip at 100 (25.82) reflects higher variance at smaller set sizes, while overall the trend is upward and stabilizing around 200. 
General CLS remains essentially flat across all sizes (48.41 at 50 to 47.88 at 200; total range $\le$0.55), indicating that larger $\mathcal{D}_\text{eval}$ chiefly benefits medical alignment without harming general performance. 
Balancing diminishing returns beyond 200 against the linear increase in scoring cost, and we adopt 200 samples per task as the default.

\begin{figure}[t]
    \centering
    \includegraphics[width=0.8\linewidth]{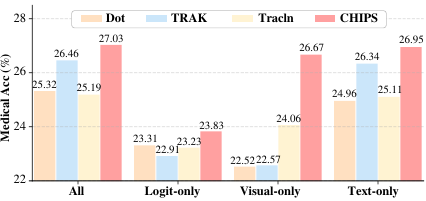}
    \caption{Medical downstream results of MetaCLIP-B16-400M continually pre-trained on 10\% data selected from different selection methods under various end-point geometry settings.}
    \label{fig:projection_ablation}
\end{figure}

\vspace{-1em}

\paragraph{Projection layers dominate alignment with text being more informative.}
\label{sec:geometry}

We ablate the end-point geometry $\vartheta=\{\mathbf W_v,\mathbf W_t,\tau\}$ into \textit{All}, \textit{Logit-only}, \textit{Visual-only}, and \textit{Text-only}, with results shown in Fig.~\ref{fig:projection_ablation}.
Across methods, \textit{Text-only} nearly matches \textit{All} (98.6--99.7\%), \textit{Visual-only} lags, and \textit{Logit-only} is weakest. 
For CHIPS, the scores are 27.03 (\textit{All}), 26.95 (\textit{Text-only}, -0.08; 99.7\%), 26.67 (\textit{Visual-only}, -0.36; 98.7\%), and 23.83 (\textit{Logit-only}, -3.20). 
These results show that alignment resides primarily in the projection heads, especially $\mathbf W_t$, with $\mathbf W_v$ providing a smaller yet complementary gain. 
Moreover, CHIPS is more robust to projection layers compared with baseline methods, retaining 99.7\% (\textit{Text-only}) and 98.7\% (\textit{Visual-only}), the highest among methods, and exhibiting the smallest text--visual gap (0.28 vs.\ 2.43/3.77/1.05 for Dot/TracIn/TRAK). 

\vspace{-1em}

\paragraph{Effect of random projection.}
\label{sec:rp}

\begin{figure}[t]
    \centering
    \includegraphics[width=0.8\linewidth]{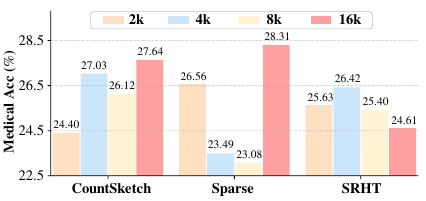}
    \caption{Medical downstream results of MetaCLIP-B16-400M continually pre-trained on 10\% data selected from CHIPS under various random projection settings.}
    \label{fig:random_projection_ablation}
\end{figure}

We compare three random projection methods in computing the alignment score $\widehat A_\alpha(z)$ and vary the target dimension $k \in \{2048,4096,8192,16384\}$, with results shown in Fig.~\ref{fig:random_projection_ablation}. 
\textit{Sparse} projections are the most effective, peaking at \textit{28.31} (16k) and remaining strong even at 2k (26.56). 
\textit{CountSketch} improves with dimension (24.40$\rightarrow$27.64 from 2k to 16k), with a small dip at 8k (26.12); at 4k it already reaches 27.03. 
\textit{SRHT} is less robust, peaking at 4k (26.42) and degrading at 16k (24.61). 
Overall, \textit{Sparse}-16k is best (+0.67 over \textit{CountSketch}-16k; +1.89 over \textit{SRHT}-4k). 
For tighter budgets, \textit{CountSketch}-4k offers a cost-efficient alternative within about 1.3 points of the best.

\vspace{-1em}

\paragraph{Balancing positive and negative curvature.}
\label{sec:alpha}

In Eq.~\eqref{eq:h_m_def}, $\alpha$ linearly mixes positive and negative pair curvature to form $\mathbf M$, which reweights gradients in $A(z)=\mathbf g_\vartheta(z)^\top\mathbf M^{-1}\mathbf u_\vartheta$. 
As shown in Fig.~\ref{fig:eval_alpha_beta_ablation}(B), Medical CLS peaks at $\alpha{=}0.6$ (27.05) and remains essentially tied at $0.8$ (27.03), while both smaller ($0.2$: 25.92; $0.4$: 25.66) and larger ($1.0$: 26.07) values underperform by about 1-1.4 points. 
General CLS is almost flat across $\alpha$ (48.36-48.46), with a modest dip at $0.6$ (47.88), indicating that $\alpha$ mainly governs target-domain discrimination with negligible general-domain impact. 
Mechanistically, moderate $\alpha$ balances attraction to positives and repulsion from negatives, improving the conditioning of $\mathbf M$ and yielding a more discriminative alignment direction; extremes overweight one term and reduce $A(z)$’s utility. 
We therefore recommend $\alpha\in[0.6,0.8]$ (default $0.6$ for medical focus; $0.4$ if slightly prioritizing general-domain retention).

\subsection{Domain Relevance and General Retention}
\label{sec:beta}
In Eq.~\eqref{eq:wR}, $\beta$ balances visual and textual similarities when computing $w_{\mathrm R}(z)$. 
As shown in Fig.~\ref{fig:eval_alpha_beta_ablation}(C), varying $\beta$ reveals a clear adaptation-retention trade-off: Medical CLS peaks at $\beta{=}0.5$ (27.03), improving by 2.01-1.21 points over the unimodal extremes ($\beta{=}0$: 25.02; $\beta{=}1$: 25.04). 
In contrast, General CLS is flattish (48.59$\rightarrow$47.88 across settings; total spread $0.71$), with the best retention at the extremes ($\beta{=}0$: 48.59; $\beta{=}1$: 48.43) and the largest dip at $\beta{=}0.5$ (47.88). 
This U-shaped retention pattern indicates that balanced relevance sharpens specialization to the medical target, whereas unimodal weighting selects more generic samples that better preserve general-domain ability. 
A slight skew toward text ($\beta{=}0.75$) offers a good compromis: 26.01 on Medical (within 1.02 of the best) with near-top 48.48 on General, which is consistent with our end-point geometry ablation where text signals are more informative (Sec.~\ref{sec:geometry}). 
Finally, the overall narrow range in General CLS ($\le$0.71) aligns with the bounded-drift property of $w_{\mathrm R}$, which limits deviation from the base alignment distribution. 
We adopt $\beta{=}0.5$ for maximal target-domain gains.
\section{Conclusion}
We propose CHIPS, showing that principled data selection can substitute for raw scale in CLIP continual pre-training. 
On 17 medical benchmarks, CHIPS attains SOTA performance among selection baselines, matches full-dataset CPT with 30\% of the data, and outperforms half-dataset CPT using only 10\%. 
Across 31 general-domain benchmarks, it yields the smallest performance drop under 10--30\% retention ratios, better preserving general capabilities. 
Limitations include reliance on a target validation distribution, we will explore unlabeled or shift-robust target signals and extensions beyond the CLIP model in the future work.

{
    \small
    \bibliographystyle{ieeenat_fullname}
    \bibliography{main}
}

\clearpage
\appendix
\setcounter{page}{1}
\maketitlesupplementary

\startcontents[appendixtoc]
\printcontents[appendixtoc]{l}{1}{
    \setcounter{tocdepth}{2}
    \section*{Appendix Contents}
}

\section{Related Work}
\label{sec:related_work}

\paragraph{CLIP Adaptation.}

Current approaches for effective CLIP adaptation to specialized vertical domains (e.g., Medical) can be broadly categorized into two paradigms: model-centric and data-centric methods \cite{medical_clip_survey}.
Model-centric approaches primarily focus on developing novel training strategies, including probabilistic fine-tuning \cite{clap4clip} and many-to-many contrastive learning \cite{cplip}, as well as parameter-efficient fine-tuning (PEFT) techniques \cite{SPT, BSR, clip_ast, clip_refine}.
In contrast, data-centric approaches emphasize the collection of large-scale domain-specific datasets for continual pre-training.
Within the general-purpose medical domain, PMC-CLIP \cite{pmc-clip} leveraged approximately 2M samples, BioMedCLIP \cite{biomedclip} curated 15M samples, BIOMEDICA \cite{biomedica} assembled 24M samples, and MedTrinity aggregated 25M samples.
For more specialized medical subdomains, QUILT \cite{quilt1m} compiled 1M samples for Histopathology, EYECLIP \cite{ophthalmology} gathered approximately 3M samples for Ophthalmology, and PanDerm \cite{dermatology} curated 2M samples for Dermatology.
In the biology and biodiversity domains, even larger-scale datasets ranging from 5M to 214M samples have been collected \cite{bioclip, bioclip2, biotrove, bioscan, dalip, openinsects}.
In this work, rather than pursuing resource-intensive dataset scaling, we investigate the intrinsic characteristics of existing domain-specific datasets and employ strategic data selection to achieve more efficient CLIP adaptation.

\paragraph{Data Attribution.}

Data attribution methods quantify the influence of individual training samples on model training efficacy. 
\citep{if} initially introduced Influence Functions (IF), proposing the LiSSA approximation to efficiently compute the inverse Hessian-vector product, wherein the computation entails calculating the Hessian inverse, iterating through validation and training samples, and deriving influence scores through gradient-based calculations. 
\cite{tracin} proposed TracIn, which offers a more intuitive first-order approximation that tracks model training dynamics across multiple checkpoints. 
\cite{EL2N} introduced EL2N, a self-influence metric that computes the L2 norm of prediction error vectors without requiring explicit validation samples. 
\cite{arnoldi} proposed the Arnoldi iteration method, which leverages dominant eigenvalue decomposition to address the computational overhead and non-H-invariance issues inherent in random projection approaches. 
\cite{gex} introduced GEX, reformulating IF to capture bimodal influence distributions by replacing gradient-based Taylor approximations with direct loss multiplications and employing ensemble models to mitigate Hessian singularity bias. 
\cite{fvm} proposed FVM, which optimizes for flat validation minima to enhance the stability of influence estimations for mislabeled sample detection. 
In this work, we propose an IF variant specifically tailored for CLIP influence estimation.

\section{Proofs} 

This section provides consolidated derivations and proofs for the three core components referenced in Sec.~\ref{sec:setup}, Sec.~\ref{sec:alignment}, and Sec.~\ref{sec:neg_curv}. 
Unless stated otherwise, expectations are taken over the mini-batch sampling and data randomness at the current parameter iterate \(\theta\). We write $\widehat{\mathbf g}=\tfrac{1}{B}\sum_{i=1}^B \nabla_\theta \ell(z_i;\theta)$, $\mathbf g_q=\mathbb E_{z\sim q}[\nabla_\theta \ell(z;\theta)]$, $\mathbf u=\nabla_\theta \mathcal L_{\text{eval}}(\theta)$, and use the Euclidean inner product and norm. Spectral and Frobenius norms are denoted by \(\|\cdot\|_2\) and \(\|\cdot\|_F\).

\subsection{Alignment Direction}
\label{app:proofs_alignment_direction}

In a neighborhood of \(\theta\) assume either (i) \(\mathcal L_{\text{eval}}\in C^2\) with Hessian-Lipschitz constant \(L_H\), or (ii) \(\mathcal L_{\text{eval}}\) is \(\rho\)-smooth.

\paragraph{Second-order upper bound.}
For \(\Delta\theta\), the second-order Taylor expansion with integral remainder gives
\begin{align}
\label{eq:full_expansion}
\mathcal L_{\text{eval}}(\theta+\Delta\theta)
&= \mathcal L_{\text{eval}}(\theta)+\mathbf u^\top\Delta\theta \nonumber \\
&\quad +\tfrac12\,\Delta\theta^\top \mathbf H_{\text{eval}}(\Theta)\,\Delta\theta
+R_3,
\end{align}
where \(\Theta\) lies on the segment between \(\theta\) and \(\theta+\Delta\theta\), and
\(|R_3|\le \tfrac{L_H}{6}\,\|\Delta\theta\|^3\).
With the one-step SGD move \(\Delta\theta=-\eta\,\widehat{\mathbf g}\) and taking expectation,
\begin{align*}
\mathbb E\big[\Delta\mathcal L_{\text{eval}}\big]
&\le -\eta\,\mathbf g_q^\top\mathbf u \\
&\quad +\tfrac12\,\eta^2\,\mathbb E\!\big[\widehat{\mathbf g}^{\!\top}\mathbf H_{\text{eval}}(\Theta)\widehat{\mathbf g}\big] \\
&\quad +\tfrac{L_H}{6}\,\eta^3\,\mathbb E\!\big[\|\widehat{\mathbf g}\|^3\big],
\end{align*}
which matches Eq.~\eqref{eq:delta_eval_second} in the main text and shows that descent is driven by the alignment term \(-\mathbf g_q^\top \mathbf u\).

\paragraph{First-order upper bound.}
If \(\mathcal L_{\text{eval}}\) is \(\rho\)-smooth, the Descent Lemma yields
\begin{equation}
\label{eq:descent_lemma}
\mathcal L_{\text{eval}}(\theta+\Delta\theta)
\le
\mathcal L_{\text{eval}}(\theta)+\mathbf u^\top\Delta\theta+\tfrac{\rho}{2}\|\Delta\theta\|^2.
\end{equation}
Setting \(\Delta\theta=-\eta\,\widehat{\mathbf g}\) and taking expectations gives Eq.~\eqref{eq:delta_eval_first}.
\par
Reegarding mini-batch variance, let \(\Sigma_q=\operatorname{Cov}_{z\sim q}[\nabla_\theta \ell(z;\theta)]\), then
\begin{equation}
\label{eq:batch_moments}
\begin{aligned}
\mathbb E\big[\|\widehat{\mathbf g}\|^2\big]
&= \|\mathbf g_q\|^2+\tfrac{1}{B}\operatorname{tr}(\Sigma_q), \\
\mathbb E\big[\|\widehat{\mathbf g}-\mathbf g_q\|^2\big]
&= \tfrac{1}{B}\operatorname{tr}(\Sigma_q).
\end{aligned}
\end{equation}
This makes explicit how the batch size \(B\) moderates the quadratic term in Eq.~\eqref{eq:delta_eval_first}.
In a local quadratic model the steepest descent direction for \(\mathcal L_{\text{eval}}\) is \(\mathbf H_{\text{eval}}^{-1}\mathbf u\). Hence a selection distribution \(q\) that increases \(\mathbf g_q^\top \mathbf H_{\text{eval}}^{-1}\mathbf u\) is desirable, which motivates the proxy alignment in Sec.~\ref{sec:alignment}.

\paragraph{AdamW-aware form.}
At iteration \(t\), let \(\mathbf g_t=\tfrac{1}{B}\sum_{i=1}^B \nabla_\theta \ell(z_i;\theta_t)\). 
Consider Adam moments and bias corrections as the following forms  
\begin{align*}
m_t &= \beta_1 m_{t-1}+(1-\beta_1)\mathbf g_t, \\
v_t &= \beta_2 v_{t-1}+(1-\beta_2)\,(\mathbf g_t\odot \mathbf g_t), \\[0.5em]
\hat m_t &= \frac{m_t}{1-\beta_1^t}, \\
\hat v_t &= \frac{v_t}{1-\beta_2^t}, \\
\mathbf P_t &= \operatorname{diag}\big((\sqrt{\hat v_t}+\epsilon)^{-1}\big),
\end{align*}
where \(\odot\) denotes the element-wise product. 
With decoupled weight decay \(w_d>0\) and diagonal mask \(\mathbf D\), the update is
\[
\Delta\theta_t
\triangleq \theta_{t+1}-\theta_t
=-\eta_t\big(\mathbf P_t \hat m_t+w_d\,\mathbf D\,\theta_t\big).
\]
Under the \(C^2\) and Hessian-Lipschitz assumption, Eq.~\eqref{eq:full_expansion} gives
\begin{align*}
\mathbb E\big[\Delta \mathcal L_{\text{eval}}\big]
\le\,&
-\eta_t\,\mathbb E\big[\hat m_t^\top \mathbf P_t \mathbf u\big]
-\eta_t\,w_d\,(\mathbf D\theta_t)^\top \mathbf u \\
&+\tfrac12\,\eta_t^2\,\mathbb E\!\big[\Delta\theta_t^\top \mathbf H_{\text{eval}}(\Theta_t)\Delta\theta_t\big] \\
&+\tfrac{L_H}{6}\,\eta_t^3\,\mathbb E\!\big[\|\Delta\theta_t\|^3\big],
\end{align*}
Moreover, if under \(\rho\)-smoothness, Eq.~\eqref{eq:descent_lemma} yields
\begin{align*}
\mathbb E\big[\Delta \mathcal L_{\text{eval}}\big]
&\le -\eta_t\,\mathbb E\big[(\mathbf P_t\hat m_t)^\top \mathbf u\big] \\
&\quad -\eta_t\,w_d\,(\mathbf D\theta_t)^\top \mathbf u \\
&\quad +\tfrac{\rho}{2}\,\eta_t^2\,\mathbb E\!\big[\|\Delta\theta_t\|^2\big].
\end{align*}
In both cases the alignment term becomes \(\mathbb E[(\mathbf P_t\hat m_t)^\top \mathbf u]\), which is in line with SDG in the main content of this paper.

\subsection{Proxy Alignment in Subspace}
\label{app:proxy_influence}

In this section, we prove Theorem~\ref{thm:proxy_general} and state two practical corollaries.

In the end-point subspace let \(\mathbf g=\mathbf g_\vartheta(z)\), \(\boldsymbol\mu=\mathbb E[\mathbf g]\), \(\mathbf g_c=\mathbf g-\boldsymbol\mu\), and \(\Sigma_g=\operatorname{Cov}[\mathbf g]\succeq 0\). Locally,
\[
\nabla_\theta \ell(z)=\mathbf J\,\mathbf g + \mathbf r(z),
\qquad
\mathbf u=\bar{\mathbf J}\,\mathbf u_\vartheta+\boldsymbol\varepsilon,
\]
with \(\mathbf J,\bar{\mathbf J}\) linear maps, residual \(\mathbf r(z)\), and deterministic mismatch \(\boldsymbol\varepsilon\). Define
\[
\mathbf S=\tfrac12(\mathbf J^\top\bar{\mathbf J}+\bar{\mathbf J}^\top \mathbf J),\quad
\mathbf A=\tfrac12(\mathbf J^\top\bar{\mathbf J}-\bar{\mathbf J}^\top \mathbf J),
\]
and consider
\[
X(z)=\mathbf u_\vartheta^\top \mathbf g_c,\qquad
Y(z)=(\nabla_\theta \ell(z))^\top \mathbf u.
\]
Collect the remaining random terms into
\[
\zeta(z)
=(\mathbf J\mathbf g_c)^\top \boldsymbol\varepsilon
+ \mathbf r(z)^\top \bar{\mathbf J}\mathbf u_\vartheta
+ \mathbf r(z)^\top \boldsymbol\varepsilon
+ \mathbf g_c^\top \mathbf A\,\mathbf u_\vartheta,
\]
and write \(\sigma_\zeta^2=\operatorname{Var}[\zeta(z)]\).

\paragraph{Proof of Theorem~\ref{thm:proxy_general}.}
Let \(\tilde{\mathbf g}=\Sigma_g^{-1/2}\mathbf g_c\), \(\alpha=\Sigma_g^{1/2}\mathbf u_\vartheta\), \(B=\Sigma_g^{1/2}\mathbf S\Sigma_g^{-1/2}\), and \(Z(z)=\mathbf g_c^\top \mathbf S \mathbf u_\vartheta=\alpha^\top B \tilde{\mathbf g}\). If \(\zeta\) is uncorrelated with \(\tilde{\mathbf g}\) then
\(\operatorname{Cov}(X,Y)=\operatorname{Cov}(X,Z)\) and
\(\operatorname{Var}(Y)=\operatorname{Var}(Z)+\sigma_\zeta^2\).
Using
\begin{align*}
\operatorname{Var}(X) &= \|\alpha\|_2^2, \\
\operatorname{Cov}(X,Z) &= \alpha^\top \mathrm{sym}(B)\,\alpha, \\
\operatorname{Var}(Z) &\le \|B\|_2^2\,\|\alpha\|_2^2.
\end{align*}
and the Rayleigh-Ritz principle, gives
\begin{equation}
\label{eq:proxy_lb_final}
\rho_{XY}
\ \ge\
\frac{\lambda_{\min}\!\big(\mathrm{sym}(B)\big)}
{\sqrt{\ \|B\|_2^2+\sigma_\zeta^2/(\mathbf u_\vartheta^\top \Sigma_g \mathbf u_\vartheta)\ }},
\end{equation}
which is Eq.~\eqref{eq:proxy_lb_general} in the main content after identifying \(B\). 
Without the uncorrelatedness assumption, Cauchy-Schwarz implies the fallback bound
\begin{equation}
\label{eq:proxy_lb_fallback}
\rho_{XY}
\ \ge\ 
\frac{\lambda_{\min}(\mathrm{sym}(B))\,\|\alpha\|_2-\sigma_\zeta}
     {\|B\|_2\,\|\alpha\|_2+\sigma_\zeta}.
\end{equation}

\subsection{Negative Pair Curvature}
\label{app:neg_curv_proof}

In the projection-temperature subspace, we write \(\mathbf g(z)=\mathbf g_\vartheta(z)\) and \(\mathbf u=\mathbf u_\vartheta\).
For softmax-type losses, a generalized Gauss-Newton curvature admits the population decomposition
\begin{equation}
\label{eq:app_H_decomp_simpler}
\begin{aligned}
\mathbf H_\vartheta &= \Phi_{\text{pos}}+\Phi_{\text{neg}}, \\
\Phi_{\text{pos}} &= \mathbb E_{z}\!\big[\mathbf g(z)\mathbf g(z)^\top\big], \\
\Phi_{\text{neg}} &= \mathbb E_{z\ne z'}\!\big[\mathbf g(z)\mathbf g(z')^\top\big],
\end{aligned}
\end{equation}
which mirrors the random-negative mechanism of InfoNCE. 
Note that \(\Phi_{\text{pos}}\succeq 0\) while \(\Phi_{\text{neg}}\) need not be PSD, and the sum \(\mathbf H_\vartheta\) is PSD.

For a mini-batch \(\{z_i\}_{i=1}^B\) let \(\mathbf g_i=\mathbf g_\vartheta(z_i)\), then
\begin{equation}
\label{eq:mini_batch_estimators}
\widehat\Phi_{\text{pos}}=\tfrac{1}{B}\sum_{i=1}^B \mathbf g_i\mathbf g_i^\top,\qquad
\widehat\Phi_{\text{neg}}=\tfrac{1}{B(B-1)}\sum_{i\ne j}\mathbf g_i\mathbf g_j^\top.
\end{equation}
As in Eq.~\eqref{eq:h_m_def}, define
\begin{equation}
\label{eq:h_m_def_app}
\mathbf H_\vartheta^{(\alpha)}=(1-\alpha)\,\Phi_{\text{pos}}+\alpha\,\Phi_{\text{neg}},
\qquad
\mathbf M=\mathbf H_\vartheta^{(\alpha)}+\lambda\,\mathbf I,
\end{equation}
with \(\alpha\in[0,1]\) and \(\lambda>0\). Let
\(A^\star(z)=\mathbf g(z)^\top \mathbf H_\vartheta^{-1}\mathbf u\),
\(A_\alpha(z)=\mathbf g(z)^\top \mathbf M^{-1}\mathbf u\),
and \(\Delta_\alpha=\mathbf H_\vartheta-\mathbf H_\vartheta^{(\alpha)}\).
Using the resolvent identity,
\begin{equation}
\label{eq:inv_pert_lemma_simpler}
\begin{aligned}
\mathbf M^{-1}-\mathbf H_\vartheta^{-1}
&= \mathbf H_\vartheta^{-1}\,(\Delta_\alpha-\lambda\mathbf I)\,\mathbf M^{-1}, \\
\big\|\mathbf M^{-1}-\mathbf H_\vartheta^{-1}\big\|_2
&\le \|\mathbf H_\vartheta^{-1}\|_2\,\|\Delta_\alpha\|_2\,\|\mathbf M^{-1}\|_2.
\end{aligned}
\end{equation}
which leads to
\[
\mathbb E_z\big[(A_\alpha-A^\star)^2\big]
\le C_2\,\|\Delta_\alpha\|_F^2\,\|\mathbf H_\vartheta^{-1}\mathbf u\|_2^2,
\]
for a constant \(C_2>0\) depending only on \(\|\Phi_{\text{pos}}\|_2\), \(\|\mathbf H_\vartheta^{-1}\|_2\), and \(\|\mathbf M^{-1}\|_2\). Whenever cross-example coupling is present, any \(\alpha>0\) that injects off-diagonal mass reduces \(\|\Delta_\alpha\|_F\).
Let \(\Pi_k\in\mathbb R^{k\times d_\vartheta}\) be a JL transform. For any fixed \(\mathbf a,\mathbf b\),
\(|\mathbf a^\top\mathbf b-(\Pi_k\mathbf a)^\top(\Pi_k\mathbf b)|\le \varepsilon\|\mathbf a\|\,\|\mathbf b\|\) holds with probability at least \(1-\delta\), where \(k=\Omega(\varepsilon^{-2}\log(1/\delta))\). 
Define \(\mathbf M_k=\Pi_k \,\mathbf M \,\Pi_k^\top\) and \(\bar{\mathbf u}_k=\mathbf M_k^{-1}\Pi_k \mathbf u\). The sketched score
\[
\widehat A_\alpha(z)=(\Pi_k\mathbf g(z))^\top \bar{\mathbf u}_k
\]
satisfies \(\mathbb E_z[(\widehat A_\alpha-A_\alpha)^2]\le C_1 \log(1/\delta)/k\) for a constant \(C_1>0\) depending on \(\|\mathbf M^{-1}\|_2\), \(\|\mathbf u\|_2\), and \(\mathbb E\|\mathbf g(z)\|_2^2\). Combining with the mixing bias via the triangle inequality yields Eq.~\eqref{eq:sketch_var_bias_s33} in the main content:
\begin{equation}
\begin{aligned}
\mathbb E_z\!\big[\big(\widehat A_\alpha(z)-A^\star(z)\big)^2\big]
&\le \underbrace{\frac{C_1\log(1/\delta)}{k}}_{\text{projection variance}}
\\
&\quad+\underbrace{C_2\,\|\Delta_\alpha\|_F^2\,\|\mathbf H_\vartheta^{-1}\mathbf u_\vartheta\|_2^2}_{\text{curvature bias}}.
\end{aligned}
\end{equation}

\section{Data Selection Implementation Details}
\label{app:data_selection}

This section provides all components needed to reproduce CHIPS and baseline methods.

\subsection{Problem Setup}
\label{app:notation}

Let the training pool be \(\mathcal D_{\text{pool}}=\{(x_i,y_i)\}_{i=1}^N\) and the evaluation set be \(\mathcal D_{\text{eval}}=\{(x_m,y_m)\}_{m=1}^M\).
Given a data retention ratio \(r\in(0,1]\), we select \(n=\lfloor r\times|\mathcal D_{\text{pool}}|\rfloor\) samples from \(\mathcal D_{\text{pool}}\) to perform CPT.
CHIPS operates in CLIP’s end-point subspace \(\vartheta=\{\mathbf W_v,\mathbf W_t,\tau\}\) where \(\mathbf W_v\) and \(\mathbf W_t\) are the projection heads for vision and text, and \(\tau>0\) is the logit scale.
For computational efficienciy, backbone encoders are used to produce features and are kept fixed for computing selection utility scores (we only need training dynamics of $\vartheta$ instead of the complete model).

Let \(\mathbf h=\mathrm{CLIP}_{\text{img}}(x)\) and \(\mathbf t=\mathrm{CLIP}_{\text{txt}}(y)\) be backbone features.
Define L2-normalized end-point embeddings
\[
\hat{\mathbf x}=\frac{\mathbf W_v\mathbf h}{\|\mathbf W_v\mathbf h\|},\qquad
\hat{\mathbf y}=\frac{\mathbf W_t\mathbf t}{\|\mathbf W_t\mathbf t\|},
\]
and similarities \(s_{ij}=\tau\,\hat{\mathbf x}_i^\top \hat{\mathbf y}_j\).
For a batch of size \(B\), write \(\mathbf S=[s_{ij}]_{i,j=1}^B\) and define the bidirectional softmax probabilities
\[
p^{i2t}_{ij}=\frac{\exp(s_{ij})}{\sum_{j'}\exp(s_{ij'})},\qquad
p^{t2i}_{ij}=\frac{\exp(s_{ij})}{\sum_{i'}\exp(s_{i'j})}.
\]
The symmetric InfoNCE loss for sample \(i\) is
\[
\ell_i(\vartheta)=\tfrac12\big(\mathrm{CE}(\mathbf S_{i,:},\,i)+\mathrm{CE}(\mathbf S_{:,i},\,i)\big),
\]
where $\mathrm{CE}$ denotes the cross-entropy loss.

Let \(\mathbf g_\vartheta(z)=\nabla_\vartheta \ell(z;\vartheta)\in\mathbb R^{d}\) denote the end-point gradient of a sample, and let
\[
\mathbf u_\vartheta\ \triangleq\ \mathbb E_{z\sim \mathcal D_{\text{eval}}}[\mathbf g_\vartheta(z)]
\]
be the evaluation mean gradient in the same subspace. In practice we maintain \(\mathbf u_\vartheta\) by an exponential moving average over random evaluation minibatches.

Moreover, to reduce computational cost we use a JL map matrix \(\Pi_k\in\mathbb R^{k\times d}\) and work with sketched vectors \(\mathbf g_k=\Pi_k\mathbf g_\vartheta\) and \(\mathbf u_k=\Pi_k\mathbf u_\vartheta\). The inner product distortion satisfies \(|\mathbf a^\top\mathbf b-(\Pi_k\mathbf a)^\top(\Pi_k\mathbf b)|\le \varepsilon\|\mathbf a\|\,\|\mathbf b\|\) with probability at least \(1-\delta\) when \(k=\Omega(\varepsilon^{-2}\log(1/\delta))\), which is also used in App.~\ref{app:neg_curv_proof}.

\subsection{CHIPS}
\label{app:chips_impl}

CHIPS ranks each \(z\in\mathcal D_{\text{pool}}\) by the multiplicative utility
\[
\mathcal I_{\text{CHIPS}}(z)\ =\ \widehat A_\alpha(z)\cdot w_{\mathrm L}(z)\cdot w_{\mathrm R}(z),
\]
then selects the top \(n\) samples for CPT. 
The three factors are implemented as follows and the complete process of CHIPS is provided in Alg.~\ref{alg:chips_selection}.

\paragraph{Curvature-aware proxy alignment.}
We estimate the curvature in \(\vartheta\) by mixing self and cross moments from symmetric InfoNCE, consistent with Sec.~\ref{sec:neg_curv}. Maintain EMA estimators
\[
\Phi_{\mathrm{pos}}=\mathbb E[\mathbf g_\vartheta(z)\mathbf g_\vartheta(z)^\top],\qquad
\Phi_{\mathrm{neg}}=\mathbb E_{z\ne z'}[\mathbf g_\vartheta(z)\mathbf g_\vartheta(z')^\top],
\]
then build
\[
\mathbf M\ =\ \mathbf H_\vartheta^{(\alpha)}+\lambda\mathbf I,\qquad
\mathbf H_\vartheta^{(\alpha)}\ =\ (1-\alpha)\,\Phi_{\mathrm{pos}}+\alpha\,\Phi_{\mathrm{neg}},
\]
with \(\alpha\in[0,1]\) and ridge \(\lambda>0\).
Without sketching, the alignment score is
\[
A_\alpha(z)\ =\ \mathbf g_\vartheta(z)^\top\,\mathbf M^{-1}\,\mathbf u_\vartheta.
\]
With a JL sketch, form \(\mathbf M_k=\Pi_k\mathbf M \Pi_k^\top\) once per update and solve \(\bar{\mathbf u}_k=\mathbf M_k^{-1}\mathbf u_k\). The sketched score is
\[
\widehat A_\alpha(z)\ =\ (\Pi_k\mathbf g_\vartheta(z))^\top \bar{\mathbf u}_k.
\]
This matches Eq.~\eqref{eq:alignment_score_main} and App.~\ref{app:neg_curv_proof}.

\paragraph{Learnability.}
Compute correctness and hardest negative margin using the current batch statistics
\begin{equation}
\begin{aligned}
p_{\mathrm{corr}}(z) & = \tfrac12\big(p^{i2t}_{ii}+p^{t2i}_{ii}\big),\\
m(z)                 & = s_{ii}-\max\!\big\{\max_{j\neq i}s_{ij},\,\max_{i'\neq i}s_{i'i}\big\}.
\end{aligned}
\end{equation}
Define the learnability weight as in Eq.~\eqref{eq:wL}
\[
w_{\mathrm L}(z)=\big(1-p_{\mathrm{corr}}(z)\big)\big(1+\sigma(-m(z))\big),
\]
which upweights near-boundary samples and downweights saturated ones.

\paragraph{Target-domain relevance.}
Let the evaluation centroids be \(\boldsymbol\mu_x=\mathbb E[\hat{\mathbf x}]\) and \(\boldsymbol\mu_y=\mathbb E[\hat{\mathbf y}]\) over \(\mathcal D_{\text{eval}}\).
Following Eq.~\eqref{eq:wR},
\[
w_{\mathrm R}(z)=\sigma\Big((1-\beta)\cos(\hat{\mathbf x},\boldsymbol\mu_x)+\beta\cos(\hat{\mathbf y},\boldsymbol\mu_y)\Big),
\]
which softly reweights toward the evaluation domain while keeping the selection drift bounded.

\begin{algorithm}[t]
\caption{CHIPS Algorithm}
\label{alg:chips_selection}
\begin{algorithmic}[1]
\Require pool \(\mathcal D_{\text{pool}}\), evaluation set \(\mathcal D_{\text{eval}}\), retention \(n\), mix \(\alpha\), ridge \(\lambda\), JL dim \(k\), relevance balance \(\beta\)
\State Estimate \(\mathbf u_\vartheta\leftarrow \mathbb E_{z\sim\mathcal D_{\text{eval}}}[\mathbf g_\vartheta(z)]\) using minibatches; cache \(\boldsymbol\mu_x,\boldsymbol\mu_y\)
\State Initialize EMA moments \(\Phi_{\mathrm{pos}}\leftarrow \mathbf 0\), \(\Phi_{\mathrm{neg}}\leftarrow \mathbf 0\)
\For{each minibatch \(\{z_i\}_{i=1}^B\) sampled from \(\mathcal D_{\text{pool}}\)}
  \State compute \(\mathbf g_i=\mathbf g_\vartheta(z_i)\) for all \(i\)
  \State update \(\Phi_{\mathrm{pos}}\) and \(\Phi_{\mathrm{neg}}\) using the batch \(U\)-statistics and EMA
\EndFor
\State \(\mathbf M\leftarrow (1-\alpha)\Phi_{\mathrm{pos}}+\alpha\Phi_{\mathrm{neg}}+\lambda\mathbf I\)
\If{\(k>0\)} \State sample JL \(\Pi_k\), set \(\mathbf M_k=\Pi_k\mathbf M\Pi_k^\top\), solve \(\bar{\mathbf u}_k=\mathbf M_k^{-1}(\Pi_k\mathbf u_\vartheta)\)
\EndIf
\For{each \(z=(x,y)\in\mathcal D_{\text{pool}}\)}
  \State compute \(\mathbf g=\mathbf g_\vartheta(z)\) and \(\widehat A_\alpha(z)=
  \begin{cases}
  \mathbf g^\top \mathbf M^{-1}\mathbf u_\vartheta,& k=0\\
  (\Pi_k\mathbf g)^\top \bar{\mathbf u}_k,& k>0
  \end{cases}\)
  \State compute \(p_{\mathrm{corr}}(z)\) and \(m(z)\), then \(w_{\mathrm L}(z)=\big(1-p_{\mathrm{corr}}(z)\big)\big(1+\sigma(-m(z))\big)\)
  \State compute \(w_{\mathrm R}(z)=\sigma\big((1-\beta)\cos(\hat{\mathbf x},\boldsymbol\mu_x)+\beta\cos(\hat{\mathbf y},\boldsymbol\mu_y)\big)\)
  \State set \(\mathcal I_{\text{CHIPS}}(z)=\widehat A_\alpha(z)\,w_{\mathrm L}(z)\,w_{\mathrm R}(z)\)
\EndFor
\State return the top \(n\) samples by \(\mathcal I_{\text{CHIPS}}\)
\end{algorithmic}
\end{algorithm}

\paragraph{More details.}

During the computation process,
\begin{itemize}
    \item \(\mathbf M^{-1}\mathbf u_\vartheta\) is reused across all samples, so the per-sample cost is dominated by \(\mathbf g_\vartheta(z)\) and a dot product in \(d\) or \(k\) dimensions.  
    \item \(\tau\) is kept positive by parameterizing \(\tau=\exp(\tilde\tau)\).
    \item We rescale \(\mathcal I_{\text{CHIPS}}\) within each shard of \(\mathcal D_{\text{pool}}\) if memory constraints require sharded processing. 
    \item We recompute \(\mathbf u_\vartheta\) periodically to track drift during CPT.
\end{itemize}

\subsection{Heuristics-based Baselines}
\label{app:heuristics}

All heuristics select the top \(n\) samples under their scores.

\begin{itemize}
\item \textbf{Random} samples uniformly at random without replacement.
\item \textbf{CLIPScore} ranks by CLIPScore \cite{clipscore} computed with the frozen base CLIP model.
\item \textbf{Concept-Balance} performs probabilistic downsampling to flatten concept frequency. Following \cite{biomedica}, we downsample overrepresented concepts such as \textit{Plots and Charts}, \textit{Tables}, and \textit{Scientific Formulae and Equations} at rate \(0.25\), then sample the remainder uniformly.
\item \textbf{Concept-Filter} keeps only samples whose metadata contain any of the eight whitelist concepts in \cite{biomedica}, (\textit{Clinical Imaging}, \textit{Microscopy}, \textit{Immuno Assays}, \textit{Illustrative Diagrams}, \textit{Chemical Structures}, \textit{Maps}, \textit{Tools and Materials}, and \textit{Hand Drawn and Screen Based Visuals}) then samples uniformly from the filtered pool.
\end{itemize}

\subsection{Influence-based Baselines in the End-point Subspace}
\label{app:influence}

For fairness and to avoid conflicts with the main text, all influence baselines use the same end-point gradients \(\mathbf g_\vartheta\) and the same optional JL sketch \(\Pi_k\) as CHIPS. Let \(\tilde{\mathbf g}_i=\Pi_k\mathbf g_\vartheta(z_i)\) if \(k>0\) (use \(\tilde{\mathbf g}_i=\mathbf g_\vartheta(z_i)\) when \(k=0\)). Let the evaluation mean gradient be \(\bar{\mathbf g}_{\text{eval}}=\frac{1}{M}\sum_{m=1}^M \tilde{\mathbf g}^{\,\text{eval}}_m\).

\paragraph{Dot \cite{diq}.}
The first-order directional alignment is
\[
\mathcal I_{\text{Dot}}(i)=\tilde{\mathbf g}_i^\top \bar{\mathbf g}_{\text{eval}}.
\]

\paragraph{TracIn \cite{tracin}.}
Given checkpoints \(\{\theta_t\}_{t=0}^{T-1}\) and learning rates \(\{\eta_t\}\), we accumulate
\[
\mathcal I_{\text{TracIn}}(i)=\sum_{t=0}^{T-1}\eta_t\,\big(\tilde{\mathbf g}_i^{\,(t)}\big)^\top \bar{\mathbf g}_{\text{eval}},
\]
where \(\tilde{\mathbf g}_i^{\,(t)}=\Pi_k\nabla_\vartheta \ell_i(\theta_t)\). We keep \(\bar{\mathbf g}_{\text{eval}}\) fixed for computational stability.

\paragraph{TRAK \cite{trak}.}
We form a regularized second-moment matrix in the same feature space
\[
\boldsymbol\Phi=\frac{1}{N}\sum_{i=1}^N \tilde{\mathbf g}_i\tilde{\mathbf g}_i^\top+\lambda_{\text{TRAK}}\mathbf I,
\]
and compute
\[
\mathcal I_{\text{TRAK}}(i)=\tilde{\mathbf g}_i^\top \boldsymbol\Phi^{-1}\bar{\mathbf g}_{\text{eval}}.
\]

\paragraph{CHIPS for comparison.}
CHIPS differs from Dot and TRAK by preconditioning with \(\mathbf M\) that mixes \(\Phi_{\text{pos}}\) and \(\Phi_{\text{neg}}\) as in Sec.~\ref{app:chips_impl}, and by multiplying learnability and relevance weights. In the sketched space,
\[
\mathcal I_{\text{CHIPS}}(i)=\big(\tilde{\mathbf g}_i^\top \mathbf M_k^{-1}\mathbf u_k\big)\,w_{\mathrm L}(i)\,w_{\mathrm R}(i),
\]
where \(\mathbf M_k=\Pi_k\big((1-\alpha)\Phi_{\mathrm{pos}}+\alpha\Phi_{\mathrm{neg}}+\lambda\mathbf I\big)\Pi_k^\top\).

\section{Score Distribution}
\label{app:distribution}

\begin{figure*}[!tp]
    \centering
    \includegraphics[width=0.9\linewidth]{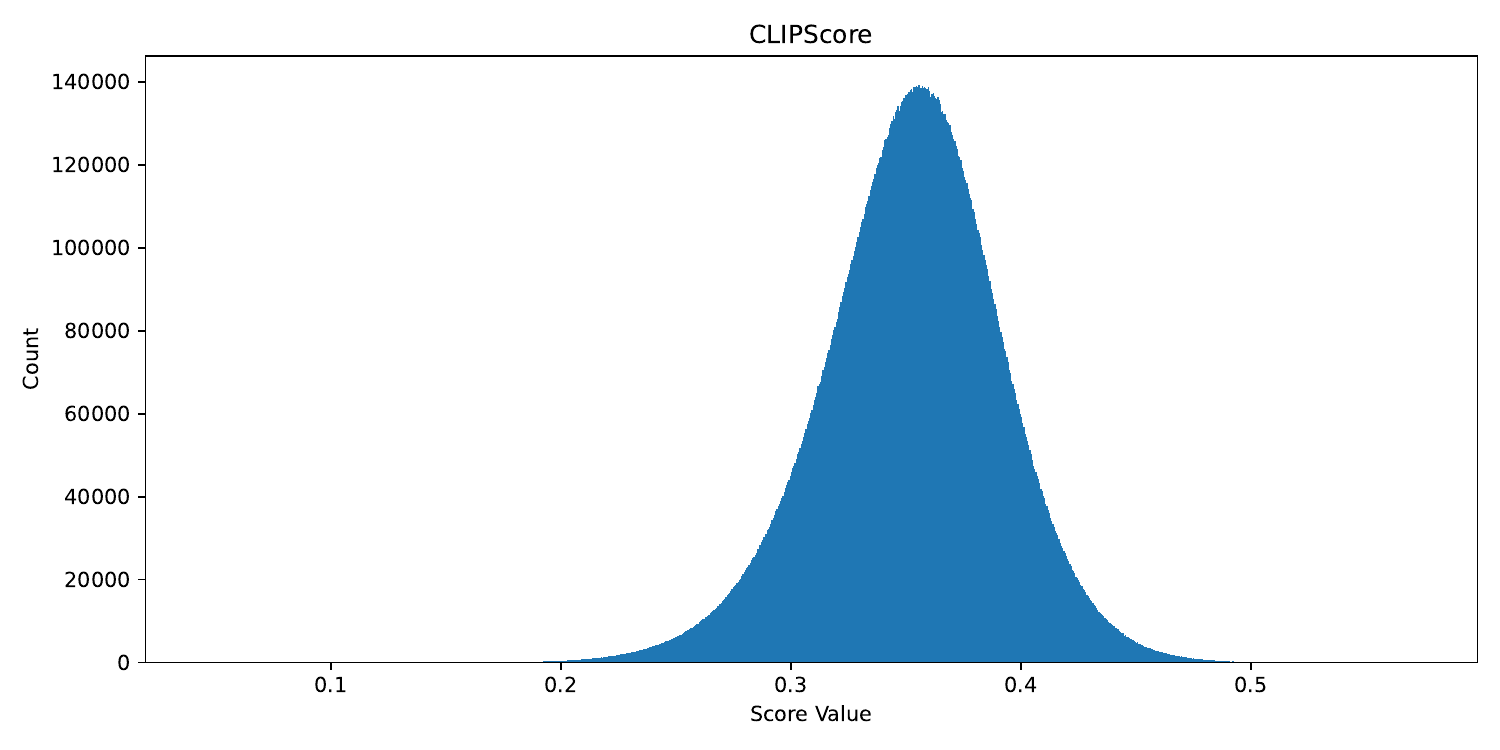}
    \caption{Distribution of CLIPScore on BIOMEDICA.}
    \label{fig:clipscore}
\end{figure*}

\begin{figure*}[!tp]
    \centering
    \includegraphics[width=0.9\linewidth]{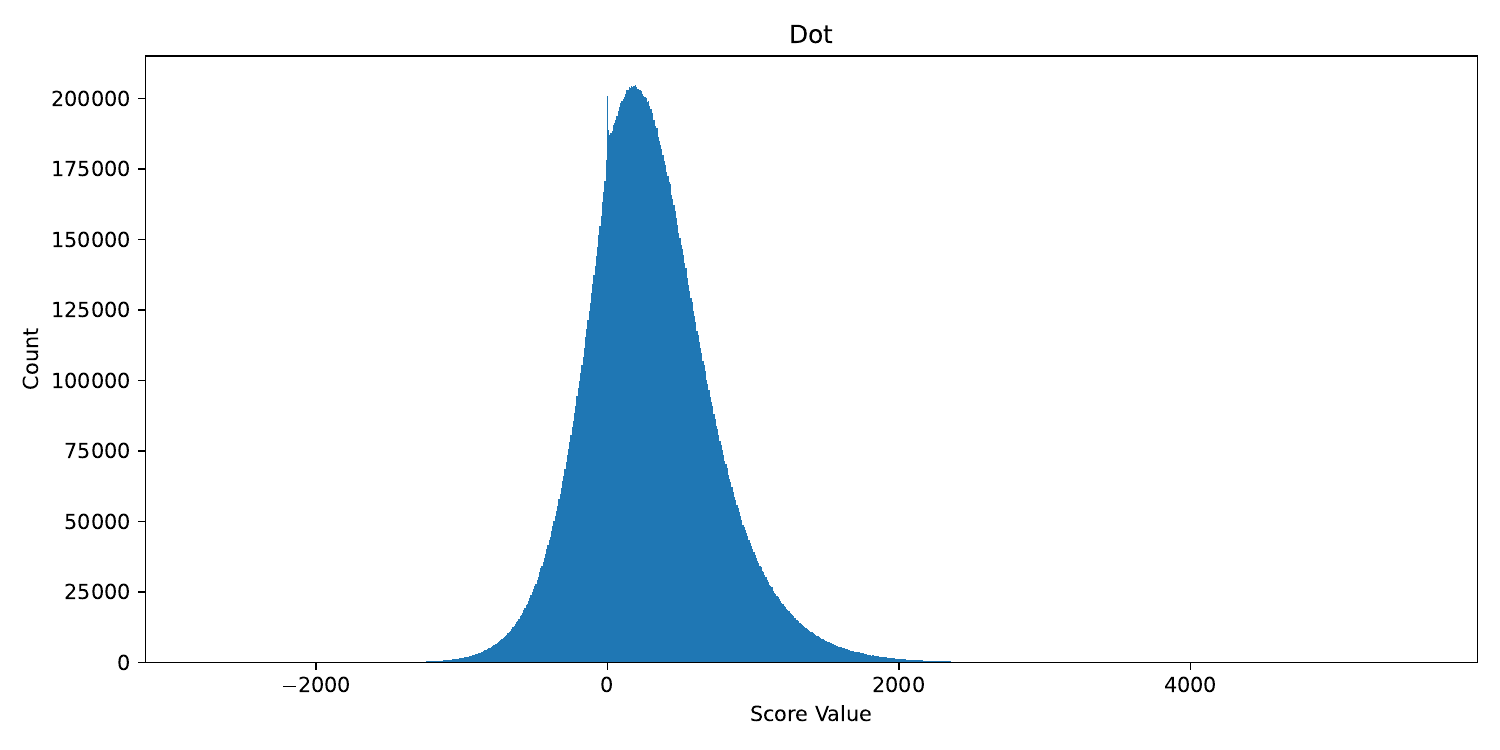}
    \caption{Distribution of Dot on BIOMEDICA.}
    \label{fig:dot}
\end{figure*}

\begin{figure*}[!tp]
    \centering
    \includegraphics[width=0.9\linewidth]{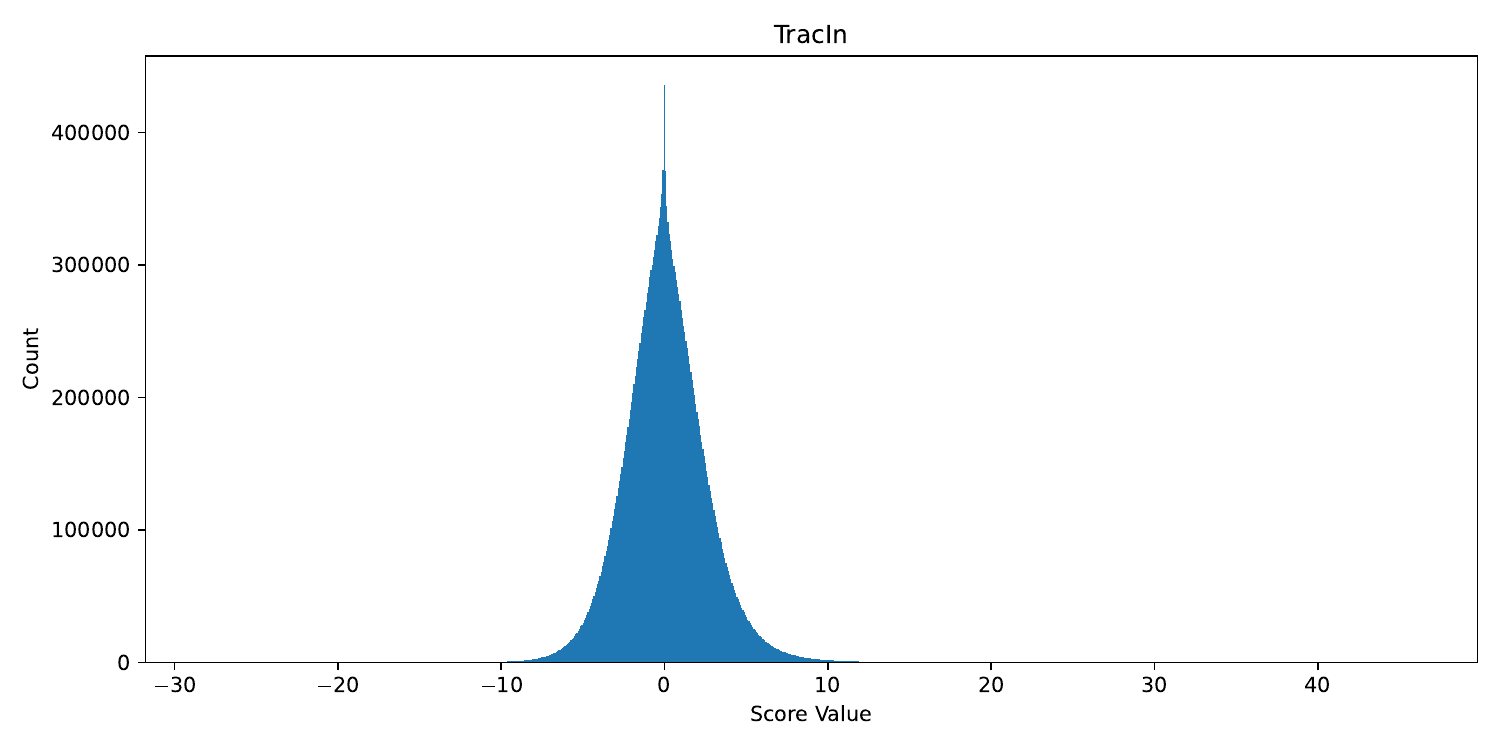}
    \caption{Distribution of TracIn on BIOMEDICA.}
    \label{fig:tracin}
\end{figure*}

\begin{figure*}[!tp]
    \centering
    \includegraphics[width=0.9\linewidth]{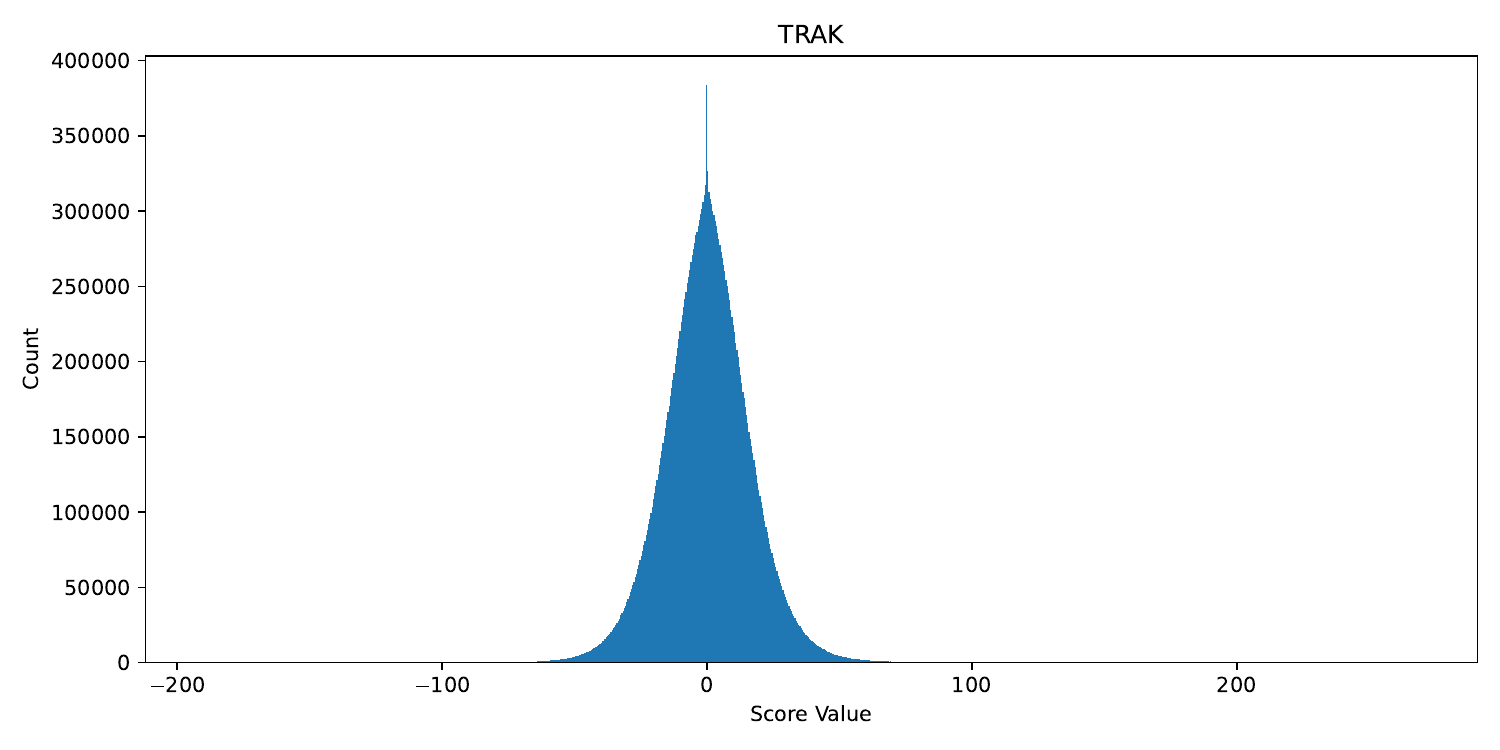}
    \caption{Distribution of TRAK on BIOMEDICA.}
    \label{fig:trak}
\end{figure*}

\begin{figure*}[!tp]
    \centering
    \includegraphics[width=0.9\linewidth]{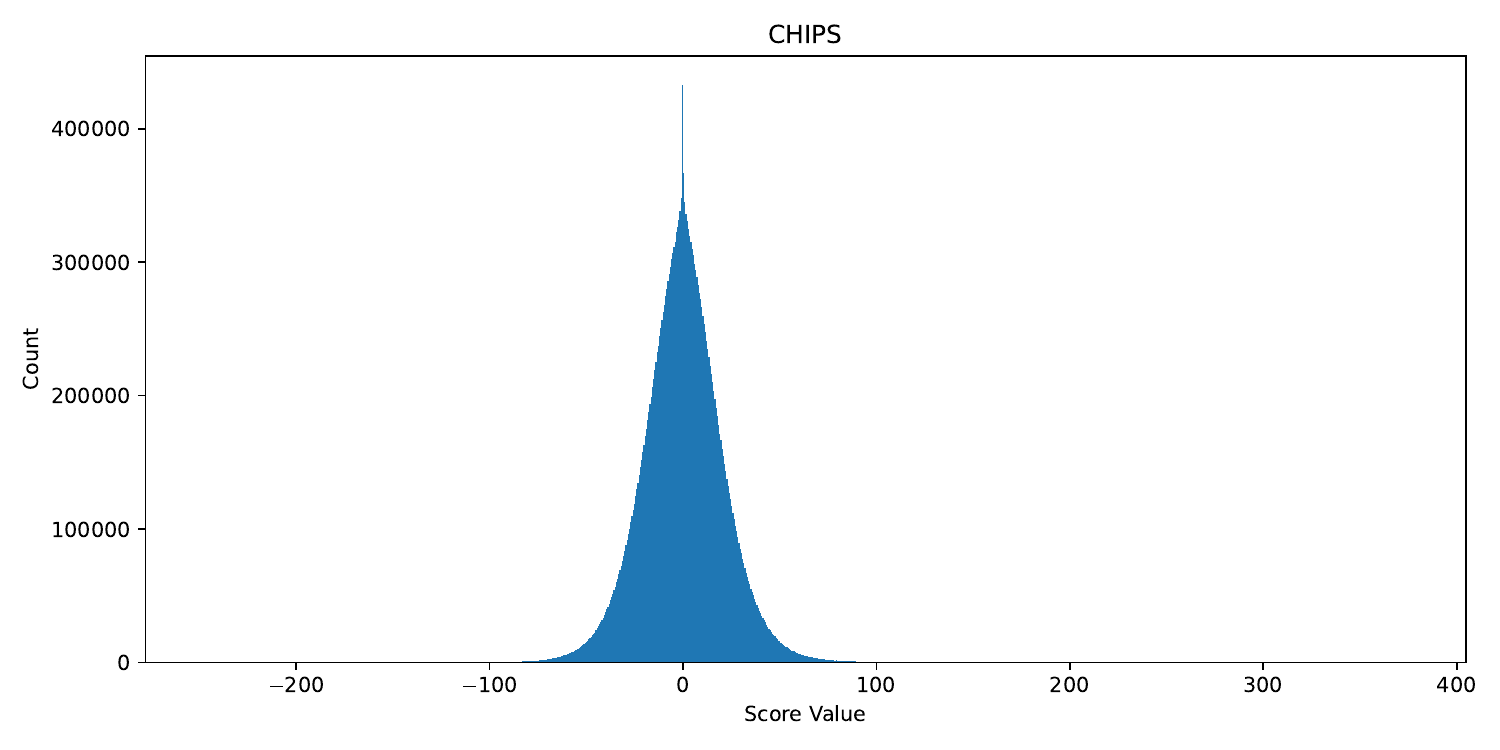}
    \caption{Distribution of CHIPS on BIOMEDICA.}
    \label{fig:chips}
\end{figure*}

The score distributions of CLIPScore, Dot, TracIn, TRAK, and CHIPS are provided in Fig.~\ref{fig:clipscore}, Fig.~\ref{fig:dot}, Fig.~\ref{fig:tracin}, Fig.~\ref{fig:trak}, and Fig.~\ref{fig:chips}.

\section{Training}
\label{app:train}

\subsection{Software and Distributed Setup}
\label{app:train:dist}

\paragraph{Framework and precision.}
We train with PyTorch in distributed data-parallel mode. On H200 GPUs we use bfloat16 for tensor computation with an fp32 master copy of parameters for the AdamW states. The logit-scale parameter is maintained as \(\tilde\tau\in\mathbb R\) and realized at runtime by \(\tau=\exp(\tilde\tau)\) to guarantee positivity. Softmax and cross-entropy are evaluated with numerically stable log-sum-exp.

\paragraph{Communication.}
Gradients are synchronized by NCCL with bucket size tuned to saturate NVLink. We enable gradient accumulation when the per-GPU micro-batch would otherwise exceed memory budget. Let \(G\) be the number of GPUs, \(B_{\text{micro}}\) the per-GPU micro-batch, and \(A\) the accumulation steps, then the global batch is
\[
B_{\text{global}} = G \times B_{\text{micro}} \times A.
\]
We choose $G=8$, \(B_{\text{micro}}=4096\) and \(A=1\) such that \(B_{\text{global}}=32{,}768\) as in the main content.

\paragraph{Determinism.}
We fix seeds for Python, NumPy, and CUDA RNGs and set cuDNN to deterministic kernels where available. Dataset sharding and sampler seeds are logged per epoch to make re-runs bitwise reproducible up to nondeterminism in fused kernels.

\subsection{Data Preprocessing and Batching}
\label{app:train:data}

\paragraph{Image pipeline.}
Training images are decoded to RGB and resized to the default input resolution of each MetaCLIP variant. 
We apply random resized crop that preserves aspect ratio, followed by horizontal flip with probability \(0.5\). 
Pixel intensities are normalized by CLIP mean and variance. 
Center crop is used at evaluation time.

\paragraph{Text pipeline.}
Texts are normalized by Unicode and then tokenized by the CLIP tokenizer. 
Sequences are truncated to the CLIP maximum length (77 tokens) with special tokens preserved.

\paragraph{Sharding and streaming.}
The training pools are sharded into balanced files on disk to avoid hot spots. Each worker maintains a streaming iterator with prefetching in pinned memory. We keep sharding consistent across methods so that all runs process the same examples at the same step counts.

\subsection{Optimization Schedule}
\label{app:train:opt}

\paragraph{Optimizer and parameter groups.}
We use AdamW with \((\beta_1,\beta_2,\epsilon)=(0.9,0.98,10^{-6})\) as stated in the main text. Weight decay is applied to all trainable parameters except LayerNorm and bias parameters. 

\paragraph{Learning-rate scheduler.}
We use a cosine decay scheduler over the total number of optimizer steps. Unless specified otherwise, the peak learning rate equals the initial value \(10^{-6}\). We optionally use a short linear warmup of \(T_{\text{warm}}\) steps when training with heavier augmentation or larger \(B_{\text{global}}\). The final learning rate floor is set to \(0\) unless explicit floor is reported.

\subsection{Counting Steps and Budget Fairness}
\label{app:train:budget}

Let \(n=\lfloor r\times|\mathcal D_{\text{train}}|\rfloor\) be the number of retained samples, \(E\) the number of epochs, and \(B_{\text{global}}\) the global batch size. The number of optimizer steps per run is
\[
T_{\text{steps}} \;=\; \left\lceil \frac{n}{B_{\text{global}}} \right\rceil \times E.
\]
All methods, including baselines and CHIPS, are trained for the same \(T_{\text{steps}}\). The wall-clock overhead of selection is measured separately and reported as a percentage of the total training time.

\section{Evaluation}
\label{app:eval}

To ensure a comprehensive and reproducible evaluation, we include 48 datasets covering both general-domain and medical-domain visual understanding tasks for this paper. Detailed descriptions are provided below.  
\subsection{General-Domain Datasets}
For General-Domain Datasets, they span a variety of recognition and retrieval tasks, including classification, fine-grained categorization, visual reasoning, robustness evaluation, and multimodal understanding. To be specific, we give an overall description for each as follows: \\
\textbf{Object \& Scene Classification:}
        \begin{itemize}
            \item \textbf{ImageNet-1K}~\cite{ImageNet}: A large-scale object classification benchmark with ~1.28 million training images and 1,000 object categories, widely used as a baseline in deep vision.  
            \item \textbf{ImageNetV2}~\cite{ImageNetV2}: A re-collected test set for ImageNet, designed to assess generalisation under dataset shift for the same 1,000 categories.  
            \item \textbf{SUN397}~\cite{SUN}: A scene recognition dataset containing 397 scene categories of diverse indoor/outdoor types, enabling evaluation on scene-level classification.  
            \item \textbf{Caltech-101}~\cite{Caltech101}: A mid-scale object classification benchmark with 101 categories and around 9,000 images, used historically for transfer-learning studies.  
            \item \textbf{VOC2007}~\cite{VOC2007}: The PASCAL Visual Object Classes 2007 dataset, including object classification (and detection) tasks across 20 categories in real-world scenes.
        \end{itemize}
\textbf{Fine-Grained Recognition:}
        \begin{itemize}
            \item \textbf{Cars}~\cite{Cars}: A fine-grained vehicle classification dataset containing many car make/model/year classes, evaluating subtle visual differences among similar objects.  
            \item \textbf{Aircraft}~\cite{Aircraft}: A fine-grained aircraft dataset distinguishing among many aircraft models and variants, used for high-granularity recognition research.  
            \item \textbf{Food101}~\cite{Food101}: A food image dataset covering 101 food types, with ~101,000 images total, for fine-grained categorisation of dishes.  
            \item \textbf{Oxford Flowers (Flowers102)}~\cite{Flowers102}: A flower species classification dataset with 102 categories and ~8,000 images, common in fine-grained vision tasks.  
            \item \textbf{Pets}~\cite{Pets}: A fine-grained pet breed classification dataset (cats \& dogs) with multiple breeds and varied poses/backgrounds.  
        \end{itemize}
\textbf{Texture, Shape, Small-Scale Objects:}
        \begin{itemize}
            \item \textbf{CIFAR-10/100}~\cite{CIFAR}: Standard small-scale image classification datasets with 10 (CIFAR-10) and 100 (CIFAR-100) classes, each containing 60,000 colour images of 32×32 size.
            \item \textbf{MNIST}~\cite{MNIST}: Classic handwritten digit classification dataset with ~70,000 grayscale images of digits 0-9, often used for benchmarking basic vision models.  
            \item \textbf{STL-10}~\cite{STL10}: A dataset derived from ImageNet with 10 classes and high-resolution images, used for unsupervised / transfer learning in small-scale settings.  
            \item \textbf{smallNORB (Azimuth/Elevation)}~\cite{NORB}: A synthetic 3D object dataset capturing objects under different viewpoints (azimuth/elevation), for studying pose-invariance and representation robustness.  
            \item \textbf{SVHN}~\cite{SVHN}: Street View House Numbers dataset with real-world digital number images, used for digit recognition in situ.  
        \end{itemize}
\textbf{Robustness \& Out-of-Distribution:}
        \begin{itemize}
            \item \textbf{ImageNet-A}~\cite{ImageNet-AO}: A subset of ImageNet containing “adversarially filtered” images challenging standard models, designed to test worst-case object recognition robustness.  
            \item \textbf{ImageNet-O}~\cite{ImageNet-AO}: An out-of-distribution test set for ImageNet models, containing images from unknown classes not present in training, to measure OOD detection/generalisation.  
        \end{itemize}
\textbf{Domain-Specific Recognition:}
        \begin{itemize}
            \item \textbf{KITTI}~\cite{Geiger2013KITTI}: A dataset collected for autonomous driving, including object, scene and motion tasks—here used in the classification context of road-scene recognition.  
            \item \textbf{EuroSAT}~\cite{Helber2019EuroSAT}: A remote sensing image classification dataset with 45 scene types, aimed at land-use and satellite-image recognition.  
            \item \textbf{RESISC45}~\cite{RESISC}: Another remote sensing scene classification dataset covering 45 classes of aerial images, evaluating models on high-altitude imagery.  
        \end{itemize}
\textbf{Visual Reasoning \& Synthetic Tasks:}
        \begin{itemize}
            \item \textbf{CLEVR (Closest-Object-Distance / Count-All)}~\cite{CLEVR}: A synthetic benchmark for visual reasoning, where models answer compositional questions about objects (distance/count) in rendered scenes.  
            \item \textbf{DTD}~\cite{DTD}: The Describable Texture Dataset, focusing on texture/material recognition with diverse patterns, supporting generalisation in less object-centric tasks.  
        \end{itemize}
\textbf{Image-Text Retrieval / Multimodal:}
        \begin{itemize}
            \item \textbf{Flickr8k}~\cite{flickr8k}: An image-caption dataset with 8,000 images, used for evaluating image-to-text and text-to-image retrieval and captioning systems.  
            \item \textbf{Flickr30k}~\cite{flickr30k}: A larger image-caption dataset with 30,000 images, used widely in cross-modal retrieval research.  
            \item \textbf{MSCOCO}~\cite{mscoco}: A large-scale multimodal dataset with 120K+ images and captions, supporting image detection, segmentation and retrieval tasks.  
            \item \textbf{Rendered-SST2}~\cite{clip}: A dataset created by rendering the sentences from the Stanford Sentiment Treebank v2 into images (positive/negative labels), used to assess optical-character-recognition via image encoders. (Train: 6,920 images; Val: 872; Test: 1,821).  
        \end{itemize}
\textbf{Geographic / Landmark Classification:}
        \begin{itemize}
            \item \textbf{Country211}~\cite{clip}: A dataset designed for geolocation classification, filtered from the YFCC100M dataset—211 countries/territories, with 150 train / 50 val / 100 test images per country.
        \end{itemize}

\subsection{Medical-Domain Datasets}
For Medical-Domain Datasets, we include a broad range of imaging modalities and clinical specialties, covering ophthalmology, radiology, dermatology, hematology, pathology, neuropathology, and non-clinical biology. 
These datasets collectively assess model performance in clinically relevant scenarios and diagnostic contexts.  From a task perspective, they can be broadly grouped into three major categories: \textit{disease classification}, \textit{organ and tissue recognition}, and \textit{pathological image analysis}, with an additional category for \textit{non-clinical biological imaging}. 
Together, they form a comprehensive benchmark for evaluating generalization and robustness of visual models in biomedical applications. In detail, the brief introduction of included datasets are listed:\\
\textbf{Disease Classification:}
    \begin{itemize}
        \item \textbf{Diabetic Retinopathy}~\cite{diabetic}: A fundus-image dataset for grading diabetic retinopathy severity in ophthalmology, used for multi-class disease classification.  
        \item \textbf{RetinaMNIST / OCTMNIST}~\cite{medmnistv2}: Retinal fundus and optical coherence tomography (OCT) image datasets for ophthalmic disease classification across multiple categories; part of the MedMNIST benchmark (about 708k 2D images in total for the MedMNIST collection).
        \item \textbf{ChestMNIST}~\cite{medmnistv2}: Based on the NIH-ChestXray14 dataset, ChestMNIST contains 112,120 chest X-ray images, formulated as a multi-label classification task for detecting 14 different diseases.  
        \item \textbf{ChestX-ray14}~\cite{chestx-ray14}: A large collection of over 100,000 frontal-view chest X-ray images from more than 30,000 patients. The labels for eight common thoracic diseases were automatically generated by text-mining the associated radiological reports.  
        \item \textbf{DermaMNIST}~\cite{medmnistv2}: This is a multi-class classification dataset of 10,015 dermatoscopic images for identifying 7 different types of common pigmented skin lesions.
        \item \textbf{BloodMNIST}~\cite{medmnistv2}: A hematology microscope-image dataset containing 17,092 images across 8 classes, used for blood-cell type classification.  
    \end{itemize}
\textbf{Organ and Tissue Recognition:}
    \begin{itemize}
        \item \textbf{OrganAMNIST / OrganCMNIST / OrganSMNIST}~\cite{medmnistv2}: This dataset is for multi-class classification of 11 body organs, containing 58,850 images derived from axial-view slices of abdominal CT scans and resized to 28x28 pixels.  
        \item \textbf{TissueMNIST}~\cite{medmnistv2}: Sourced from the Broad Bioimage Benchmark Collection, TissueMNIST is a large-scale dataset of 236,386 human kidney cortex cells, organized into 8 categories for a multi-class classification task.  
    \end{itemize}
\textbf{Pathological Image Analysis:}
    \begin{itemize}
        \item \textbf{PCAM}~\cite{pcam}: The dataset is a large-scale collection of 96x96 pixel histopathology image patches extracted from the Camelyon16 challenge, designed for the task of identifying metastatic cancer in lymph node sections.  
        \item \textbf{LC25000}~\cite{LC25000}: The LC25000 dataset contains 25,000 color histopathological images across five classes, featuring both cancerous and benign tissues from the lung and colon.
        \item \textbf{PathMNIST}~\cite{medmnistv2}: PathMNIST is a multi-class classification dataset derived from colorectal cancer histology slides. It is comprised of 107,180 image patches, categorized into 9 distinct tissue types.  
        \item \textbf{Amyloid CAA/Diffuse}~\cite{amyloid}: It includes 100495 annotations on 20099 candidate amyloid beta neuropathologies, which is a neuropathology image dataset for subtypes of amyloid pathology in brain tissue, used for subtype classification tasks. 
    \end{itemize}
\textbf{Non-Clinical Biological Imaging:}
    \begin{itemize}
        \item \textbf{Pollen}~\cite{icpr2020_pollen}: The Pollen13K dataset is a large-scale collection of over 13,000 microscopic pollen grain images from aerobiological samples, used for biological particle classification. It was chosen to test model generalization on complex, non-clinical imagery. 
    \end{itemize}

\section{FLOPs Computation}
\label{app:computation}

This appendix consolidates the FLOPs counting procedure for BIOMEDICA \cite{biomedica} used to report scoring cost.

All totals below measure a single complete \textit{scoring} pass over \(\mathcal D_{\text{train}}\) and use the batch primitives in Tab.~\ref{tab:batch_flops}:
\(C_{\text{fwd}}(B)\), \(C_{\text{bwd}}(B)\), \(C_{\text{fb}}(B)=C_{\text{fwd}}(B)+C_{\text{bwd}}(B)\), and \(C_{\text{jvp}}(B)\).
The random projection (CountSketch) cost per application is \(C_{\text{rp}}\approx 2P\) (see Sec.~\ref{app:computation}).
We denote \(n_{\text{train}}=\lceil N_{\text{train}}/B_{\text{train}}\rceil\) and \(n_{\text{eval}}=\lceil N_{\text{eval}}/B_{\text{eval}}\rceil\).
TracIn uses \(E\) epochs for accumulation. TRAK and CHIPS use \(I\) conjugate-gradient (CG) iterations.

\paragraph{TracIn.}
TracIn combines a single evaluation-direction construction with per-batch JVPs and trajectory accumulation:
\begin{align}
C_{\text{TracIn}}
&= C_{\text{eval-dir}} + C_{\text{JVP-train}} + C_{\text{accum}}, \label{eq:tracin-total}\\[2pt]
C_{\text{eval-dir}}
&= n_{\text{eval}}\,\big(C_{\text{fb}}(B_{\text{eval}}) + C_{\text{rp}}\big), \label{eq:tracin-eval}\\
C_{\text{JVP-train}}
&= n_{\text{train}}\,C_{\text{jvp}}(B_{\text{train}}), \label{eq:tracin-jvp}\\
C_{\text{accum}}
&= E\,n_{\text{train}}\,C_{\text{fb}}(B_{\text{train}}). \label{eq:tracin-accum}
\end{align}

\paragraph{TRAK.}
TRAK builds a second-order score with CG in the same end-point geometry.
We write the shared backbone block once and reuse it:
\begin{align}
\mathrm{Base}
&= n_{\text{eval}}\,\big(C_{\text{fb}}(B_{\text{eval}}) + C_{\text{rp}}\big)
 \;+\; C_{\text{proto-eval}} \nonumber\\
&\quad +\; n_{\text{train}}\,\big(C_{\text{fb}}(B_{\text{train}})+C_{\text{rp}}\big) \nonumber\\
&\quad +\; I\,n_{\text{train}}\,\big(C_{\text{jvp}}(B_{\text{train}})+C_{\text{fb}}(B_{\text{train}})+C_{\text{rp}}\big) \nonumber\\
&\quad +\; n_{\text{train}}\,\big(C_{\text{jvp}}(B_{\text{train}})+C_{\text{fwd}}(B_{\text{train}})\big). \label{eq:trak-base}
\end{align}
The TRAK total is simply
\begin{align}
C_{\text{TRAK}}=\mathrm{Base}. \label{eq:trak-total}
\end{align}

\begin{table*}[!t]
\centering
\small
\begin{tabular}{l l l r r}
\toprule
\textbf{Quantity} & \textbf{Meaning} & \textbf{General formula} & \(\boldsymbol{B_{\text{train}}}\) & \(\boldsymbol{B_{\text{eval}}}\) \\
\midrule
\(C_{\text{lin}}\)   & two linear projections to \(d\)            & \(2\,B\,(d_v{+}d_t)\,d\)                            & \(4.294967296\times 10^{10}\)  & \(4.456448\times 10^{9}\) \\
\(C_{\text{norm}}\)  & two L2 normalizations                      & \(\approx 6\,B\,d\)                                 & \(1.00663296\times 10^{8}\)    & \(1.04448\times 10^{7}\) \\
\(C_{\text{mm}}\)    & logits matmul for one direction            & \(2\,B^{2}\,d\)                                      & \(1.099511627776\times 10^{12}\) & \(1.183744\times 10^{10}\) \\
\midrule
\(C_{\text{fwd}}\)   & forward, both directions                   & \(C_{\text{lin}} + C_{\text{norm}} + 2C_{\text{mm}}\) & \(2.242073591808\times 10^{12}\) & \(2.81417728\times 10^{10}\) \\
\(C_{\text{bwd}}\)   & backward to end-point heads and \(\tau\)   & \(\approx 2\,(C_{\text{lin}} + 2C_{\text{mm}})\)      & \(4.483945857024\times 10^{12}\) & \(5.6262656\times 10^{10}\) \\
\(C_{\text{fb}}\)    & forward + backward                         & \(C_{\text{fwd}} + C_{\text{bwd}}\)                   & \(6.726019448832\times 10^{12}\) & \(8.44044288\times 10^{10}\) \\
\(C_{\text{jvp}}\)   & JVP cost upper bound\(^\dagger\)           & \(\approx 2\,C_{\text{fwd}}\)                         & \(4.484147183616\times 10^{12}\) & \(5.62835456\times 10^{10}\) \\
\midrule
\(C_{\text{proto-eval}}\) & eval prototypes only                   & \(C_{\text{lin}}(B_{\text{eval}})+C_{\text{norm}}(B_{\text{eval}})\) & \multicolumn{2}{c}{\(4.4668928\times 10^{9}\)} \\
\bottomrule
\end{tabular}
\caption{Batch-level FLOPs primitives and instantiated costs. Symmetric CLIP loss computes logits in both directions. \(^\dagger\)\(C_{\text{jvp}}\) uses the empirical bound \(C_{\text{jvp}}\approx 2\,C_{\text{fwd}}\) for this architecture and batching.}
\label{tab:batch_flops}
\end{table*}

\paragraph{CHIPS.}
CHIPS shares \(\mathrm{Base}\) with TRAK and adds three lightweight terms for
(i) negative-pair curvature in the sketched space,
(ii) hardest-negative margin bookkeeping,
and (iii) two prototype similarities per batch for relevance.
To avoid overlong lines we factor these contributions:
\begin{align}
\Delta_{\text{neg}}
&\approx c_{\text{neg}}\,k \\
\Delta_{\text{margin}}
&\approx 2\,B_{\text{train}}^{2} \\
\Delta_{\text{rel}}
&\approx 4\,B_{\text{train}}\,d
\end{align}
which are vector-level ops per CG iteration, hardest-negative search in a \(B_{\text{train}}\times B_{\text{train}}\) block, and two prototype similarities per batch.

The CHIPS total is then
\begin{align}
C_{\text{CHIPS}}
&= \mathrm{Base} \nonumber\\
&\quad +\; I\,\Delta_{\text{neg}} \nonumber\\
&\quad +\; n_{\text{train}}\,\Delta_{\text{margin}} \nonumber\\
&\quad +\; n_{\text{train}}\,\Delta_{\text{rel}}. \label{eq:chips-total}
\end{align}
\(C_{\text{rp}}\) is incurred when forming the sketched evaluation direction and wherever sketched vectors are refreshed.
\(\Delta_{\text{neg}}\) accounts for a handful of axpy-like operations per CG iteration in the \(k\)-dimensional sketched space.
\(\Delta_{\text{margin}}\) is a max-reduction over off-diagonal logits already materialized for symmetric InfoNCE.
\(\Delta_{\text{rel}}\) computes two dot products per sample with cached prototypes \((\boldsymbol\mu_x,\boldsymbol\mu_y)\).

\subsection{Numerical Totals}
\label{app:computation:nums}

We instantiate Eqs.~\eqref{eq:tracin-total}-\eqref{eq:chips-total} with the fixed values above.
We use \(E=10\) epochs for TracIn and \(I=5\) CG iterations for TRAK and CHIPS.

\begin{align*}
\textbf{TracIn} \ (E{=}10):\quad & 5.258869\times 10^{16}.\\[2pt]
\textbf{TRAK} \ (I{=}5):\quad & 5.094585\times 10^{16}.\\[2pt]
\textbf{CHIPS} \ (I{=}5):\quad & 5.094747\times 10^{16}.
\end{align*}

\subsection{Assumptions and Further notes}
\label{app:computation:assump}

\begin{itemize}[leftmargin=1.25em]
\item Symmetric CLIP computes logits in both directions; \(C_{\text{fwd}}(B)\) already includes the two \(B\times B\) matmuls in Eq.~\eqref{tab:batch_flops}.
\item The empirical bound \(C_{\text{jvp}}(B)\approx 2\,C_{\text{fwd}}(B)\) holds for our implementation and batch shapes.
\item The evaluation mean-gradient uses a single split with \(B_{\text{eval}}=3400\). \(C_{\text{proto-eval}}\) counts only projections and L2 norms.
\item CountSketch is the default random projection; for other sketches replace \(C_{\text{rp}}\) with the appropriate cost model (sparse RP \(2sP\), SRHT \(2m\log_2 m\), dense Gaussian \(2kP\)).
\item FLOPs are operation counts independent of arithmetic precision and exclude file I/O and host preprocessing.
\item \textit{Scaling summary.} With fixed \(B\) and \(d\), the dominant terms scale as
TracIn \(=\Theta(E\,n_{\text{train}}\,C_{\text{fb}})\),
TRAK \(=\Theta(I\,n_{\text{train}}\,C_{\text{fb}})\),
CHIPS \(=\Theta(I\,n_{\text{train}}\,C_{\text{fb}}) + O(Ik + n_{\text{train}}B^2 + n_{\text{train}}Bd)\).
At the reported \(k\) and \(B\), the \(O(Ik)\) and \(O(n_{\text{train}}Bd)\) extras are negligible and \(\Delta_{\text{margin}}\) is amortized by already-computed logits.
\end{itemize}

\section{Full Results}
\label{app:full_results}

\subsection{Main Experiment}

The full results of the main experiment are detailed in Tab.~\ref{tab:full_main_medical_1} and Tab.~\ref{tab:full_main_medical_2} for the medical domain, and in Tab.~\ref{tab:full_main_general_1} through Tab.~\ref{tab:full_main_general_4} for the general domain.

\begin{table*}[tp]
\centering
\small
\begin{tabular}{lcccccccc}
\toprule
\multicolumn{1}{c}{\textbf{Model}}    & \textbf{Diabetic} & \textbf{PCAM} & \textbf{LC25000} & \textbf{Pollen} & \textbf{Amyloid CAA} & \textbf{Amyloid Diffuse} & \textbf{BloodMNIST} & \textbf{ChestMNIST} \\ \midrule
PubMedCLIP                            & 63.25        & 51.46         & 8.16             & 25.27           & 94.12        & 26.88        & 8.45           & 19.01          \\
BioMedCLIP                            & 2.26         & 47.53         & 20.19            & 72.91           & 1.65         & 87.92        & 10.14          & 3.43           \\
BMCLIP                                & 69.95        & 73.96         & 38.00            & 17.28           & 1.80         & 74.72        & 9.44           & 8.77           \\ \midrule
Vanilla                               & 2.58         & 55.77         & 20.15            & 10.56           & 52.61        & 63.31        & 17.95          & 57.67          \\
\tableLineColorGray Full Dataset      & 60.89        & 71.62         & 44.83            & 9.81            & 88.30        & 77.27        & 23.33          & 62.11          \\ \midrule
\textit{$r$=10\%}                     &              &               &                  &                 &              &              &                &                \\
Random                                & 67.29        & 68.43         & 38.91            & 10.18           & 19.85        & 68.07        & 20.61          & 8.44           \\
Concept-Balance                       & 61.18        & 66.89         & 39.71            & 10.87           & 11.07        & 82.57        & 21.81          & 15.62          \\
Concept-Filter                        & 54.91        & 68.60         & 40.00            & 11.06           & 6.85         & 82.21        & 23.41          & 9.93           \\
CLIPScore                             & 2.49         & 60.68         & 40.67            & 10.56           & 72.81        & 55.49        & 20.99          & 27.80          \\
Dot                                   & 71.67        & 57.20         & 39.89            & 23.51           & 72.64        & 18.34        & 16.87          & 5.50           \\
TracIn                                & 65.40        & 56.75         & 40.03            & 26.34           & 90.73        & 12.84        & 16.69          & 3.69           \\
TRAK                                  & 43.23        & 59.23         & 39.57            & 12.88           & 77.16        & 14.74        & 25.93          & 3.60           \\
\tableLineColor \textbf{CHIPS (ours)} & 68.35        & 63.66         & 40.51            & 25.96           & 88.06        & 12.38        & 21.43          & 6.24           \\ \midrule
\textit{$r$=20\%}                     &              &               &                  &                 &              &              &                &                \\
Random                                & 65.86        & 57.13         & 40.55            & 11.13           & 10.30        & 62.52        & 21.92          & 15.43          \\
Concept-Balance                       & 55.50        & 69.32         & 40.13            & 10.37           & 3.49         & 82.83        & 24.00          & 15.01          \\
Concept-Filter                        & 54.92        & 69.78         & 41.04            & 15.21           & 3.25         & 68.44        & 25.46          & 6.93           \\
CLIPScore                             & 4.34         & 59.92         & 47.44            & 11.25           & 48.38        & 16.31        & 18.82          & 11.17          \\
Dot                                   & 72.11        & 60.07         & 31.49            & 33.63           & 62.01        & 45.43        & 13.30          & 11.07          \\
TracIn                                & 72.77        & 57.24         & 36.61            & 19.17           & 75.28        & 16.98        & 19.70          & 3.00           \\
TRAK                                  & 71.76        & 58.73         & 41.71            & 28.16           & 42.51        & 15.15        & 16.75          & 3.31           \\
\tableLineColor \textbf{CHIPS (ours)} & 72.50        & 68.67         & 48.07            & 30.36           & 53.58        & 23.50        & 22.49          & 13.97          \\ \midrule
\textit{$r$=30\%}                     &              &               &                  &                 &              &              &                &                \\
Random                                & 70.95        & 70.79         & 41.39            & 10.94           & 29.82        & 64.34        & 18.42          & 14.17          \\
Concept-Balance                       & 55.50        & 62.83         & 41.01            & 10.43           & 5.07         & 66.69        & 24.61          & 10.15          \\
Concept-Filter                        & 54.92        & 52.14         & 41.15            & 21.06           & 5.73         & 76.31        & 25.96          & 3.97           \\
CLIPScore                             & 16.89        & 53.70         & 44.16            & 11.94           & 29.20        & 23.78        & 13.94          & 5.59           \\
Dot                                   & 70.35        & 57.56         & 32.45            & 20.74           & 44.69        & 25.38        & 15.40          & 6.39           \\
TracIn                                & 72.13        & 58.48         & 29.81            & 22.19           & 50.72        & 31.45        & 14.24          & 7.28           \\
TRAK                                  & 69.39        & 59.76         & 42.11            & 25.14           & 35.00        & 16.40        & 16.92          & 1.64           \\
\tableLineColor \textbf{CHIPS (ours)} & 73.42        & 73.66         & 49.88            & 36.34           & 67.90        & 25.15        & 25.02          & 14.39          \\ \midrule
\textit{$r$=50\%}                     &              &               &                  &                 &              &              &                &                \\
Random                                & 63.57        & 72.99         & 39.65            & 10.56           & 12.41        & 69.59        & 18.82          & 17.01          \\ \bottomrule
\end{tabular}
\caption{Full medical-domain results (Part 1/2) of the main experiment.}
\label{tab:full_main_medical_1}
\end{table*}

\begin{table*}[tp]
\centering
\small
\begin{tabular}{lccccccccc}
\toprule
\multicolumn{1}{c}{\textbf{Model}}    & \textbf{ChestXray14} & \textbf{Derma} & \textbf{Oct} & \textbf{OrganA} & \textbf{OrganC} & \textbf{OrganS} & \textbf{Path} & \textbf{Retina} & \textbf{Tissue} \\ \midrule
PubMedCLIP                            & 12.83                & 9.63           & 26.50        & 23.07           & 22.63           & 23.81           & 21.73         & 18.00           & 8.45            \\
BioMedCLIP                            & 9.78                 & 1.20           & 25.00        & 20.28           & 22.55           & 21.92           & 8.48          & 16.00           & 4.66            \\
BMCLIP                                & 9.57                 & 61.65          & 25.70        & 27.15           & 20.37           & 21.35           & 43.52         & 19.75           & 5.02            \\ \midrule
Vanilla                               & 9.35                 & 10.82          & 26.40        & 10.42           & 9.91            & 11.46           & 22.21         & 18.75           & 5.24            \\
\tableLineColorGray Full Dataset      & 8.64                 & 11.72          & 21.50        & 17.51           & 14.89           & 14.20           & 40.91         & 17.50           & 12.40           \\ \midrule
\textit{$r$=10\%}                     &                      &                &              &                 &                 &                 &               &                 &                 \\
Random                                & 6.57                 & 11.27          & 23.00        & 13.80           & 10.35           & 10.49           & 25.75         & 17.75           & 4.68            \\
Concept-Balance                       & 7.63                 & 11.67          & 21.60        & 15.07           & 11.14           & 11.09           & 27.49         & 16.75           & 5.03            \\
Concept-Filter                        & 7.44                 & 11.72          & 22.30        & 14.79           & 10.43           & 11.46           & 31.98         & 17.00           & 5.14            \\
CLIPScore                             & 6.86                 & 11.47          & 27.20        & 8.51            & 8.99            & 11.60           & 25.79         & 16.75           & 5.25            \\
Dot                                   & 6.17                 & 12.17          & 20.60        & 9.48            & 10.36           & 11.19           & 25.42         & 33.75           & 4.16            \\
TracIn                                & 6.53                 & 12.02          & 18.00        & 13.98           & 12.85           & 15.28           & 22.98         & 40.50           & 4.96            \\
TRAK                                  & 25.84                & 12.17          & 20.90        & 10.74           & 10.43           & 11.22           & 28.68         & 26.75           & 4.46            \\
\tableLineColor \textbf{CHIPS (ours)} & 16.63                & 11.87          & 18.20        & 12.03           & 9.64            & 10.71           & 23.90         & 30.25           & 5.29            \\ \midrule
\textit{$r$=20\%}                     &                      &                &              &                 &                 &                 &               &                 &                 \\
Random                                & 17.84                & 11.62          & 24.00        & 14.75           & 10.13           & 10.93           & 29.86         & 17.50           & 6.60            \\
Concept-Balance                       & 9.46                 & 11.97          & 25.30        & 17.65           & 13.19           & 12.77           & 33.75         & 17.50           & 5.42            \\
Concept-Filter                        & 8.41                 & 11.97          & 22.40        & 17.28           & 13.01           & 12.85           & 34.54         & 16.75           & 6.42            \\
CLIPScore                             & 4.94                 & 11.27          & 27.40        & 10.10           & 11.25           & 11.69           & 22.34         & 20.25           & 4.45            \\
Dot                                   & 7.15                 & 11.77          & 22.00        & 12.35           & 12.50           & 12.87           & 25.14         & 23.50           & 5.38            \\
TracIn                                & 11.96                & 11.82          & 17.40        & 13.39           & 13.94           & 19.12           & 28.91         & 37.25           & 5.07            \\
TRAK                                  & 6.94                 & 11.02          & 23.50        & 12.42           & 13.49           & 14.43           & 25.10         & 34.50           & 5.41            \\
\tableLineColor \textbf{CHIPS (ours)} & 8.96                 & 11.87          & 21.40        & 13.50           & 13.80           & 14.87           & 23.20         & 38.50           & 5.92            \\ \midrule
\textit{$r$=30\%}                     &                      &                &              &                 &                 &                 &               &                 &                 \\
Random                                & 7.93                 & 12.07          & 22.10        & 15.32           & 10.76           & 11.33           & 30.60         & 17.25           & 6.42            \\
Concept-Balance                       & 8.40                 & 12.07          & 23.60        & 16.53           & 12.38           & 11.92           & 33.11         & 16.75           & 6.48            \\
Concept-Filter                        & 7.52                 & 12.02          & 25.20        & 17.38           & 13.72           & 13.57           & 36.16         & 17.00           & 6.81            \\
CLIPScore                             & 4.19                 & 11.32          & 26.90        & 9.64            & 11.16           & 10.69           & 28.69         & 22.00           & 4.49            \\
Dot                                   & 10.49                & 11.62          & 21.40        & 13.21           & 12.61           & 12.56           & 30.03         & 23.50           & 7.69            \\
TracIn                                & 17.75                & 11.97          & 18.00        & 14.57           & 14.57           & 16.74           & 33.68         & 28.75           & 5.69            \\
TRAK                                  & 6.49                 & 11.42          & 19.20        & 13.80           & 14.61           & 15.91           & 22.98         & 28.25           & 7.75            \\
\tableLineColor \textbf{CHIPS (ours)} & 10.65                & 11.37          & 21.30        & 13.13           & 14.28           & 15.33           & 24.43         & 30.25           & 7.69            \\ \midrule
\textit{$r$=50\%}                     &                      &                &              &                 &                 &                 &               &                 &                 \\
Random                                & 9.86                 & 11.82          & 22.60        & 16.86           & 12.66           & 12.90           & 35.31         & 17.25           & 10.06           \\ \bottomrule
\end{tabular}
\caption{Full medical-domain results (Part 2/2) of the main experiment. Abbreviations: Derma = DermaMNIST, Oct = OctMNIST, OrganA = OrganAMNIST, OrganC = OrganCMNIST, OrganS = OrganSMNIST, Path = PathMNIST, Retina = RetinaMNIST, Tissue = TissueMNIST.}
\label{tab:full_main_medical_2}
\end{table*}

\begin{table*}[tp]
\small
\centering
\begin{tabular}{lcccccccccc}
\toprule
\multicolumn{1}{c}{\textbf{Model}}    & \textbf{Cars} & \textbf{Country211} & \textbf{FER} & \textbf{Aircraft} & \textbf{Food101} & \textbf{GTSRB} & \textbf{\begin{tabular}[c]{@{}c@{}}Imagenet\\ -A\end{tabular}} & \textbf{\begin{tabular}[c]{@{}c@{}}Imagenet\\ -O\end{tabular}} & \textbf{\begin{tabular}[c]{@{}c@{}}Imagenet\\ 1k\end{tabular}} & \textbf{\begin{tabular}[c]{@{}c@{}}Imagenet\\ v2\end{tabular}} \\ \midrule
PubMedCLIP                            & 8.51          & 3.09                & 20.30            & 4.32              & 15.43            & 15.37          & 4.97                & 14.55               & 14.95               & 13.00                \\
BioMedCLIP                            & 0.37          & 0.37                & 13.33            & 1.02              & 0.99             & 2.49           & 0.72                & 0.75                & 0.07                & 0.07                 \\
BMCLIP                                & 92.13         & 27.18               & 37.57            & 31.38             & 90.40            & 56.22          & 54.93               & 33.85               & 71.97               & 64.80                \\ \midrule
Vanilla                               & 74.29         & 22.41               & 42.57            & 28.50             & 87.24            & 43.80          & 46.97               & 39.55               & 70.78               & 62.59                \\
\tableLineColorGray Full Dataset      & 66.11         & 20.41               & 45.00            & 20.37             & 81.00            & 36.94          & 42.24               & 37.60               & 66.28               & 58.90                \\ \midrule
\textit{$r$=10\%}                     &               &                     &                  &                   &                  &                &                     &                     &                     & \multicolumn{1}{l}{} \\
Random                                & 71.43         & 21.76               & 43.73            & 27.06             & 84.99            & 41.94          & 45.60               & 39.20               & 69.18               & 61.31                \\
Concept-Balance                       & 71.28         & 21.81               & 44.90            & 27.48             & 84.82            & 42.76          & 45.73               & 39.35               & 69.12               & 61.05                \\
Concept-Filter                        & 70.92         & 21.61               & 43.86            & 26.91             & 84.59            & 42.36          & 44.91               & 39.30               & 68.75               & 60.85                \\
CLIPScore                             & 74.93         & 22.42               & 42.09            & 28.77             & 86.90            & 44.25          & 47.16               & 39.65               & 70.60               & 62.63                \\
Dot                                   & 65.32         & 17.74               & 38.12            & 19.98             & 76.25            & 29.78          & 37.57               & 31.10               & 60.91               & 53.14                \\
TracIn                                & 64.82         & 17.30               & 35.94            & 19.71             & 74.90            & 27.40          & 36.41               & 29.95               & 59.22               & 51.16                \\
TRAK                                  & 65.10         & 17.69               & 37.46            & 20.58             & 76.27            & 28.89          & 37.61               & 30.15               & 60.54               & 52.58                \\
\tableLineColor \textbf{CHIPS (ours)} & 65.61         & 17.80               & 38.72            & 20.43             & 75.94            & 29.65          & 38.13               & 30.25               & 61.07               & 53.01                \\ \midrule
\textit{$r$=20\%}                     &               &                     &                  &                   &                  &                &                     &                     &                     & \multicolumn{1}{l}{} \\
Random                                & 70.20         & 21.60               & 45.29            & 25.71             & 84.47            & 41.35          & 45.23               & 39.30               & 68.67               & 60.53                \\
Concept-Balance                       & 69.12         & 21.29               & 45.89            & 25.77             & 83.75            & 39.30          & 44.53               & 38.40               & 68.26               & 60.30                \\
Concept-Filter                        & 69.42         & 21.20               & 44.61            & 25.89             & 83.36            & 38.61          & 44.03               & 38.40               & 67.87               & 59.94                \\
CLIPScore                             & 71.88         & 20.70               & 41.32            & 26.79             & 85.41            & 41.09          & 45.59               & 35.95               & 68.13               & 60.13                \\
Dot                                   & 70.84         & 18.83               & 39.24            & 22.86             & 81.28            & 29.65          & 40.45               & 31.75               & 63.04               & 55.01                \\
TracIn                                & 69.54         & 18.55               & 39.29            & 22.23             & 80.37            & 29.30          & 39.48               & 30.55               & 62.51               & 54.46                \\
TRAK                                  & 69.44         & 18.79               & 37.98            & 23.34             & 80.05            & 31.58          & 41.32               & 31.85               & 63.56               & 55.44                \\
\tableLineColor \textbf{CHIPS (ours)} & 69.32         & 19.54               & 39.40            & 24.36             & 80.05            & 31.06          & 40.33               & 32.10               & 62.97               & 55.01                \\ \midrule
\textit{$r$=30\%}                     &               &                     &                  &                   &                  &                &                     &                     &                     & \multicolumn{1}{l}{} \\
Random                                & 69.36         & 21.36               & 46.22            & 25.23             & 84.10            & 39.74          & 44.61               & 38.40               & 68.21               & 60.20                \\
Concept-Balance                       & 69.03         & 21.50               & 46.20            & 25.92             & 83.55            & 38.97          & 44.56               & 38.60               & 68.33               & 60.48                \\
Concept-Filter                        & 68.88         & 20.82               & 43.81            & 25.08             & 82.21            & 36.93          & 42.68               & 37.75               & 67.13               & 59.44                \\
CLIPScore                             & 72.11         & 20.48               & 42.05            & 26.46             & 84.78            & 38.42          & 44.92               & 35.40               & 67.68               & 59.30                \\
Dot                                   & 69.36         & 18.16               & 37.94            & 22.47             & 79.58            & 27.87          & 39.15               & 30.40               & 61.90               & 53.53                \\
TracIn                                & 68.96         & 18.20               & 39.84            & 22.89             & 79.51            & 29.26          & 39.65               & 30.15               & 61.87               & 53.63                \\
TRAK                                  & 68.82         & 18.36               & 38.63            & 23.91             & 79.20            & 30.37          & 39.47               & 30.75               & 62.41               & 54.55                \\
\tableLineColor \textbf{CHIPS (ours)} & 68.30         & 18.39               & 37.89            & 23.46             & 79.27            & 30.36          & 38.96               & 31.40               & 62.33               & 54.45                \\ \midrule
\textit{$r$=50\%}                     &               &                     &                  &                   &                  &                &                     &                     &                     & \multicolumn{1}{l}{} \\
Random                                & 68.47         & 21.03               & 46.15            & 23.28             & 82.96            & 38.74          & 43.67               & 37.90               & 67.61               & 59.90                \\ \bottomrule
\end{tabular}
\caption{Full general-domain results (Part 1/4) of the main experiment.}
\label{tab:full_main_general_1}
\end{table*}

\begin{table*}[tp]
\centering
\small
\begin{tabular}{lccccccccc}
\toprule
\multicolumn{1}{c}{\textbf{Model}}    & \textbf{MNIST} & \textbf{\begin{tabular}[c]{@{}c@{}}Rendered\\ SST2\end{tabular}} & \textbf{STL10} & \textbf{Sun397} & \textbf{\begin{tabular}[c]{@{}c@{}}Sun397\\ Official\end{tabular}} & \textbf{VOC} & \textbf{Caltech101} & \textbf{CIFAR10} & \textbf{CIFAR100} \\ \midrule
PubMedCLIP                            & 25.55          & 54.37                                                             & 59.24          & 18.27           & 16.86                                                               & 34.95            & 22.73               & 38.40            & 10.54             \\
BioMedCLIP                            & 9.58           & 50.08                                                             & 9.70           & 0.36            & 0.24                                                                & 3.55             & 0.51                & 10.31            & 1.19              \\
BMCLIP                                & 86.27          & 63.92                                                             & 97.86          & 69.79           & 69.28                                                               & 80.87            & 83.55               & 96.98            & 81.53             \\ \midrule
Vanilla                               & 47.83          & 60.57                                                             & 97.28          & 66.83           & 68.20                                                               & 72.18            & 83.80               & 90.24            & 66.69             \\
\tableLineColorGray Full Dataset      & 27.61          & 55.24                                                             & 97.29          & 65.17           & 65.64                                                               & 66.12            & 83.73               & 91.43            & 66.79             \\ \midrule
\textit{$r$=10\%}                     &                &                                                                   &                &                 &                                                                     &                  &                     &                  &                   \\
Random                                & 45.07          & 55.85                                                             & 97.05          & 66.82           & 67.89                                                               & 69.95            & 85.55               & 91.94            & 67.61             \\
Concept-Balance                       & 37.41          & 54.26                                                             & 97.12          & 67.01           & 68.09                                                               & 69.44            & 85.85               & 92.34            & 68.40             \\
Concept-Filter                        & 41.63          & 56.29                                                             & 96.96          & 66.88           & 67.80                                                               & 69.87            & 85.55               & 92.25            & 67.94             \\
CLIPScore                             & 47.82          & 60.57                                                             & 97.28          & 66.82           & 68.45                                                               & 72.70            & 84.40               & 90.83            & 67.90             \\
Dot                                   & 38.78          & 55.63                                                             & 95.54          & 58.91           & 61.10                                                               & 68.78            & 81.79               & 85.03            & 58.04             \\
TracIn                                & 39.09          & 52.94                                                             & 95.24          & 57.33           & 59.55                                                               & 68.56            & 82.17               & 82.60            & 54.60             \\
TRAK                                  & 44.99          & 56.84                                                             & 95.80          & 58.33           & 60.34                                                               & 66.35            & 83.22               & 82.17            & 54.39             \\
\tableLineColor \textbf{CHIPS (ours)} & 38.82          & 57.84                                                             & 94.91          & 58.00           & 57.88                                                               & 68.66            & 82.48               & 80.76            & 55.33             \\ \midrule
\textit{$r$=20\%}                     &                &                                                                   &                &                 &                                                                     &                  &                     &                  &                   \\
Random                                & 34.08          & 54.48                                                             & 97.17          & 66.71           & 67.57                                                               & 67.92            & 85.39               & 92.22            & 67.98             \\
Concept-Balance                       & 38.87          & 55.68                                                             & 97.14          & 66.54           & 67.39                                                               & 67.91            & 85.09               & 91.96            & 67.62             \\
Concept-Filter                        & 36.40          & 57.11                                                             & 96.88          & 66.43           & 66.99                                                               & 69.49            & 85.19               & 91.68            & 67.26             \\
CLIPScore                             & 47.78          & 58.54                                                             & 97.04          & 62.08           & 64.06                                                               & 71.39            & 83.63               & 88.03            & 61.89             \\
Dot                                   & 38.50          & 57.77                                                             & 95.36          & 56.67           & 59.19                                                               & 66.22            & 81.61               & 84.61            & 57.69             \\
TracIn                                & 37.95          & 55.19                                                             & 94.67          & 56.22           & 58.38                                                               & 67.15            & 81.71               & 84.93            & 56.99             \\
TRAK                                  & 43.87          & 51.62                                                             & 95.14          & 55.84           & 58.63                                                               & 68.10            & 82.04               & 85.28            & 58.14             \\
\tableLineColor \textbf{CHIPS (ours)} & 42.91          & 50.63                                                             & 95.42          & 55.94           & 58.51                                                               & 65.89            & 81.38               & 85.72            & 59.00             \\ \midrule
\textit{$r$=30\%}                     &                &                                                                   &                &                 &                                                                     &                  &                     &                  &                   \\
Random                                & 27.07          & 54.26                                                             & 97.15          & 66.51           & 67.18                                                               & 67.28            & 85.08               & 92.34            & 68.13             \\
Concept-Balance                       & 31.72          & 54.31                                                             & 97.00          & 66.68           & 67.37                                                               & 68.08            & 85.09               & 92.18            & 68.32             \\
Concept-Filter                        & 34.15          & 57.33                                                             & 96.80          & 65.84           & 66.55                                                               & 68.56            & 84.34               & 91.07            & 65.50             \\
CLIPScore                             & 46.31          & 57.06                                                             & 97.05          & 62.11           & 63.81                                                               & 69.97            & 83.32               & 88.35            & 62.38             \\
Dot                                   & 37.90          & 54.64                                                             & 95.31          & 55.82           & 57.89                                                               & 66.75            & 80.82               & 83.64            & 55.78             \\
TracIn                                & 37.04          & 53.21                                                             & 94.96          & 55.72           & 57.92                                                               & 67.11            & 80.84               & 83.05            & 54.84             \\
TRAK                                  & 32.68          & 50.08                                                             & 95.30          & 55.68           & 57.78                                                               & 64.52            & 80.81               & 84.56            & 57.45             \\
\tableLineColor \textbf{CHIPS (ours)} & 34.25          & 49.92                                                             & 95.36          & 55.69           & 57.91                                                               & 64.36            & 80.79               & 84.61            & 57.64             \\ \midrule
\textit{$r$=50\%}                     &                &                                                                   &                &                 &                                                                     &                  &                     &                  &                   \\
Random                                & 30.17          & 54.09                                                             & 97.21          & 66.02           & 66.73                                                               & 66.76            & 84.60               & 92.11            & 67.94             \\ \bottomrule
\end{tabular}
\caption{Full general-domain results (Part 2/4) of the main experiment.}
\label{tab:full_main_general_2}
\end{table*}

\begin{table*}[tp]
\centering
\small
\begin{tabular}{lccccccc}
\toprule
\multicolumn{1}{c}{\textbf{Model}}    & \textbf{\begin{tabular}[c]{@{}c@{}}CLEVR\\ Closest\end{tabular}} & \textbf{\begin{tabular}[c]{@{}c@{}}CLEVR\\ Count\end{tabular}} & \textbf{DMLAB} & \textbf{DTD} & \textbf{Eurosat} & \textbf{Flowers} & \textbf{KITTI} \\ \midrule
PubMedCLIP                            & 22.28                                                           & 16.23                                                          & 18.85          & 10.59        & 19.70            & 14.54            & 31.36          \\
BioMedCLIP                            & 24.51                                                           & 16.85                                                          & 16.86          & 1.17         & 11.26            & 0.59             & 29.54          \\
BMCLIP                                & 15.79                                                           & 33.98                                                          & 13.80          & 55.64        & 63.02            & 76.11            & 26.30          \\ \midrule
Vanilla                               & 22.47                                                           & 29.11                                                          & 16.11          & 56.22        & 55.96            & 73.57            & 24.47          \\
\tableLineColorGray Full Dataset      & 20.45                                                           & 21.03                                                          & 12.08          & 50.48        & 50.06            & 70.69            & 32.49          \\ \midrule
\textit{$r$=10\%}                     &                                                                 &                                                                &                &              &                  &                  &                \\
Random                                & 22.57                                                           & 29.39                                                          & 12.28          & 53.56        & 53.31            & 72.68            & 31.08          \\
Concept-Balance                       & 22.59                                                           & 31.12                                                          & 12.02          & 52.93        & 52.39            & 72.94            & 28.83          \\
Concept-Filter                        & 22.41                                                           & 31.32                                                          & 11.99          & 52.66        & 48.26            & 72.68            & 29.54          \\
CLIPScore                             & 22.45                                                           & 31.71                                                          & 15.08          & 56.60        & 56.54            & 74.21            & 22.93          \\
Dot                                   & 21.38                                                           & 25.25                                                          & 15.86          & 47.82        & 42.61            & 66.14            & 17.58          \\
TracIn                                & 21.39                                                           & 22.96                                                          & 14.83          & 45.80        & 40.17            & 66.08            & 17.02          \\
TRAK                                  & 22.55                                                           & 21.23                                                          & 14.04          & 48.14        & 37.98            & 68.12            & 20.82          \\
\tableLineColor \textbf{CHIPS (ours)} & 22.59                                                           & 21.31                                                          & 14.92          & 48.86        & 40.27            & 66.29            & 19.83          \\ \midrule
\textit{$r$=20\%}                     &                                                                 &                                                                &                &              &                  &                  &                \\
Random                                & 22.16                                                           & 26.30                                                          & 11.98          & 52.66        & 51.19            & 73.05            & 30.66          \\
Concept-Balance                       & 21.53                                                           & 29.22                                                          & 11.80          & 52.07        & 52.26            & 72.48            & 29.68          \\
Concept-Filter                        & 21.51                                                           & 30.64                                                          & 11.92          & 51.91        & 50.33            & 71.74            & 31.50          \\
CLIPScore                             & 22.47                                                           & 28.19                                                          & 14.92          & 55.00        & 48.13            & 70.60            & 20.39          \\
Dot                                   & 20.93                                                           & 20.89                                                          & 16.73          & 46.44        & 41.33            & 65.67            & 20.39          \\
TracIn                                & 18.71                                                           & 21.03                                                          & 15.45          & 43.88        & 40.69            & 64.55            & 16.60          \\
TRAK                                  & 21.21                                                           & 24.05                                                          & 15.61          & 45.11        & 41.83            & 66.45            & 18.71          \\
\tableLineColor \textbf{CHIPS (ours)} & 21.52                                                           & 25.25                                                          & 16.21          & 44.47        & 42.89            & 66.53            & 16.46          \\ \midrule
\textit{$r$=30\%}                     &                                                                 &                                                                &                &              &                  &                  &                \\
Random                                & 21.86                                                           & 26.93                                                          & 11.85          & 52.34        & 52.33            & 72.56            & 32.07          \\
Concept-Balance                       & 21.23                                                           & 27.33                                                          & 11.83          & 51.44        & 52.69            & 72.56            & 30.38          \\
Concept-Filter                        & 21.63                                                           & 27.87                                                          & 11.84          & 51.65        & 48.91            & 71.59            & 31.50          \\
CLIPScore                             & 22.44                                                           & 27.65                                                          & 14.88          & 55.11        & 47.78            & 70.92            & 18.00          \\
Dot                                   & 21.01                                                           & 20.81                                                          & 15.41          & 42.82        & 39.65            & 64.90            & 16.60          \\
TracIn                                & 18.77                                                           & 20.52                                                          & 15.44          & 43.14        & 37.85            & 65.05            & 17.02          \\
TRAK                                  & 21.40                                                           & 20.57                                                          & 15.64          & 44.26        & 41.35            & 64.68            & 15.61          \\
\tableLineColor \textbf{CHIPS (ours)} & 21.60                                                           & 20.78                                                          & 15.76          & 44.20        & 42.19            & 65.30            & 14.77          \\ \midrule
\textit{$r$=50\%}                     &                                                                 &                                                                &                &              &                  &                  &                \\
Random                                & 21.29                                                           & 24.69                                                          & 11.74          & 51.38        & 50.54            & 71.85            & 32.77          \\ \bottomrule
\end{tabular}
\caption{Full general-domain results (Part 3/4) of the main experiment.}
\label{tab:full_main_general_3}
\end{table*}

\begin{table*}[tp]
\centering
\small
\begin{tabular}{lccccc}
\toprule
\multicolumn{1}{c}{\textbf{Model}}    & \textbf{Pets} & \textbf{RESISC45} & \textbf{\begin{tabular}[c]{@{}c@{}}Smallnorb\\ Azimuth\end{tabular}} & \textbf{\begin{tabular}[c]{@{}c@{}}Smallnorb\\ Elevation\end{tabular}} & \textbf{SVHN} \\ \midrule
PubMedCLIP                            & 23.36         & 14.22             & 5.51                                                                 & 10.97                                                                  & 7.43          \\
BioMedCLIP                            & 2.89          & 2.21              & 5.64                                                                 & 10.95                                                                  & 9.83          \\
BMCLIP                                & 91.39         & 63.38             & 6.29                                                                 & 10.54                                                                  & 50.43         \\ \midrule
Vanilla                               & 90.54         & 66.08             & 5.42                                                                 & 10.92                                                                  & 24.34         \\
\tableLineColorGray Full Dataset      & 88.77         & 63.05             & 6.77                                                                 & 12.12                                                                  & 19.39         \\ \midrule
\textit{$r$=10\%}                     &               &                   &                                                                      &                                                                        &               \\
Random                                & 90.52         & 64.08             & 5.93                                                                 & 10.88                                                                  & 18.41         \\
Concept-Balance                       & 90.11         & 63.76             & 5.82                                                                 & 10.47                                                                  & 19.17         \\
Concept-Filter                        & 89.83         & 62.35             & 5.82                                                                 & 10.53                                                                  & 19.56         \\
CLIPScore                             & 90.02         & 66.38             & 5.67                                                                 & 11.51                                                                  & 25.80         \\
Dot                                   & 88.09         & 60.21             & 5.84                                                                 & 10.44                                                                  & 26.41         \\
TracIn                                & 88.23         & 60.14             & 5.47                                                                 & 11.37                                                                  & 24.96         \\
TRAK                                  & 87.57         & 60.54             & 5.26                                                                 & 11.60                                                                  & 26.31         \\
\tableLineColor \textbf{CHIPS (ours)} & 86.97         & 58.44             & 5.55                                                                 & 11.82                                                                  & 25.23         \\ \midrule
\textit{$r$=20\%}                     &               &                   &                                                                      &                                                                        &               \\
Random                                & 90.16         & 63.87             & 6.11                                                                 & 10.97                                                                  & 18.19         \\
Concept-Balance                       & 89.45         & 63.49             & 6.07                                                                 & 11.05                                                                  & 18.78         \\
Concept-Filter                        & 89.32         & 61.79             & 5.90                                                                 & 11.12                                                                  & 20.29         \\
CLIPScore                             & 89.02         & 64.83             & 5.34                                                                 & 10.80                                                                  & 27.57         \\
Dot                                   & 86.48         & 60.22             & 6.51                                                                 & 10.48                                                                  & 25.77         \\
TracIn                                & 88.25         & 59.75             & 5.37                                                                 & 10.96                                                                  & 25.78         \\
TRAK                                  & 86.59         & 60.29             & 5.67                                                                 & 11.80                                                                  & 24.32         \\
\tableLineColor \textbf{CHIPS (ours)} & 87.24         & 59.68             & 5.49                                                                 & 11.02                                                                  & 24.63         \\ \midrule
\textit{$r$=30\%}                     &               &                   &                                                                      &                                                                        &               \\
Random                                & 89.92         & 63.79             & 6.27                                                                 & 10.74                                                                  & 17.19         \\
Concept-Balance                       & 89.29         & 64.56             & 6.41                                                                 & 10.78                                                                  & 18.74         \\
Concept-Filter                        & 89.23         & 62.60             & 5.97                                                                 & 11.74                                                                  & 20.28         \\
CLIPScore                             & 88.39         & 64.11             & 5.13                                                                 & 10.89                                                                  & 26.40         \\
Dot                                   & 86.59         & 60.65             & 5.69                                                                 & 11.64                                                                  & 23.29         \\
TracIn                                & 87.76         & 59.51             & 5.54                                                                 & 10.67                                                                  & 23.28         \\
TRAK                                  & 86.84         & 59.44             & 5.60                                                                 & 11.30                                                                  & 23.07         \\
\tableLineColor \textbf{CHIPS (ours)} & 87.08         & 60.00             & 5.84                                                                 & 11.18                                                                  & 22.45         \\ \midrule
\textit{$r$=50\%}                     &               &                   &                                                                      &                                                                        &               \\
Random                                & 89.64         & 63.71             & 6.19                                                                 & 11.04                                                                  & 17.09         \\ \bottomrule
\end{tabular}
\caption{Full general-domain results (Part 4/4) of the main experiment.}
\label{tab:full_main_general_4}
\end{table*}

\subsection{Generalization Experiment}

The full results of the main experiment are detailed in Tab.~\ref{tab:full_generalization_medical_1} and Tab.~\ref{tab:full_generalization_medical_2} for the medical domain, and in Tab.~\ref{tab:full_generalization_general_1} through Tab.~\ref{tab:full_generalization_general_4} for the general domain.

\begin{table*}[tp]
\centering
\small
\begin{tabular}{lcccccccc}
\toprule
\multicolumn{1}{c}{\textbf{Model}} & \textbf{Diabetic} & \textbf{PCAM} & \textbf{LC25000} & \textbf{Pollen} & \textbf{Amyloid CAA} & \textbf{Amyloid Diffuse} & \textbf{BloodMNIST} & \textbf{ChestMNIST} \\ \midrule
\textit{B32-400M}                  &                   &               &                  &                 &                      &                          &                     &                     \\
Random                             & 53.80             & 65.05         & 40.88            & 67.57           & 93.21                & 23.61                    & 14.56               & 1.24                \\
TracIn                             & 57.38             & 55.25         & 37.12            & 50.47           & 96.33                & 12.32                    & 21.66               & 2.16                \\
\tableLineColor \textbf{CHIPS (ours)}                       & 59.94             & 55.22         & 36.27            & 69.83           & 98.35                & 13.97                    & 16.22               & 1.56                \\ \midrule
\textit{B32-CC}                    &                   &               &                  &                 &                      &                          &                     &                     \\
Random                             & 73.20             & 62.22         & 21.31            & 32.24           & 1.54                 & 87.94                    & 35.52               & 3.53                \\
TracIn                             & 66.56             & 56.44         & 19.17            & 23.26           & 5.01                 & 85.16                    & 30.20               & 11.05               \\
\tableLineColor \textbf{CHIPS (ours)}                       & 51.66             & 56.75         & 19.44            & 39.85           & 6.46                 & 87.88                    & 28.73               & 34.37               \\ \midrule
\textit{B16-400M}                  &                   &               &                  &                 &                      &                          &                     &                     \\
Random                             & 67.29             & 68.43         & 38.91            & 10.18           & 19.85                & 68.07                    & 20.61               & 8.44                \\
TracIn                             & 65.40             & 56.75         & 40.03            & 26.34           & 90.73                & 12.84                    & 16.69               & 3.69                \\
\tableLineColor \textbf{CHIPS (ours)}                      & 68.35             & 63.66         & 40.51            & 25.96           & 88.06                & 12.38                    & 21.43               & 6.24                \\ \midrule
\textit{B16-CC}                    &                   &               &                  &                 &                      &                          &                     &                     \\
Random                             & 72.85             & 59.01         & 20.61            & 16.66           & 49.21                & 87.68                    & 16.90               & 2.33                \\
TracIn                             & 74.49             & 52.07         & 20.33            & 19.04           & 91.57                & 37.16                    & 12.36               & 2.00                \\
\tableLineColor \textbf{CHIPS (ours)}                        & 76.91             & 54.99         & 32.75            & 22.88           & 34.65                & 82.59                    & 14.88               & 2.90                \\ \midrule
\textit{L14-400M}                  &                   &               &                  &                 &                      &                          &                     &                     \\
Random                             & 73.65             & 55.76         & 43.68            & 19.61           & 11.95                & 87.16                    & 9.97                & 5.72                \\
TracIn                             & 65.62             & 60.57         & 40.61            & 16.03           & 18.51                & 51.40                    & 19.47               & 5.10                \\
\tableLineColor \textbf{CHIPS (ours)}                       & 65.34             & 61.27         & 39.65            & 19.48           & 44.86                & 73.60                    & 11.14               & 5.84                \\ \midrule
\textit{L14-CC}                    &                   &               &                  &                 &                      &                          &                     &                     \\
Random                             & 47.71             & 54.50         & 24.16            & 68.13           & 11.29                & 86.15                    & 18.06               & 41.76               \\
TracIn                             & 51.44             & 56.73         & 26.40            & 60.72           & 42.54                & 83.83                    & 18.80               & 49.22               \\
\tableLineColor \textbf{CHIPS (ours)}                  & 54.22             & 56.09         & 21.73            & 63.54           & 33.75                & 84.75                    & 20.50               & 46.61               \\ \midrule
\textit{H14-CC}                             &                   &               &                  &                 &                      &                          &                     &                     \\
Random                             & 72.41             & 64.77         & 50.27            & 12.19           & 92.56                & 86.96                    & 24.44               & 2.43                \\
TracIn                             & 73.09             & 65.87         & 50.75            & 12.70           & 76.15                & 83.24                    & 14.24               & 2.92                \\
\tableLineColor \textbf{CHIPS (ours)}                       & 78.64             & 65.30         & 55.12            & 11.50           & 82.87                & 86.85                    & 20.94               & 8.00                \\ \bottomrule
\end{tabular}
\caption{Full medical-domain results (Part 1/2) of the generalization experiment.}
\label{tab:full_generalization_medical_1}
\end{table*}

\begin{table*}[tp]
\centering
\small
\begin{tabular}{lccccccccc}
\toprule
\multicolumn{1}{c}{\textbf{Model}}    & \textbf{ChestXray14} & \textbf{Derma} & \textbf{Oct} & \textbf{OrganA} & \textbf{OrganC} & \textbf{OrganS} & \textbf{Path} & \textbf{Retina} & \multicolumn{1}{l}{\textbf{Tissue}} \\ \midrule
\textit{B32-400M}                     &                      &                &              &                 &                 &                 &               &                 & \multicolumn{1}{l}{}                \\
Random                                & 6.05                 & 13.97          & 25.00        & 10.77           & 6.86            & 5.30            & 34.50         & 9.75            & 5.31                                \\
TracIn                                & 3.80                 & 17.11          & 24.80        & 9.93            & 10.32           & 9.19            & 16.56         & 36.00           & 12.49                               \\
\tableLineColor \textbf{CHIPS (ours)} & 2.64                 & 12.52          & 24.20        & 13.31           & 11.33           & 10.59           & 27.87         & 19.25           & 11.63                               \\ \midrule
\textit{B32-CC}                       &                      &                &              &                 &                 &                 &               &                 & \multicolumn{1}{l}{}                \\
Random                                & 4.42                 & 9.63           & 25.00        & 16.22           & 8.98            & 6.24            & 31.91         & 14.25           & 9.18                                \\
TracIn                                & 6.71                 & 16.06          & 24.30        & 18.37           & 11.37           & 11.03           & 27.80         & 19.00           & 10.78                               \\
\tableLineColor \textbf{CHIPS (ours)} & 9.14                 & 13.82          & 24.70        & 23.67           & 14.28           & 13.67           & 29.82         & 13.00           & 10.36                               \\ \midrule
\textit{B16-400M}                     &                      &                &              &                 &                 &                 &               &                 & \multicolumn{1}{l}{}                \\
Random                                & 6.57                 & 11.27          & 23.00        & 13.80           & 10.35           & 10.49           & 25.75         & 17.75           & 4.68                                \\
TracIn                                & 6.53                 & 12.02          & 18.00        & 13.98           & 12.85           & 15.28           & 22.98         & 40.50           & 4.96                                \\
\tableLineColor \textbf{CHIPS (ours)} & 16.63                & 11.87          & 18.20        & 12.03           & 9.64            & 10.71           & 23.90         & 30.25           & 5.29                                \\ \midrule
\textit{B16-CC}                       &                      &                &              &                 &                 &                 &               &                 & \multicolumn{1}{l}{}                \\
Random                                & 9.92                 & 11.82          & 25.50        & 17.61           & 11.04           & 12.47           & 31.50         & 6.00            & 5.38                                \\
TracIn                                & 9.12                 & 12.82          & 23.30        & 14.47           & 14.05           & 15.37           & 41.14         & 8.50            & 5.94                                \\
\tableLineColor \textbf{CHIPS (ours)} & 10.10                & 12.37          & 24.20        & 13.61           & 13.56           & 14.42           & 43.15         & 9.75            & 6.22                                \\ \midrule
\textit{L14-400M}                     &                      &                &              &                 &                 &                 &               &                 & \multicolumn{1}{l}{}                \\
Random                                & 9.81                 & 19.10          & 23.20        & 26.17           & 16.99           & 17.37           & 36.39         & 38.50           & 23.02                               \\
TracIn                                & 10.51                & 21.05          & 23.30        & 23.10           & 16.13           & 17.55           & 36.92         & 31.00           & 11.04                               \\
\tableLineColor \textbf{CHIPS (ours)} & 8.74                 & 25.74          & 22.20        & 23.01           & 15.06           & 17.07           & 35.04         & 38.50           & 17.37                               \\ \midrule
\textit{L14-CC}                       &                      &                &              &                 &                 &                 &               &                 & \multicolumn{1}{l}{}                \\
Random                                & 13.06                & 21.40          & 26.20        & 25.68           & 19.81           & 19.23           & 30.74         & 20.25           & 7.31                                \\
TracIn                                & 15.82                & 27.83          & 23.60        & 16.71           & 15.68           & 11.29           & 21.88         & 20.25           & 6.11                                \\
\tableLineColor \textbf{CHIPS (ours)} & 16.39                & 28.99          & 23.30        & 19.39           & 16.38           & 15.26           & 25.42         & 16.75           & 7.74                                \\ \midrule
\textit{H14-CC}                                &                      &                &              &                 &                 &                 &               &                 & \multicolumn{1}{l}{}                \\
Random                                & 8.97                 & 62.19          & 24.80        & 25.52           & 16.69           & 14.35           & 38.29         & 8.00            & 4.83                                \\
TracIn                                & 13.05                & 64.69          & 23.80        & 31.44           & 20.53           & 18.83           & 37.51         & 22.25           & 6.53                                \\
\tableLineColor \textbf{CHIPS (ours)} & 8.89                 & 54.16          & 23.20        & 27.00           & 18.78           & 15.67           & 34.21         & 19.25           & 7.24                                \\ \bottomrule
\end{tabular}
\caption{Full medical-domain results (Part 2/2) of the generalization experiment.}
\label{tab:full_generalization_medical_2}
\end{table*}

\begin{table*}[tp]
\centering
\small
\begin{tabular}{lcccccccccc}
\toprule
\multicolumn{1}{c}{\textbf{Model}}    & \textbf{Cars} & \textbf{Country211} & \textbf{FER} & \textbf{Aircraft} & \textbf{Food101} & \textbf{GTSRB} & \textbf{\begin{tabular}[c]{@{}c@{}}Imagenet\\ -A\end{tabular}} & \textbf{\begin{tabular}[c]{@{}c@{}}Imagenet\\ -O\end{tabular}} & \textbf{\begin{tabular}[c]{@{}c@{}}Imagenet\\ 1k\end{tabular}} & \textbf{\begin{tabular}[c]{@{}c@{}}Imagenet\\ v2\end{tabular}} \\ \midrule
\textit{B32-400M}                     &               &                     &              &                   &                  &                &                                                                &                                                                & \multicolumn{1}{l}{}                                           & \multicolumn{1}{l}{}                                           \\
Random                                & 66.66         & 16.73               & 30.54        & 25.44             & 78.84            & 38.45          & 27.85                                                          & 45.80                                                          & 63.49                                                          & 55.78                                                          \\
TracIn                                & 66.00         & 15.50               & 33.74        & 22.41             & 77.07            & 27.69          & 27.07                                                          & 40.75                                                          & 61.35                                                          & 52.69                                                          \\
\tableLineColor \textbf{CHIPS (ours)} & 65.97         & 15.85               & 31.51        & 24.42             & 77.74            & 30.26          & 28.40                                                          & 42.10                                                          & 62.46                                                          & 53.85                                                          \\ \midrule
\textit{B32-CC}                       &               &                     &              &                   &                  &                &                                                                &                                                                & \multicolumn{1}{l}{}                                           & \multicolumn{1}{l}{}                                           \\
Random                                & 74.99         & 17.20               & 43.52        & 23.34             & 80.65            & 39.71          & 28.37                                                          & 46.35                                                          & 65.90                                                          & 57.95                                                          \\
TracIn                                & 73.96         & 15.58               & 35.82        & 23.46             & 78.49            & 37.19          & 29.13                                                          & 40.85                                                          & 63.71                                                          & 55.60                                                          \\
\tableLineColor \textbf{CHIPS (ours)} & 73.37         & 16.05               & 45.75        & 24.15             & 79.41            & 35.33          & 29.73                                                          & 41.55                                                          & 64.16                                                          & 56.40                                                          \\ \midrule
\textit{B16-400M}                     &               &                     &              &                   &                  &                &                                                                &                                                                & \multicolumn{1}{l}{}                                           & \multicolumn{1}{l}{}                                           \\
Random                                & 71.43         & 21.76               & 43.73        & 27.06             & 84.99            & 41.94          & 45.60                                                          & 39.20                                                          & 69.18                                                          & 61.31                                                          \\
TracIn                                & 70.63         & 19.09               & 37.11        & 23.91             & 81.94            & 31.37          & 40.48                                                          & 31.20                                                          & 63.62                                                          & 55.34                                                          \\
\tableLineColor \textbf{CHIPS (ours)} & 68.40         & 21.47               & 40.69        & 25.32             & 82.60            & 31.02          & 42.36                                                          & 32.50                                                          & 66.88                                                          & 56.22                                                          \\ \midrule
\textit{B16-CC}                       &               &                     &              &                   &                  &                &                                                                &                                                                & \multicolumn{1}{l}{}                                           & \multicolumn{1}{l}{}                                           \\
Random                                & 81.12         & 22.35               & 48.75        & 30.42             & 86.97            & 51.00          & 48.59                                                          & 40.85                                                          & 71.06                                                          & 64.12                                                          \\
TracIn                                & 75.25         & 17.76               & 34.20        & 22.59             & 81.33            & 36.29          & 41.97                                                          & 31.20                                                          & 62.65                                                          & 55.68                                                          \\
\tableLineColor \textbf{CHIPS (ours)} & 75.96         & 18.10               & 35.33        & 23.49             & 82.34            & 40.96          & 43.88                                                          & 32.20                                                          & 64.15                                                          & 57.34                                                          \\ \midrule
\textit{L14-400M}                     &               &                     &              &                   &                  &                &                                                                &                                                                & \multicolumn{1}{l}{}                                           & \multicolumn{1}{l}{}                                           \\
Random                                & 82.93         & 30.34               & 41.36        & 38.97             & 89.35            & 48.79          & 65.73                                                          & 29.35                                                          & 75.23                                                          & 68.72                                                          \\
TracIn                                & 78.54         & 26.27               & 36.57        & 35.25             & 87.81            & 39.90          & 61.49                                                          & 24.30                                                          & 69.84                                                          & 63.35                                                          \\
\tableLineColor \textbf{CHIPS (ours)} & 80.26         & 26.73               & 39.45        & 34.71             & 88.02            & 41.71          & 62.24                                                          & 25.20                                                          & 70.88                                                          & 64.61                                                          \\ \midrule
\textit{L14-CC}                       &               &                     &              &                   &                  &                &                                                                &                                                                & \multicolumn{1}{l}{}                                           & \multicolumn{1}{l}{}                                           \\
Random                                & 88.02         & 33.86               & 54.99        & 42.18             & 93.31            & 56.33          & 71.47                                                          & 29.05                                                          & 78.28                                                          & 71.32                                                          \\
TracIn                                & 83.62         & 29.36               & 46.68        & 41.19             & 90.38            & 48.76          & 66.41                                                          & 24.65                                                          & 73.51                                                          & 66.43                                                          \\
\tableLineColor \textbf{CHIPS (ours)} & 84.65         & 29.78               & 47.94        & 40.98             & 91.25            & 46.98          & 68.07                                                          & 25.15                                                          & 74.15                                                          & 67.55                                                          \\ \midrule
\textit{H14-CC}                       &               &                     &              &                   &                  &                &                                                                &                                                                & \multicolumn{1}{l}{}                                           & \multicolumn{1}{l}{}                                           \\
Random                                & 89.64         & 37.46               & 52.49        & 51.43             & 93.87            & 58.34          & 74.92                                                          & 29.45                                                          & 79.75                                                          & 73.69                                                          \\
TracIn                                & 88.73         & 35.35               & 53.19        & 45.15             & 92.81            & 57.78          & 73.16                                                          & 24.05                                                          & 76.19                                                          & 69.91                                                          \\
\tableLineColor \textbf{CHIPS (ours)} & 88.30         & 35.64               & 55.46        & 47.58             & 93.09            & 57.86          & 74.25                                                          & 24.85                                                          & 76.63                                                          & 70.39                                                          \\ \bottomrule
\end{tabular}
\caption{Full general-domain results (Part 1/4) of the generalization experiment.}
\label{tab:full_generalization_general_1}
\end{table*}

\begin{table*}[tp]
\centering
\small
\begin{tabular}{lccccccccc}
\toprule
\multicolumn{1}{c}{\textbf{Model}}    & \textbf{MNIST} & \textbf{\begin{tabular}[c]{@{}c@{}}Rendered\\ SST2\end{tabular}} & \textbf{STL10} & \textbf{Sun397} & \textbf{\begin{tabular}[c]{@{}c@{}}Sun397\\ Official\end{tabular}} & \textbf{VOC} & \textbf{Caltech101} & \textbf{CIFAR10} & \textbf{CIFAR100}    \\ \midrule
\textit{B32-400M}                     &                &                                                                  &                &                 &                                                                    &              &                     &                  & \multicolumn{1}{l}{} \\
Random                                & 39.99          & 54.81                                                            & 96.11          & 63.75           & 64.66                                                              & 66.37        & 84.21               & 91.19            & 67.92                \\
TracIn                                & 35.04          & 53.76                                                            & 94.42          & 59.56           & 60.19                                                              & 68.32        & 82.86               & 88.24            & 63.02                \\
\tableLineColor \textbf{CHIPS (ours)} & 38.52          & 55.13                                                            & 95.08          & 59.43           & 60.70                                                              & 66.67        & 82.70               & 88.68            & 63.70                \\ \midrule
\textit{B32-CC}                       &                &                                                                  &                &                 &                                                                    &              &                     &                  & \multicolumn{1}{l}{} \\
Random                                & 43.11          & 53.87                                                            & 96.42          & 66.50           & 66.95                                                              & 76.30        & 85.80               & 95.17            & 77.92                \\
TracIn                                & 38.90          & 52.83                                                            & 95.40          & 63.29           & 64.03                                                              & 75.01        & 86.06               & 94.27            & 76.24                \\
\tableLineColor \textbf{CHIPS (ours)} & 36.16          & 51.62                                                            & 95.71          & 63.95           & 64.63                                                              & 75.73        & 84.62               & 94.01            & 75.47                \\ \midrule
\textit{B16-400M}                     &                &                                                                  &                &                 &                                                                    &              &                     &                  & \multicolumn{1}{l}{} \\
Random                                & 45.07          & 55.85                                                            & 97.05          & 66.82           & 67.89                                                              & 69.95        & 85.55               & 91.94            & 67.61                \\
TracIn                                & 39.09          & 52.94                                                            & 95.24          & 57.33           & 59.55                                                              & 68.56        & 82.17               & 82.60            & 54.60                \\
\tableLineColor \textbf{CHIPS (ours)} & 38.82          & 57.84                                                            & 94.91          & 58.00           & 57.88                                                              & 68.66        & 82.48               & 80.76            & 55.33                \\ \midrule
\textit{B16-CC}                       &                &                                                                  &                &                 &                                                                    &              &                     &                  & \multicolumn{1}{l}{} \\
Random                                & 59.17          & 54.75                                                            & 98.35          & 68.43           & 69.29                                                              & 78.17        & 84.54               & 96.09            & 79.97                \\
TracIn                                & 54.70          & 53.27                                                            & 97.12          & 58.73           & 60.19                                                              & 69.29        & 82.28               & 92.97            & 72.27                \\
\tableLineColor \textbf{CHIPS (ours)} & 55.66          & 54.42                                                            & 97.61          & 61.01           & 62.31                                                              & 70.19        & 82.68               & 93.50            & 73.19                \\ \midrule
\textit{L14-400M}                     &                &                                                                  &                &                 &                                                                    &              &                     &                  & \multicolumn{1}{l}{} \\
Random                                & 52.89          & 64.52                                                            & 99.20          & 71.34           & 72.54                                                              & 74.47        & 85.98               & 95.97            & 76.67                \\
TracIn                                & 58.40          & 58.26                                                            & 97.34          & 62.08           & 63.28                                                              & 62.04        & 82.28               & 89.52            & 70.96                \\
\tableLineColor \textbf{CHIPS (ours)} & 59.23          & 62.93                                                            & 97.74          & 61.34           & 62.34                                                              & 62.95        & 82.83               & 91.16            & 70.80                \\ \midrule
\textit{L14-CC}                       &                &                                                                  &                &                 &                                                                    &              &                     &                  & \multicolumn{1}{l}{} \\
Random                                & 60.93          & 68.26                                                            & 99.25          & 72.03           & 73.77                                                              & 80.58        & 88.15               & 97.60            & 84.84                \\
TracIn                                & 57.61          & 56.73                                                            & 98.61          & 66.52           & 67.98                                                              & 74.63        & 84.45               & 95.20            & 79.35                \\
\tableLineColor \textbf{CHIPS (ours)} & 64.90          & 64.80                                                            & 98.84          & 67.41           & 68.64                                                              & 75.45        & 84.42               & 95.46            & 78.86                \\ \midrule
\textit{H14-CC}                       &                &                                                                  &                &                 &                                                                    &              &                     &                  & \multicolumn{1}{l}{} \\
Random                                & 70.45          & 70.68                                                            & 99.49          & 73.01           & 75.34                                                              & 73.45        & 87.54               & 98.10            & 86.61                \\
TracIn                                & 33.99          & 66.89                                                            & 99.19          & 68.30           & 69.37                                                              & 67.22        & 83.43               & 97.15            & 84.05                \\
\tableLineColor \textbf{CHIPS (ours)} & 20.47          & 63.87                                                            & 99.20          & 68.98           & 70.47                                                              & 69.35        & 85.18               & 97.09            & 84.10                \\ \bottomrule
\end{tabular}
\caption{Full general-domain results (Part 2/4) of the generalization experiment.}
\label{tab:full_generalization_general_2}
\end{table*}

\begin{table*}[tp]
\centering
\small
\begin{tabular}{lccccccc}
\toprule
\multicolumn{1}{c}{\textbf{Model}}    & \textbf{\begin{tabular}[c]{@{}c@{}}CLEVR\\ Closest\end{tabular}} & \textbf{\begin{tabular}[c]{@{}c@{}}CLEVR\\ Count\end{tabular}} & \textbf{DMLAB} & \textbf{DTD} & \textbf{Eurosat} & \textbf{Flowers} & \textbf{KITTI} \\ \midrule
\textit{B32-400M}                     &                                                                 &                                                                &                &              &                  &                  &                \\
Random                                & 21.01                                                           & 23.33                                                          & 19.06          & 50.32        & 50.06            & 70.06            & 32.63          \\
TracIn                                & 21.85                                                           & 22.33                                                          & 16.89          & 47.55        & 47.96            & 66.60            & 24.61          \\
\tableLineColor \textbf{CHIPS (ours)} & 22.40                                                           & 23.06                                                          & 15.15          & 48.30        & 45.43            & 69.15            & 32.07          \\ \midrule
\textit{B32-CC}                       &                                                                 &                                                                &                &              &                  &                  &                \\
Random                                & 23.78                                                           & 21.09                                                          & 12.20          & 55.64        & 49.52            & 69.07            & 15.33          \\
TracIn                                & 22.59                                                           & 19.37                                                          & 12.34          & 53.03        & 45.76            & 63.77            & 15.61          \\
\tableLineColor \textbf{CHIPS (ours)} & 22.55                                                           & 20.12                                                          & 13.17          & 54.52        & 43.00            & 66.11            & 19.69          \\ \midrule
\textit{B16-400M}                     &                                                                 &                                                                &                &              &                  &                  &                \\
Random                                & 22.57                                                           & 29.39                                                          & 12.28          & 53.56        & 53.31            & 72.68            & 31.08          \\
TracIn                                & 21.39                                                           & 22.96                                                          & 14.83          & 45.80        & 40.17            & 66.08            & 17.02          \\
\tableLineColor \textbf{CHIPS (ours)} & 22.59                                                           & 21.31                                                          & 14.92          & 48.86        & 40.27            & 66.29            & 19.83          \\ \midrule
\textit{B16-CC}                       &                                                                 &                                                                &                &              &                  &                  &                \\
Random                                & 22.55                                                           & 28.51                                                          & 21.39          & 61.65        & 52.61            & 74.94            & 27.00          \\
TracIn                                & 22.69                                                           & 25.08                                                          & 19.25          & 51.60        & 53.44            & 62.06            & 22.22          \\
\tableLineColor \textbf{CHIPS (ours)} & 22.72                                                           & 26.79                                                          & 19.38          & 52.87        & 55.43            & 65.07            & 22.08          \\ \midrule
\textit{L14-400M}                     &                                                                 &                                                                &                &              &                  &                  &                \\
Random                                & 21.07                                                           & 34.80                                                          & 20.12          & 59.84        & 62.15            & 77.90            & 28.41          \\
TracIn                                & 22.93                                                           & 27.41                                                          & 19.26          & 51.97        & 50.56            & 73.78            & 30.38          \\
\tableLineColor \textbf{CHIPS (ours)} & 23.67                                                           & 27.62                                                          & 17.88          & 53.72        & 50.19            & 75.05            & 29.54          \\ \midrule
\textit{L14-CC}                       &                                                                 &                                                                &                &              &                  &                  &                \\
Random                                & 22.54                                                           & 28.61                                                          & 16.64          & 66.38        & 66.76            & 81.53            & 24.33          \\
TracIn                                & 22.54                                                           & 27.45                                                          & 20.95          & 59.89        & 56.39            & 78.09            & 25.04          \\
\tableLineColor \textbf{CHIPS (ours)} & 22.54                                                           & 28.74                                                          & 18.50          & 60.37        & 53.33            & 78.60            & 32.77          \\ \midrule
\textit{H14-CC}                       &                                                                 &                                                                &                &              &                  &                  &                \\
Random                                & 10.25                                                           & 20.78                                                          & 14.86          & 68.99        & 69.37            & 83.72            & 27.14          \\
TracIn                                & 21.85                                                           & 23.17                                                          & 13.62          & 63.24        & 62.37            & 81.18            & 22.93          \\
\tableLineColor \textbf{CHIPS (ours)} & 20.85                                                           & 21.61                                                          & 13.02          & 66.17        & 61.69            & 81.43            & 25.32          \\ \bottomrule
\end{tabular}
\caption{Full general-domain results (Part 3/4) of the generalization experiment.}
\label{tab:full_generalization_general_3}
\end{table*}

\begin{table*}[tp]
\centering
\small
\begin{tabular}{lccccc}
\toprule
\multicolumn{1}{c}{\textbf{Model}}    & \textbf{Pets} & \textbf{RESISC45} & \textbf{\begin{tabular}[c]{@{}c@{}}Smallnorb\\ Azimuth\end{tabular}} & \textbf{\begin{tabular}[c]{@{}c@{}}Smallnorb\\ Elevation\end{tabular}} & \textbf{SVHN} \\ \midrule
\textit{B32-400M}                     &               &                   &                                                                      &                                                                        &               \\
Random                                & 85.75         & 57.11             & 5.35                                                                 & 12.03                                                                  & 23.40         \\
TracIn                                & 84.79         & 55.87             & 5.74                                                                 & 11.45                                                                  & 27.67         \\
\tableLineColor \textbf{CHIPS (ours)} & 86.67         & 56.70             & 5.24                                                                 & 12.04                                                                  & 25.37         \\ \midrule
\textit{B32-CC}                       &               &                   &                                                                      &                                                                        &               \\
Random                                & 88.74         & 59.37             & 5.70                                                                 & 12.20                                                                  & 18.77         \\
TracIn                                & 87.84         & 56.25             & 5.60                                                                 & 11.65                                                                  & 22.40         \\
\tableLineColor \textbf{CHIPS (ours)} & 88.47         & 58.83             & 6.09                                                                 & 12.47                                                                  & 21.03         \\ \midrule
\textit{B16-400M}                     &               &                   &                                                                      &                                                                        &               \\
Random                                & 90.52         & 64.08             & 5.93                                                                 & 10.88                                                                  & 18.41         \\
TracIn                                & 88.23         & 60.14             & 5.47                                                                 & 11.37                                                                  & 24.96         \\
\tableLineColor \textbf{CHIPS (ours)} & 86.97         & 58.44             & 5.55                                                                 & 11.82                                                                  & 25.23         \\ \midrule
\textit{B16-CC}                       &               &                   &                                                                      &                                                                        &               \\
Random                                & 91.09         & 66.79             & 4.95                                                                 & 10.77                                                                  & 37.11         \\
TracIn                                & 87.57         & 59.35             & 5.33                                                                 & 11.01                                                                  & 36.07         \\
\tableLineColor \textbf{CHIPS (ours)} & 87.82         & 60.21             & 5.28                                                                 & 10.77                                                                  & 36.69         \\ \midrule
\textit{L14-400M}                     &               &                   &                                                                      &                                                                        &               \\
Random                                & 92.89         & 68.68             & 5.51                                                                 & 11.04                                                                  & 22.41         \\
TracIn                                & 90.30         & 63.06             & 5.93                                                                 & 11.94                                                                  & 27.65         \\
\tableLineColor \textbf{CHIPS (ours)} & 91.41         & 63.32             & 5.85                                                                 & 12.17                                                                  & 27.55         \\ \midrule
\textit{L14-CC}                       &               &                   &                                                                      &                                                                        &               \\
Random                                & 94.06         & 75.13             & 4.85                                                                 & 11.13                                                                  & 46.73         \\
TracIn                                & 90.81         & 66.89             & 5.40                                                                 & 11.11                                                                  & 49.62         \\
\tableLineColor \textbf{CHIPS (ours)} & 92.37         & 66.44             & 5.21                                                                 & 11.56                                                                  & 50.27         \\ \midrule
\textit{H14-CC}                       &               &                   &                                                                      &                                                                        &               \\
Random                                & 95.53         & 72.56             & 6.30                                                                 & 12.30                                                                  & 44.79         \\
TracIn                                & 94.19         & 68.65             & 5.82                                                                 & 12.40                                                                  & 50.89         \\
\tableLineColor \textbf{CHIPS (ours)} & 94.49         & 68.44             & 6.44                                                                 & 12.21                                                                  & 50.99         \\ \bottomrule
\end{tabular}
\caption{Full general-domain results (Part 4/4) of the generalization experiment.}
\label{tab:full_generalization_general_4}
\end{table*}

\subsection{Ablation Experiment}

The full results of the main experiment are detailed in Tab.~\ref{tab:full_ablation_medical_1} and Tab.~\ref{tab:full_ablation_medical_2} for the medical domain, and in Tab.~\ref{tab:full_ablation_general_1} through Tab.~\ref{tab:full_ablation_general_4} for the general domain.

\begin{table*}[tp]
\centering
\small
\begin{tabular}{lcccccccc}
\toprule
\multicolumn{1}{c}{\textbf{Model}}    & \textbf{Diabetic} & \textbf{PCAM} & \textbf{LC25000} & \textbf{Pollen} & \textbf{Amyloid CAA} & \textbf{Amyloid Diffuse} & \textbf{BloodMNIST} & \textbf{ChestMNIST} \\ \midrule
\textit{r=10\%}                       &                   &               &                  &                 &                      &                          &                     &                     \\
Alignment-only                        & 59.51             & 57.64         & 39.41            & 12.76           & 88.69                & 13.17                    & 23.62               & 3.71                \\
Alignment-Margin                      & 59.56             & 57.65         & 40.21            & 11.69           & 87.51                & 13.94                    & 23.56               & 3.26                \\
\tableLineColor \textbf{CHIPS (ours)} & 68.35             & 63.66         & 40.51            & 25.96           & 88.06                & 12.38                    & 21.43               & 6.24                \\ \midrule
\textit{r=20\%}                       &                   &               &                  &                 &                      &                          &                     &                     \\
Alignment-only                        & 72.41             & 68.04         & 46.98            & 30.54           & 55.27                & 21.32                    & 20.56               & 12.24               \\
Alignment-Margin                      & 72.02             & 68.32         & 46.87            & 30.10           & 58.24                & 21.65                    & 21.34               & 12.34               \\
\tableLineColor \textbf{CHIPS (ours)} & 72.50             & 68.67         & 48.07            & 30.36           & 53.58                & 23.50                    & 22.49               & 13.97               \\ \midrule
\textit{r=30\%}                       &                   &               &                  &                 &                      &                          &                     &                     \\
Alignment-only                        & 71.30             & 72.76         & 43.79            & 32.14           & 60.26                & 22.71                    & 21.88               & 11.77               \\
Alignment-Margin                      & 71.24             & 70.43         & 48.16            & 32.42           & 61.79                & 25.12                    & 23.44               & 13.19               \\
\tableLineColor \textbf{CHIPS (ours)} & 73.42             & 73.66         & 49.88            & 36.34           & 67.90                & 25.15                    & 25.02               & 14.39               \\ \bottomrule
\end{tabular}
\caption{Full medical-domain results (Part 1/2) of the ablation experiment.}
\label{tab:full_ablation_medical_1}
\end{table*}

\begin{table*}[tp]
\centering
\small
\begin{tabular}{lccccccccc}
\toprule
\multicolumn{1}{c}{\textbf{Model}}    & \textbf{ChestXray14} & \textbf{Derma} & \textbf{Oct} & \textbf{OrganA} & \textbf{OrganC} & \textbf{OrganS} & \textbf{Path} & \textbf{Retina} & \multicolumn{1}{l}{\textbf{Tissue}} \\ \midrule
\textit{r=10\%}                       &                      &                &              &                 &                 &                 &               &                 & \multicolumn{1}{l}{}                \\
Alignment-only                        & 21.49                & 12.07          & 20.20        & 11.65           & 10.08           & 10.81           & 27.28         & 27.00           & 4.86                                \\
Alignment-Margin                      & 21.02                & 12.17          & 19.60        & 12.10           & 10.00           & 10.28           & 26.70         & 29.25           & 4.99                                \\
\tableLineColor \textbf{CHIPS (ours)} & 16.63                & 11.87          & 18.20        & 12.03           & 9.64            & 10.71           & 23.90         & 30.25           & 5.29                                \\ \midrule
\textit{r=20\%}                       &                      &                &              &                 &                 &                 &               &                 & \multicolumn{1}{l}{}                \\
Alignment-only                        & 9.57                 & 11.22          & 20.80        & 14.10           & 13.86           & 15.24           & 24.57         & 32.00           & 5.99                                \\
Alignment-Margin                      & 9.07                 & 11.12          & 20.80        & 14.53           & 14.17           & 15.04           & 24.55         & 35.75           & 5.39                                \\
\tableLineColor \textbf{CHIPS (ours)} & 8.96                 & 11.87          & 21.40        & 13.50           & 13.80           & 14.87           & 23.20         & 38.50           & 5.92                                \\ \midrule
\textit{r=30\%}                       &                      &                &              &                 &                 &                 &               &                 & \multicolumn{1}{l}{}                \\
Alignment-only                        & 8.28                 & 11.22          & 21.80        & 13.62           & 12.99           & 14.87           & 21.56         & 30.75           & 7.46                                \\
Alignment-Margin                      & 10.49                & 11.32          & 20.50        & 14.35           & 13.38           & 12.69           & 24.08         & 30.50           & 6.40                                \\
\tableLineColor \textbf{CHIPS (ours)} & 10.65                & 11.37          & 21.30        & 13.13           & 14.28           & 15.33           & 24.43         & 30.25           & 7.69                                \\ \bottomrule
\end{tabular}
\caption{Full medical-domain results (Part 2/2) of the ablation experiment.}
\label{tab:full_ablation_medical_2}
\end{table*}

\begin{table*}[tp]
\centering
\small
\begin{tabular}{lcccccccccc}
\toprule
\multicolumn{1}{c}{\textbf{Model}}    & \textbf{Cars} & \textbf{Country211} & \textbf{FER} & \textbf{Aircraft} & \textbf{Food101} & \textbf{GSTRB} & \textbf{\begin{tabular}[c]{@{}c@{}}Imagenet\\ -A\end{tabular}} & \textbf{\begin{tabular}[c]{@{}c@{}}Imagenet\\ -O\end{tabular}} & \textbf{\begin{tabular}[c]{@{}c@{}}Imagenet\\ 1k\end{tabular}} & \textbf{\begin{tabular}[c]{@{}c@{}}Imagenet\\ v2\end{tabular}} \\ \midrule
\textit{r=10\%}                       &               &                     &              &                   &                  &                &                                                                &                                                                &                                                                &                                                                \\
Alignment-only                        & 69.99         & 19.50               & 40.85        & 25.59             & 82.55            & 31.56          & 42.47                                                          & 32.95                                                          & 64.84                                                          & 56.63                                                          \\
Alignment-Margin                      & 70.12         & 21.67               & 40.81        & 25.92             & 82.60            & 31.62          & 42.17                                                          & 32.45                                                          & 65.01                                                          & 56.78                                                          \\
\tableLineColor \textbf{CHIPS (ours)} & 68.40         & 21.47               & 40.69        & 25.32             & 82.60            & 31.02          & 42.36                                                          & 32.50                                                          & 66.88                                                          & 56.22                                                          \\ \midrule
\textit{r=20\%}                       &               &                     &              &                   &                  &                &                                                                &                                                                &                                                                &                                                                \\
Alignment-only                        & 68.97         & 18.80               & 39.19        & 24.60             & 80.06            & 31.50          & 40.11                                                          & 31.60                                                          & 62.95                                                          & 55.05                                                          \\
Alignment-Margin                      & 68.95         & 19.59               & 39.23        & 24.51             & 80.03            & 31.92          & 40.15                                                          & 31.60                                                          & 62.89                                                          & 54.99                                                          \\
\tableLineColor \textbf{CHIPS (ours)} & 69.32         & 19.54               & 39.40        & 24.36             & 80.05            & 31.06          & 40.33                                                          & 32.10                                                          & 62.97                                                          & 55.01                                                          \\ \midrule
\textit{r=30\%}                       &               &                     &              &                   &                  &                &                                                                &                                                                &                                                                &                                                                \\
Alignment-only                        & 68.86         & 18.41               & 38.35        & 24.78             & 79.45            & 30.78          & 39.45                                                          & 31.60                                                          & 62.63                                                          & 54.52                                                          \\
Alignment-Margin                      & 68.74         & 18.48               & 42.28        & 24.18             & 79.93            & 30.43          & 39.83                                                          & 31.40                                                          & 62.56                                                          & 54.74                                                          \\
\tableLineColor \textbf{CHIPS (ours)} & 68.30         & 18.39               & 37.89        & 23.46             & 79.27            & 30.36          & 38.96                                                          & 31.40                                                          & 62.33                                                          & 54.45                                                          \\ \bottomrule
\end{tabular}
\caption{Full general-domain results (Part 1/4) of the ablation experiment.}
\label{tab:full_ablation_general_1}
\end{table*}

\begin{table*}[tp]
\centering
\small
\begin{tabular}{lccccccccc}
\toprule
\multicolumn{1}{c}{\textbf{Model}}    & \textbf{MNIST} & \textbf{\begin{tabular}[c]{@{}c@{}}Rendered\\ SST2\end{tabular}} & \textbf{STL10} & \textbf{Sun397} & \textbf{\begin{tabular}[c]{@{}c@{}}Sun397\\ Official\end{tabular}} & \textbf{VOC} & \textbf{Caltech101} & \textbf{CIFAR10} & \textbf{CIFAR100}    \\ \midrule
\textit{r=10\%}                       &               &                     &              &                   &                  &                &                                                                &                                                                &                                                                \\
Alignment-only                        & 45.04         & 56.78               & 95.42        & 58.44             & 60.47            & 66.18          & 83.27                                                          & 82.55                                                          & 55.19                                                          \\
Alignment-Margin                      & 44.47         & 57.44               & 95.34        & 58.34             & 60.40            & 66.66          & 82.93                                                          & 82.57                                                          & 55.34                                                          \\
\tableLineColor \textbf{CHIPS (ours)} & 38.82         & 57.84               & 94.91        & 58.00             & 57.88            & 68.66          & 82.48                                                          & 80.76                                                          & 55.33                                                          \\ \midrule
\textit{r=20\%}                       &               &                     &              &                   &                  &                &                                                                &                                                                &                                                                \\
Alignment-only                        & 43.42         & 50.74               & 95.19        & 56.52             & 58.73            & 66.31          & 81.38                                                          & 84.85                                                          & 57.73                                                          \\
Alignment-Margin                      & 42.65         & 50.36               & 94.99        & 56.47             & 58.76            & 65.93          & 81.10                                                          & 85.76                                                          & 59.11                                                          \\
\tableLineColor \textbf{CHIPS (ours)} & 42.91         & 50.63               & 95.42        & 55.94             & 58.51            & 65.89          & 81.38                                                          & 85.72                                                          & 59.00                                                          \\ \midrule
\textit{r=30\%}                       &               &                     &              &                   &                  &                &                                                                &                                                                &                                                                \\
Alignment-only                        & 38.92         & 50.14               & 95.21        & 56.16             & 58.28            & 65.77          & 80.12                                                          & 85.44                                                          & 57.97                                                          \\
Alignment-Margin                      & 35.97         & 50.03               & 95.17        & 55.95             & 58.43            & 62.85          & 80.30                                                          & 84.61                                                          & 56.85                                                          \\
\tableLineColor \textbf{CHIPS (ours)} & 34.25         & 49.92               & 95.36        & 55.69             & 57.91            & 64.36          & 80.79                                                          & 84.61                                                          & 57.64                                                          \\ \bottomrule
\end{tabular}
\caption{Full general-domain results (Part 2/4) of the ablation experiment.}
\label{tab:full_ablation_general_2}
\end{table*}

\begin{table*}[tp]
\centering
\small
\begin{tabular}{lccccccc}
\toprule
\multicolumn{1}{c}{\textbf{Model}}    & \textbf{\begin{tabular}[c]{@{}c@{}}CLEVR\\ Closest\end{tabular}} & \textbf{\begin{tabular}[c]{@{}c@{}}CLEVR\\ Count\end{tabular}} & \textbf{DMLAB} & \textbf{DTD} & \textbf{Eurosat} & \textbf{Flowers} & \textbf{KITTI} \\ \midrule
\textit{r=10\%}                       &               &                     &              &                   &                  &                &                                                                \\
Alignment-only                        & 22.62         & 21.30               & 13.90        & 48.67             & 38.98            & 68.11          & 23.49                                                          \\
Alignment-Margin                      & 22.58         & 21.41               & 14.14        & 49.36             & 39.11            & 68.01          & 23.21                                                          \\
\tableLineColor \textbf{CHIPS (ours)} & 22.59         & 21.31               & 14.92        & 48.86             & 40.27            & 66.29          & 19.83                                                          \\ \midrule
\textit{r=20\%}                       &               &                     &              &                   &                  &                &                                                                \\
Alignment-only                        & 21.27         & 24.39               & 15.90        & 44.79             & 41.89            & 66.22          & 17.86                                                          \\
Alignment-Margin                      & 21.00         & 25.33               & 15.90        & 44.26             & 42.15            & 66.30          & 17.44                                                          \\
\tableLineColor \textbf{CHIPS (ours)} & 21.52         & 25.25               & 16.21        & 44.47             & 42.89            & 66.53          & 16.46                                                          \\ \midrule
\textit{r=30\%}                       &               &                     &              &                   &                  &                &                                                                \\
Alignment-only                        & 21.24         & 21.56               & 15.40        & 45.21             & 42.76            & 65.03          & 15.61                                                          \\
Alignment-Margin                      & 22.24         & 20.59               & 16.86        & 44.15             & 42.06            & 65.67          & 17.72                                                          \\
\tableLineColor \textbf{CHIPS (ours)} & 21.60         & 20.78               & 15.76        & 44.20             & 42.19            & 65.30          & 14.77                                                          \\ \bottomrule
\end{tabular}
\caption{Full general-domain results (Part 3/4) of the ablation experiment.}
\label{tab:full_ablation_general_3}
\end{table*}

\begin{table*}[tp]
\centering
\small
\begin{tabular}{lccccc}
\toprule
\multicolumn{1}{c}{\textbf{Model}}    & \textbf{Pets} & \textbf{RESISC45} & \textbf{\begin{tabular}[c]{@{}c@{}}Smallnorb\\ Azimuth\end{tabular}} & \textbf{\begin{tabular}[c]{@{}c@{}}Smallnorb\\ Elevation\end{tabular}} & \textbf{SVHN} \\ \midrule
\textit{r=10\%}                       &               &                     &              &                   &                  \\
Alignment-only                        & 87.46         & 60.33               & 5.51         & 11.34             & 26.19            \\
Alignment-Margin                      & 87.35         & 60.21               & 5.47         & 11.32             & 26.04            \\
\tableLineColor \textbf{CHIPS (ours)} & 86.97         & 58.44               & 5.55         & 11.82             & 25.23            \\ \midrule
\textit{r=20\%}                       &               &                     &              &                   &                  \\
Alignment-only                        & 87.35         & 59.48               & 5.62         & 11.06             & 24.34            \\
Alignment-Margin                      & 87.35         & 59.49               & 5.32         & 10.74             & 23.96            \\
\tableLineColor \textbf{CHIPS (ours)} & 87.24         & 59.68               & 5.49         & 11.02             & 24.63            \\ \midrule
\textit{r=30\%}                       &               &                     &              &                   &                  \\
Alignment-only                        & 86.70         & 59.60               & 5.53         & 10.79             & 23.94            \\
Alignment-Margin                      & 87.00         & 59.68               & 5.82         & 10.99             & 23.79            \\
\tableLineColor \textbf{CHIPS (ours)} & 87.08         & 60.00               & 5.84         & 11.18             & 22.45            \\ \bottomrule
\end{tabular}
\caption{Full general-domain results (Part 4/4) of the ablation experiment.}
\label{tab:full_ablation_general_4}
\end{table*}

\subsection{Analysis Experiment}

\begin{itemize}
    \item \textbf{Hyperparameter Analysis:} The analysis of evaluation set size, mixing hyperparameter $\alpha$, and balance hyperparameter $\beta$ is presented in Tab.~\ref{tab:full_analysis_1_medical_1} through Tab.~\ref{tab:full_analysis_1_general_4}.
    \item \textbf{End-point Subspace:} The analysis of the end-point subspace $\vartheta$ is shown in Tab.~\ref{tab:full_analysis_2_medical_1} and Tab.~\ref{tab:full_analysis_2_medical_2}.
    \item \textbf{JL Random Projection:} The analysis of JL random projection is provided in Tab.~\ref{tab:full_analysis_3_medical_1} and Tab.~\ref{tab:full_analysis_3_medical_2}. 
\end{itemize}


\begin{table*}[tp]
\centering
\small
\begin{tabular}{lcccccccc}
\toprule
\multicolumn{1}{c}{\textbf{Model}} & \textbf{Diabetic} & \textbf{PCAM} & \textbf{LC25000} & \textbf{Pollen} & \textbf{Amyloid CAA} & \textbf{Amyloid Diffuse} & \textbf{BloodMNIST} & \textbf{ChestMNIST} \\ \midrule
\textit{Evaluation Set Size}       &                   &               &                  &                 &                      &                          &                     &                     \\
50                                 & 67.33             & 57.66         & 39.68            & 18.10           & 87.22                & 12.69                    & 21.51               & 2.79                \\
100                                & 57.57             & 58.45         & 39.39            & 14.90           & 82.92                & 13.02                    & 24.76               & 2.50                \\
150                                & 54.62             & 56.38         & 40.61            & 11.38           & 85.64                & 13.07                    & 24.06               & 4.61                \\
\tableLineColor 200                                & 68.35             & 63.66         & 40.51            & 25.96           & 88.06                & 12.38                    & 21.43               & 6.24                \\
250                                & 68.07             & 66.49         & 40.55            & 22.20           & 87.99                & 12.63                    & 23.91               & 2.14                \\ \midrule
\textit{$\alpha$}                   &                   &               &                  &                 &                      &                          &                     &                     \\
0.2                                & 52.04             & 57.57         & 40.53            & 13.70           & 79.23                & 13.61                    & 24.55               & 4.28                \\
0.4                                & 51.45             & 57.18         & 41.20            & 13.51           & 74.95                & 14.03                    & 24.90               & 3.28                \\
\tableLineColor 0.6                                & 56.65             & 55.74         & 40.21            & 14.39           & 89.81                & 13.85                    & 25.17               & 2.90                \\
0.8                                & 68.35             & 63.66         & 40.51            & 25.96           & 88.06                & 12.38                    & 21.43               & 6.24                \\
1.0                                & 51.01             & 55.88         & 40.67            & 14.77           & 83.71                & 13.51                    & 23.79               & 3.29                \\ \midrule
\textit{$\beta$}                    &                   &               &                  &                 &                      &                          &                     &                     \\
0                                  & 71.38             & 59.27         & 44.43            & 27.03           & 43.02                & 25.05                    & 18.36               & 3.13                \\
0.25                               & 54.72             & 56.48         & 39.89            & 11.75           & 90.43                & 12.74                    & 22.42               & 3.08                \\
\tableLineColor 0.50                               & 68.35             & 63.66         & 40.51            & 25.96           & 88.06                & 12.38                    & 21.43               & 6.24                \\
0.75                               & 52.53             & 57.27         & 38.93            & 13.83           & 86.47                & 13.79                    & 23.21               & 4.25                \\
1.0                                & 71.53             & 59.11         & 41.89            & 26.40           & 54.35                & 23.21                    & 15.76               & 2.71                \\ \bottomrule
\end{tabular}
\caption{Full medical-domain results (Part 1/2) of the analysis on evaluation set size, $\alpha$, and $\beta$.}
\label{tab:full_analysis_1_medical_1}
\end{table*}

\begin{table*}[tp]
\centering
\small
\begin{tabular}{lccccccccc}
\toprule
\multicolumn{1}{c}{\textbf{Model}} & \textbf{ChestXray14} & \textbf{Derma} & \textbf{Oct} & \textbf{OrganA} & \textbf{OrganC} & \textbf{OrganS} & \textbf{Path} & \textbf{Retina} & \multicolumn{1}{l}{\textbf{Tissue}} \\ \midrule
\textit{Evaluation Set Size}       &                      &                &              &                 &                 &                 &               &                 & \multicolumn{1}{l}{}                \\
50                                 & 16.07                & 12.07          & 21.50        & 8.66            & 8.70            & 10.04           & 26.56         & 35.25           & 5.20                                \\
100                                & 21.20                & 12.17          & 20.70        & 9.23            & 8.87            & 9.49            & 26.24         & 32.75           & 5.87                                \\
150                                & 28.48                & 12.32          & 21.00        & 11.42           & 10.32           & 10.59           & 30.21         & 35.00           & 4.92                                \\
\tableLineColor 200                                & 16.63                & 11.87          & 18.20        & 12.03           & 9.64            & 10.71           & 23.90         & 30.25           & 5.29                                \\
250                                & 16.49                & 12.07          & 19.50        & 12.48           & 10.13           & 9.88            & 24.38         & 30.25           & 5.02                                \\ \midrule
\textit{$\alpha$}                   &                      &                &              &                 &                 &                 &               &                 & \multicolumn{1}{l}{}                \\
0.2                                & 30.17                & 12.12          & 20.30        & 10.93           & 9.54            & 10.30           & 28.23         & 30.00           & 4.92                                \\
0.4                                & 29.49                & 12.17          & 20.70        & 10.98           & 10.08           & 10.82           & 28.08         & 29.25           & 4.82                                \\
\tableLineColor 0.6                                & 28.46                & 12.32          & 20.70        & 11.76           & 10.66           & 11.74           & 29.33         & 33.00           & 4.95                                \\
0.8                                & 16.63                & 11.87          & 18.20        & 12.03           & 9.64            & 10.71           & 23.90         & 30.25           & 5.29                                \\
1.0                                & 28.75                & 12.07          & 20.30        & 11.49           & 10.30           & 11.39           & 29.04         & 30.50           & 5.04                                \\ \midrule
\textit{$\beta$}                    &                      &                &              &                 &                 &                 &               &                 & \multicolumn{1}{l}{}                \\
0                                  & 7.56                 & 11.32          & 19.90        & 10.15           & 11.19           & 12.91           & 24.43         & 37.00           & 5.80                                \\
0.25                               & 19.36                & 12.07          & 20.60        & 11.63           & 10.02           & 10.11           & 26.38         & 35.50           & 5.18                                \\
\tableLineColor 0.50                               & 16.63                & 11.87          & 18.20        & 12.03           & 9.64            & 10.71           & 23.90         & 30.25           & 5.29                                \\
0.75                               & 22.38                & 11.82          & 19.90        & 11.30           & 10.70           & 11.05           & 29.28         & 32.75           & 5.45                                \\
1.0                                & 6.39                 & 10.97          & 19.50        & 10.09           & 11.51           & 12.71           & 24.32         & 38.75           & 5.68                                \\ \bottomrule
\end{tabular}
\caption{Full medical-domain results (Part 2/2) of the analysis on evaluation set size, $\alpha$, and $\beta$.}
\label{tab:full_analysis_1_medical_2}
\end{table*}

\begin{table*}[tp]
\centering
\small
\begin{tabular}{lcccccccccc}
\toprule
\multicolumn{1}{c}{\textbf{Model}} & \textbf{Cars} & \textbf{Country211} & \textbf{FER} & \textbf{Aircraft} & \textbf{Food101} & \textbf{GSTRB} & \textbf{\begin{tabular}[c]{@{}c@{}}Imagenet\\ -A\end{tabular}} & \textbf{\begin{tabular}[c]{@{}c@{}}Imagenet\\ -O\end{tabular}} & \textbf{\begin{tabular}[c]{@{}c@{}}Imagenet\\ 1k\end{tabular}} & \textbf{\begin{tabular}[c]{@{}c@{}}Imagenet\\ v2\end{tabular}} \\ \midrule
\textit{Evaluation Set Size}       &               &                     &              &                   &                  &                &                                                                &                                                                & \multicolumn{1}{l}{}                                           & \multicolumn{1}{l}{}                                           \\
50                                 & 70.15         & 19.38               & 40.58        & 25.65             & 82.56            & 31.46          & 42.68                                                          & 33.05                                                          & 65.04                                                          & 57.01                                                          \\
100                                & 69.79         & 19.40               & 41.70        & 25.26             & 82.52            & 33.02          & 42.25                                                          & 32.10                                                          & 65.15                                                          & 57.00                                                          \\
150                                & 69.48         & 19.48               & 41.07        & 24.75             & 82.70            & 32.54          & 42.09                                                          & 33.40                                                          & 64.97                                                          & 56.96                                                          \\
\tableLineColor 200                                & 68.40         & 21.47               & 40.69        & 25.32             & 82.60            & 31.02          & 42.36                                                          & 32.50                                                          & 66.88                                                          & 56.22                                                          \\
250                                & 67.38         & 19.41               & 40.35        & 25.08             & 82.91            & 30.02          & 42.45                                                          & 32.25                                                          & 64.91                                                          & 56.80                                                          \\ \midrule
\textit{$\alpha$}                   &               &                     &              &                   &                  &                &                                                                &                                                                & \multicolumn{1}{l}{}                                           & \multicolumn{1}{l}{}                                           \\
0.2                                & 69.59         & 19.26               & 41.71        & 24.99             & 82.4             & 32.38          & 42.72                                                          & 32.35                                                          & 64.97                                                          & 56.81                                                          \\
0.4                                & 69.82         & 19.34               & 40.97        & 25.26             & 82.65            & 32.07          & 42.71                                                          & 32.40                                                          & 65.12                                                          & 57.15                                                          \\
\tableLineColor 0.6                                & 68.40         & 21.47               & 40.69        & 25.32             & 82.60            & 31.02          & 42.36                                                          & 32.50                                                          & 66.88                                                          & 56.22                                                          \\
0.8                                & 69.88         & 19.40               & 40.92        & 25.77             & 82.56            & 31.95          & 42.85                                                          & 31.70                                                          & 65.08                                                          & 57.13                                                          \\
1.0                                & 70.07         & 19.50               & 41.31        & 25.56             & 82.63            & 31.50          & 43.00                                                          & 32.15                                                          & 65.12                                                          & 57.08                                                          \\ \midrule
\textit{$\beta$}                    &               &                     &              &                   &                  &                &                                                                &                                                                & \multicolumn{1}{l}{}                                           & \multicolumn{1}{l}{}                                           \\
0                                  & 71.00         & 19.63               & 42.77        & 25.77             & 82.05            & 32.64          & 42.29                                                          & 31.70                                                          & 64.59                                                          & 56.71                                                          \\
0.25                               & 69.98         & 19.42               & 40.14        & 25.26             & 82.51            & 31.57          & 41.99                                                          & 32.30                                                          & 64.99                                                          & 56.68                                                          \\
\tableLineColor 0.50                               & 68.40         & 21.47               & 40.69        & 25.32             & 82.60            & 31.02          & 42.36                                                          & 32.50                                                          & 66.88                                                          & 56.22                                                          \\
0.75                               & 70.54         & 19.26               & 41.39        & 25.26             & 82.66            & 33.66          & 42.52                                                          & 32.10                                                          & 65.11                                                          & 57.02                                                          \\
1.0                                & 70.39         & 19.41               & 41.84        & 26.76             & 81.91            & 32.07          & 41.99                                                          & 32.25                                                          & 64.63                                                          & 56.26                                                          \\ \bottomrule
\end{tabular}
\caption{Full general-domain results (Part 1/4) of the analysis on evaluation set size, $\alpha$, and $\beta$.}
\label{tab:full_analysis_1_general_1}
\end{table*}

\begin{table*}[tp]
\centering
\small
\begin{tabular}{lccccccccc}
\toprule
\multicolumn{1}{c}{\textbf{Model}} & \textbf{MNIST} & \textbf{\begin{tabular}[c]{@{}c@{}}Rendered\\ SST2\end{tabular}} & \textbf{STL10} & \textbf{Sun397} & \textbf{\begin{tabular}[c]{@{}c@{}}Sun397\\ Official\end{tabular}} & \textbf{VOC} & \textbf{Caltech101} & \textbf{CIFAR10} & \textbf{CIFAR100}    \\ \midrule
\textit{Evaluation Set Size}       &                &                                                                  &                &                 &                                                                    &              &                     &                  & \multicolumn{1}{l}{} \\
50                                 & 47.03          & 57.77                                                            & 95.29          & 58.61           & 60.61                                                              & 66.67        & 82.74               & 82.78            & 54.91                \\
100                                & 46.74          & 56.84                                                            & 95.40          & 58.71           & 60.62                                                              & 65.63        & 83.47               & 82.48            & 54.73                \\
150                                & 43.58          & 57.00                                                            & 95.31          & 59.10           & 60.79                                                              & 64.98        & 83.57               & 82.77            & 55.25                \\
\tableLineColor 200                & 38.82          & 57.84                                                            & 94.91          & 58.00           & 57.88                                                              & 68.66        & 82.48               & 80.76            & 55.33                \\
250                                & 42.26          & 58.26                                                            & 95.09          & 57.58           & 60.68                                                              & 63.62        & 83.25               & 80.11            & 55.04                \\ \midrule
\textit{$\alpha$}                  &                &                                                                  &                &                 &                                                                    &              &                     &                  & \multicolumn{1}{l}{} \\
0.2                                & 45.55          & 57.44                                                            & 95.31          & 58.91           & 60.49                                                              & 66.91        & 83.06               & 82.61            & 55.31                \\
0.4                                & 45.44          & 57.83                                                            & 95.40          & 58.78           & 60.65                                                              & 67.05        & 83.11               & 83.08            & 55.68                \\
\tableLineColor 0.6                & 38.82          & 57.84                                                            & 94.91          & 58.00           & 57.88                                                              & 68.66        & 82.48               & 80.76            & 55.33                \\
0.8                                & 45.82          & 57.33                                                            & 95.39          & 58.92           & 60.55                                                              & 66.19        & 83.27               & 82.96            & 55.15                \\
1.0                                & 47.11          & 57.66                                                            & 95.51          & 58.73           & 60.53                                                              & 66.57        & 83.27               & 82.95            & 55.21                \\ \midrule
\textit{$\beta$}                   &                &                                                                  &                &                 &                                                                    &              &                     &                  & \multicolumn{1}{l}{} \\
0                                  & 42.85          & 52.33                                                            & 95.51          & 57.74           & 60.47                                                              & 66.73        & 82.65               & 85.81            & 58.97                \\
0.25                               & 44.89          & 58.26                                                            & 94.31          & 58.89           & 60.89                                                              & 66.29        & 82.97               & 82.37            & 54.13                \\
\tableLineColor 0.50               & 38.82          & 57.84                                                            & 94.91          & 58.00           & 57.88                                                              & 68.66        & 82.48               & 80.76            & 55.33                \\
0.75                               & 42.92          & 58.37                                                            & 95.34          & 58.86           & 60.79                                                              & 66.25        & 83.58               & 84.02            & 56.88                \\
1.0                                & 40.30          & 51.89                                                            & 95.14          & 57.72           & 60.44                                                              & 66.75        & 82.51               & 85.63            & 59.00                \\ \bottomrule
\end{tabular}
\caption{Full general-domain results (Part 2/4) of the analysis on evaluation set size, $\alpha$, and $\beta$.}
\label{tab:full_analysis_1_general_2}
\end{table*}

\begin{table*}[tp]
\centering
\small
\begin{tabular}{lccccccc}
\toprule
\multicolumn{1}{c}{\textbf{Model}} & \textbf{\begin{tabular}[c]{@{}c@{}}CLEVR\\ Closest\end{tabular}} & \textbf{\begin{tabular}[c]{@{}c@{}}CLEVR\\ Count\end{tabular}} & \textbf{DMLAB} & \textbf{DTD} & \textbf{Eurosat} & \textbf{Flowers} & \textbf{KITTI} \\ \midrule
\textit{Evaluation Set Size}       &                                                                 &                                                                &                &              &                  &                  &                \\
50                                 & 22.65                                                           & 21.22                                                          & 14.08          & 48.24        & 39.11            & 68.11            & 23.21          \\
100                                & 22.65                                                           & 21.45                                                          & 14.08          & 48.83        & 38.69            & 68.04            & 22.22          \\
150                                & 22.60                                                           & 22.14                                                          & 14.63          & 48.35        & 39.52            & 67.47            & 23.07          \\
\tableLineColor 200                & 22.59                                                           & 21.31                                                          & 14.92          & 48.86        & 40.27            & 66.29            & 19.83          \\
250                                & 22.44                                                           & 21.19                                                          & 13.76          & 48.35        & 37.52            & 67.05            & 25.32          \\ \midrule
\textit{$\alpha$}                  &                                                                 &                                                                &                &              &                  &                  &                \\
0.2                                & 22.59                                                           & 21.78                                                          & 14.44          & 47.93        & 38.48            & 68.21            & 22.22          \\
0.4                                & 22.59                                                           & 21.87                                                          & 14.42          & 48.51        & 38.28            & 68.14            & 22.78          \\
\tableLineColor 0.6                & 22.59                                                           & 21.31                                                          & 14.92          & 48.86        & 40.27            & 66.29            & 19.83          \\
0.8                                & 22.61                                                           & 21.41                                                          & 14.03          & 48.40        & 38.52            & 67.99            & 21.94          \\
1.0                                & 22.63                                                           & 21.91                                                          & 14.18          & 48.62        & 38.28            & 67.99            & 21.10          \\ \midrule
\textit{$\beta$}                   &                                                                 &                                                                &                &              &                  &                  &                \\
0                                  & 22.10                                                           & 26.25                                                          & 15.35          & 46.91        & 42.26            & 68.35            & 20.53          \\
0.25                               & 22.66                                                           & 21.09                                                          & 13.89          & 48.14        & 39.11            & 68.48            & 23.07          \\
\tableLineColor 0.50               & 22.59                                                           & 21.31                                                          & 14.92          & 48.86        & 40.27            & 66.29            & 19.83          \\
0.75                               & 22.55                                                           & 21.87                                                          & 13.98          & 47.93        & 37.81            & 67.75            & 23.21          \\
1.0                                & 22.06                                                           & 26.15                                                          & 15.76          & 47.23        & 42.78            & 68.24            & 20.53          \\ \bottomrule
\end{tabular}
\caption{Full general-domain results (Part 3/4) of the analysis on evaluation set size, $\alpha$, and $\beta$.}
\label{tab:full_analysis_1_general_3}
\end{table*}

\begin{table*}[tp]
\centering
\small
\begin{tabular}{lccccc}
\toprule
\multicolumn{1}{c}{\textbf{Model}} & \textbf{Pets} & \textbf{RESISC45} & \textbf{\begin{tabular}[c]{@{}c@{}}Smallnorb\\ Azimuth\end{tabular}} & \textbf{\begin{tabular}[c]{@{}c@{}}Smallnorb\\ Elevation\end{tabular}} & \textbf{SVHN} \\ \midrule
\textit{Evaluation Set Size}       &               &                   &                                                                      &                                                                        &               \\
50                                 & 87.63         & 60.05             & 5.44                                                                 & 11.27                                                                  & 25.64         \\
100                                & 88.06         & 60.27             & 5.51                                                                 & 11.51                                                                  & 26.43         \\
150                                & 88.01         & 60.24             & 5.56                                                                 & 11.77                                                                  & 26.67         \\
\tableLineColor 200                & 86.97         & 58.44             & 5.55                                                                 & 11.82                                                                  & 25.23         \\
250                                & 87.49         & 60.38             & 5.42                                                                 & 11.32                                                                  & 26.01         \\ \midrule
\textit{$\alpha$}                  &               &                   &                                                                      &                                                                        &               \\
0.2                                & 87.76         & 60.56             & 5.42                                                                 & 11.41                                                                  & 25.56         \\
0.4                                & 87.46         & 60.46             & 5.79                                                                 & 11.49                                                                  & 25.97         \\
\tableLineColor 0.6                & 86.97         & 58.44             & 5.55                                                                 & 11.82                                                                  & 25.23         \\
0.8                                & 87.74         & 60.62             & 5.28                                                                 & 11.49                                                                  & 26.41         \\
1.0                                & 87.90         & 60.43             & 5.33                                                                 & 11.28                                                                  & 26.49         \\ \midrule
\textit{$\beta$}                   &               &                   &                                                                      &                                                                        &               \\
0                                  & 87.93         & 62.27             & 5.77                                                                 & 10.63                                                                  & 25.60         \\
0.25                               & 87.74         & 60.14             & 5.57                                                                 & 11.54                                                                  & 25.85         \\
\tableLineColor 0.50               & 86.97         & 58.44             & 5.55                                                                 & 11.82                                                                  & 25.23         \\
0.75                               & 87.44         & 59.79             & 5.72                                                                 & 11.55                                                                  & 26.83         \\
1.0                                & 87.27         & 61.63             & 5.73                                                                 & 11.06                                                                  & 25.86         \\ \bottomrule
\end{tabular}
\caption{Full general-domain results (Part 4/4) of the analysis on evaluation set size, $\alpha$, and $\beta$.}
\label{tab:full_analysis_1_general_4}
\end{table*}


\begin{table*}[tp]
\centering
\small
\begin{tabular}{lcccccccc}
\toprule
\multicolumn{1}{c}{\textbf{Model}}    & \textbf{Diabetic} & \textbf{PCAM} & \textbf{LC25000} & \textbf{Pollen} & \textbf{Amyloid CAA} & \textbf{Amyloid Diffuse} & \textbf{BloodMNIST} & \textbf{ChestMNIST} \\ \midrule
\textit{All}                          &                   &               &                  &                 &                      &                          &                     &                     \\
Dot                                   & 71.67             & 57.20         & 39.89            & 23.51           & 72.64                & 18.34                    & 16.87               & 5.50                \\
TracIn                                & 65.40             & 56.75         & 40.03            & 26.34           & 90.73                & 12.84                    & 16.69               & 3.69                \\
TRAK                                  & 43.23             & 59.23         & 39.57            & 12.88           & 77.16                & 14.74                    & 25.93               & 3.60                \\
\tableLineColor \textbf{CHIPS (ours)} & 68.35             & 63.66         & 40.51            & 25.96           & 88.06                & 12.38                    & 21.43               & 6.24                \\ \midrule
\textit{Logit-only}                   &                   &               &                  &                 &                      &                          &                     &                     \\
Dot                                   & 62.45             & 59.27         & 31.28            & 25.96           & 13.48                & 63.18                    & 20.90               & 9.60                \\
TracIn                                & 58.24             & 61.00         & 33.89            & 16.09           & 8.30                 & 71.48                    & 19.38               & 10.15               \\
TRAK                                  & 64.29             & 62.24         & 35.12            & 26.02           & 14.73                & 59.51                    & 17.36               & 11.74               \\
\tableLineColor \textbf{CHIPS (ours)} & 59.94             & 61.30         & 35.36            & 20.93           & 13.72                & 70.29                    & 17.89               & 21.50               \\ \midrule
\textit{Visual-only}                  &                   &               &                  &                 &                      &                          &                     &                     \\
Dot                                   & 61.96             & 56.15         & 37.71            & 23.00           & 13.33                & 47.26                    & 20.05               & 6.74                \\
TracIn                                & 59.11             & 53.51         & 34.99            & 18.16           & 28.44                & 31.75                    & 18.15               & 13.78               \\
TRAK                                  & 71.23             & 59.48         & 44.83            & 32.75           & 23.59                & 24.27                    & 15.02               & 15.37               \\
\tableLineColor \textbf{CHIPS (ours)} & 64.01             & 57.61         & 41.25            & 42.61           & 33.29                & 53.67                    & 20.93               & 5.48                \\ \midrule
\textit{Text-only}                    &                   &               &                  &                 &                      &                          &                     &                     \\
Dot                                   & 71.29             & 61.81         & 40.77            & 24.07           & 40.91                & 12.56                    & 17.10               & 20.57               \\
TracIn                                & 61.86             & 60.90         & 39.23            & 25.71           & 77.58                & 13.00                    & 18.65               & 11.51               \\
TRAK                                  & 51.61             & 57.08         & 39.55            & 21.36           & 34.21                & 46.31                    & 20.61               & 11.22               \\
\tableLineColor \textbf{CHIPS (ours)} & 69.76             & 60.02         & 46.83            & 35.80           & 25.76                & 55.13                    & 18.65               & 12.80               \\ \bottomrule
\end{tabular}
\caption{Full medical-domain results (Part 1/2) of the analysis on end-point subspace.}
\label{tab:full_analysis_2_medical_1}
\end{table*}

\begin{table*}[tp]
\centering
\small
\begin{tabular}{lccccccccc}
\toprule
\multicolumn{1}{c}{\textbf{Model}}    & \textbf{ChestXray14} & \textbf{Derma} & \textbf{Oct} & \textbf{OrganA} & \textbf{OrganC} & \textbf{OrganS} & \textbf{Path} & \textbf{Retina} & \multicolumn{1}{l}{\textbf{Tissue}} \\ \midrule
\textit{All}                          &                      &                &              &                 &                 &                 &               &                 & \multicolumn{1}{l}{}                \\
Dot                                   & 6.17                 & 12.17          & 20.60        & 9.48            & 10.36           & 11.19           & 25.42         & 33.75           & 4.16                                \\
TracIn                                & 6.53                 & 12.02          & 18.00        & 13.98           & 12.85           & 15.28           & 22.98         & 40.50           & 4.96                                \\
TRAK                                  & 25.84                & 12.17          & 20.90        & 10.74           & 10.43           & 11.22           & 28.68         & 26.75           & 4.46                                \\
\tableLineColor \textbf{CHIPS (ours)} & 16.63                & 11.87          & 18.20        & 12.03           & 9.64            & 10.71           & 23.90         & 30.25           & 5.29                                \\ \midrule
\textit{Logit-only}                   &                      &                &              &                 &                 &                 &               &                 & \multicolumn{1}{l}{}                \\
Dot                                   & 9.63                 & 11.72          & 24.10        & 9.22            & 9.10            & 7.31            & 13.89         & 21.50           & 6.13                                \\
TracIn                                & 10.82                & 11.77          & 22.40        & 10.54           & 8.82            & 7.50            & 15.08         & 22.25           & 5.21                                \\
TRAK                                  & 6.84                 & 11.92          & 22.80        & 9.20            & 7.80            & 7.34            & 15.10         & 23.25           & 5.53                                \\
\tableLineColor \textbf{CHIPS (ours)} & 9.49                 & 12.02          & 22.50        & 9.42            & 8.68            & 7.68            & 16.18         & 19.00           & 5.16                                \\ \midrule
\textit{Visual-only}                  &                      &                &              &                 &                 &                 &               &                 & \multicolumn{1}{l}{}                \\
Dot                                   & 3.66                 & 11.62          & 21.30        & 11.04           & 9.30            & 10.72           & 24.21         & 22.75           & 4.72                                \\
TracIn                                & 6.45                 & 10.97          & 19.80        & 11.80           & 12.04           & 13.90           & 24.25         & 26.00           & 4.95                                \\
TRAK                                  & 6.20                 & 11.77          & 18.90        & 11.51           & 9.90            & 10.42           & 17.19         & 40.00           & 5.56                                \\
\tableLineColor \textbf{CHIPS (ours)} & 19.32                & 12.02          & 19.60        & 10.91           & 10.69           & 12.22           & 23.69         & 26.50           & 5.31                                \\ \midrule
\textit{Text-only}                    &                      &                &              &                 &                 &                 &               &                 & \multicolumn{1}{l}{}                \\
Dot                                   & 11.54                & 11.97          & 22.50        & 11.72           & 11.81           & 12.30           & 23.79         & 33.00           & 4.55                                \\
TracIn                                & 10.52                & 12.07          & 19.80        & 10.27           & 12.34           & 13.11           & 23.97         & 40.25           & 4.69                                \\
TRAK                                  & 18.09                & 12.27          & 22.30        & 12.52           & 11.82           & 12.82           & 25.97         & 27.25           & 6.31                                \\
\tableLineColor \textbf{CHIPS (ours)} & 19.51                & 11.42          & 19.70        & 10.09           & 9.31            & 9.67            & 17.66         & 39.25           & 5.09                                \\ \bottomrule
\end{tabular}
\caption{Full medical-domain results (Part 2/2) of the analysis on end-point subspace.}
\label{tab:full_analysis_2_medical_2}
\end{table*}


\begin{table*}[tp]
\centering
\small
\begin{tabular}{lcccccccc}
\toprule
\multicolumn{1}{c}{\textbf{Model}} & \textbf{Diabetic} & \textbf{PCAM} & \textbf{LC25000} & \textbf{Pollen} & \textbf{Amyloid CAA} & \textbf{Amyloid Diffuse} & \textbf{BloodMNIST} & \textbf{ChestMNIST} \\ \midrule
\textit{CountSketch}               &                   &               &                  &                 &                      &                          &                     &                     \\
2k                                 & 65.13             & 59.60         & 48.05            & 21.43           & 35.20                & 32.39                    & 16.69               & 4.16                \\
4k                                 & 68.35             & 63.66         & 40.51            & 25.96           & 88.06                & 12.38                    & 21.43               & 6.24                \\
8k                                 & 63.37             & 57.97         & 39.76            & 17.03           & 73.36                & 14.87                    & 22.45               & 2.14                \\
16k                                & 62.10             & 55.73         & 40.00            & 21.50           & 79.72                & 23.67                    & 22.27               & 3.38                \\ \midrule
\textit{Sparse}                    &                   &               &                  &                 &                      &                          &                     &                     \\
2k                                 & 66.86             & 61.11         & 40.91            & 25.27           & 18.14                & 68.09                    & 17.45               & 5.81                \\
4k                                 & 68.68             & 60.15         & 44.19            & 23.76           & 23.67                & 39.46                    & 10.38               & 5.01                \\
8k                                 & 65.29             & 65.76         & 43.33            & 27.03           & 41.30                & 12.49                    & 14.53               & 4.12                \\
16k                                & 68.51             & 62.10         & 43.89            & 35.89           & 43.41                & 56.96                    & 24.26               & 8.14                \\ \midrule
\textit{SRHT}                      &                   &               &                  &                 &                      &                          &                     &                     \\
2k                                 & 32.69             & 60.39         & 47.09            & 33.19           & 68.64                & 39.88                    & 18.91               & 6.90                \\
4k                                 & 69.35             & 58.69         & 39.04            & 14.39           & 88.25                & 15.72                    & 18.65               & 8.84                \\
8k                                 & 41.47             & 56.59         & 39.31            & 27.66           & 88.67                & 20.62                    & 14.21               & 1.82                \\
16k                                & 65.89             & 58.36         & 40.35            & 41.04           & 32.65                & 44.25                    & 12.72               & 3.77 \\ \bottomrule              
\end{tabular}
\caption{Full medical-domain results (Part 1/2) of the analysis on JL random projection.}
\label{tab:full_analysis_3_medical_1}
\end{table*}

\begin{table*}[tp]
\centering
\small
\begin{tabular}{lccccccccc}
\toprule
\multicolumn{1}{c}{\textbf{Model}} & \textbf{ChestXray14} & \textbf{Derma} & \textbf{Oct} & \textbf{OrganA} & \textbf{OrganC} & \textbf{OrganS} & \textbf{Path} & \textbf{Retina} & \multicolumn{1}{l}{\textbf{Tissue}} \\ \midrule
\textit{CountSketch}               &                      &                &              &                 &                 &                 &               &                 & \multicolumn{1}{l}{}                \\
2k                                 & 10.10                & 11.32          & 17.80        & 14.33           & 15.17           & 16.28           & 21.18         & 28.00           & 5.63                                \\
4k                                 & 16.63                & 11.87          & 18.20        & 12.03           & 9.64            & 10.71           & 23.90         & 30.25           & 5.29                                \\
8k                                 & 21.02                & 12.77          & 17.40        & 14.20           & 14.00           & 13.61           & 27.45         & 30.50           & 5.88                                \\
16k                                & 20.71                & 12.37          & 21.40        & 14.04           & 14.62           & 13.73           & 25.96         & 36.00           & 8.09                                \\ \midrule
\textit{Sparse}                    &                      &                &              &                 &                 &                 &               &                 & \multicolumn{1}{l}{}                \\
2k                                 & 11.39                & 11.12          & 23.90        & 14.60           & 13.47           & 13.90           & 26.13         & 35.75           & 6.81                                \\
4k                                 & 10.47                & 11.72          & 20.60        & 10.70           & 11.31           & 11.35           & 24.35         & 30.00           & 6.68                                \\
8k                                 & 7.93                 & 12.17          & 15.10        & 11.33           & 12.20           & 15.33           & 23.16         & 24.50           & 5.33                                \\
16k                                & 7.06                 & 11.42          & 20.70        & 12.30           & 12.07           & 12.35           & 28.83         & 31.75           & 5.60                                \\ \midrule
\textit{SRHT}                      &                      &                &              &                 &                 &                 &               &                 & \multicolumn{1}{l}{}                \\
2k                                 & 13.51                & 11.57          & 22.50        & 10.75           & 10.47           & 10.39           & 23.23         & 27.00           & 5.27                                \\
4k                                 & 9.45                 & 12.02          & 17.10        & 9.60            & 10.72           & 11.03           & 31.63         & 36.00           & 6.47                                \\
8k                                 & 22.18                & 11.87          & 17.40        & 13.21           & 13.21           & 13.23           & 28.70         & 27.00           & 5.91                                \\
16k                                & 7.81                 & 11.22          & 18.30        & 12.30           & 12.48           & 11.63           & 24.75         & 26.25           & 6.55 \\ \bottomrule                               
\end{tabular}
\caption{Full medical-domain results (Part 2/2) of the analysis on JL random projection.}
\label{tab:full_analysis_3_medical_2}
\end{table*}

\section{Additional Results}
\label{app:additional_results}

We provide additional experiments on MedTrinity \cite{medtrinity} in Tab.~\ref{tab:medtrinity_medical_1} and Tab.~\ref{tab:medtrinity_medical_2}.

\begin{table*}[tp]
\centering
\small
\begin{tabular}{lcccccccc}
\toprule
\multicolumn{1}{c}{\textbf{Model}} & \textbf{Diabetic} & \textbf{PCAM} & \textbf{LC25000} & \textbf{Pollen} & \textbf{Amyloid CAA} & \textbf{Amyloid Diffuse} & \textbf{BloodMNIST} & \textbf{ChestMNIST} \\ \midrule
\textit{B32-400M}                  &                   &               &                  &                 &                      &                          &                     &                     \\
Random 10\%                        & 63.65             & 50.20         & 32.51            & 72.91           & 97.26                & 12.82                    & 14.88               & 6.86                \\
Random 50\%                        & 34.53             & 52.70         & 25.68            & 72.91           & 97.76                & 12.50                    & 9.97                & 28.41               \\ \midrule
\textit{B32-CC}                    &                   &               &                  &                 &                      &                          &                     &                     \\
Random 10\%                        & 3.45              & 63.08         & 8.80             & 11.00           & 1.84                 & 87.94                    & 29.49               & 2.75                \\
Random 50\%                        & 3.11              & 56.25         & 23.89            & 51.85           & 2.86                 & 87.94                    & 21.60               & 17.87               \\ \midrule
\textit{B16-400M}                  &                   &               &                  &                 &                      &                          &                     &                     \\
Random 10\%                        & 10.74             & 56.77         & 39.28            & 7.23            & 88.27                & 59.82                    & 14.44               & 9.08                \\
Random 50\%                        & 17.08             & 66.73         & 32.91            & 8.30            & 96.46                & 14.03                    & 7.28                & 1.24                \\ \midrule
\textit{B16-CC}                    &                   &               &                  &                 &                      &                          &                     &                     \\
Random 10\%                        & 46.68             & 50.49         & 6.40             & 21.06           & 60.28                & 87.95                    & 6.58                & 1.56                \\
Random 50\%                        & 22.16             & 50.83         & 17.95            & 61.28           & 62.36                & 87.95                    & 8.21                & 1.07             \\ \bottomrule
\end{tabular}
\caption{Medical-domain results (Part 1/2) of MedTrinity \cite{medtrinity} dataset.}
\label{tab:medtrinity_medical_1}
\end{table*}

\begin{table*}[tp]
\centering
\small
\begin{tabular}{lccccccccc}
\toprule
\multicolumn{1}{c}{\textbf{Model}} & \textbf{ChestXray14} & \textbf{Derma} & \textbf{Oct} & \textbf{OrganA} & \textbf{OrganC} & \textbf{OrganS} & \textbf{Path} & \textbf{Retina} & \multicolumn{1}{l}{\textbf{Tissue}} \\ \midrule
\textit{B32-400M}                  &                   &               &                  &                 &                      &                          &                     &                     &                      \\
Random 10\%                        & 6.05              & 13.97         & 25.00            & 10.77           & 6.86                 & 5.30                     & 34.50               & 9.75                & 5.31                 \\
Random 50\%                        & 11.29             & 11.32         & 22.60            & 10.68           & 6.69                 & 6.30                     & 35.72               & 8.50                & 5.02                 \\ \midrule
\textit{B32-CC}                    &                   &               &                  &                 &                      &                          &                     &                     &                      \\
Random 10\%                        & 4.42              & 9.63          & 25.00            & 16.22           & 8.98                 & 6.24                     & 31.91               & 14.25               & 9.18                 \\
Random 50\%                        & 7.56              & 11.07         & 25.00            & 21.59           & 14.56                & 14.72                    & 46.03               & 16.00               & 8.01                 \\ \midrule
\textit{B16-400M}                  &                   &               &                  &                 &                      &                          &                     &                     &                      \\
Random 10\%                        & 4.78              & 10.62         & 12.40            & 13.98           & 8.57                 & 7.73                     & 27.73               & 12.75               & 9.46                 \\
Random 50\%                        & 2.86              & 10.62         & 16.20            & 13.17           & 11.93                & 10.32                    & 30.28               & 7.50                & 9.70                 \\ \midrule
\textit{B16-CC}                    &                   &               &                  &                 &                      &                          &                     &                     &                      \\
Random 10\%                        & 9.89              & 10.62         & 25.90            & 13.04           & 12.71                & 13.33                    & 35.65               & 5.00                & 6.15                 \\
Random 50\%                        & 6.45              & 18.50         & 25.40            & 19.92           & 14.79                & 15.54                    & 42.45               & 5.75                & 4.92                 \\ \bottomrule
\end{tabular}
\caption{Medical-domain results (Part 2/2) of MedTrinity \cite{medtrinity} dataset.}
\label{tab:medtrinity_medical_2}
\end{table*} 


\end{document}